\documentclass[10pt,journal,compsoc]{IEEEtran}
\usepackage{setspace}
\usepackage{amsmath}
\usepackage{amssymb}
\usepackage{mathtools}
\usepackage{amsthm}
\usepackage{multirow}

\usepackage{lineno}
\usepackage{graphicx}	% A package for graphics use (see figures)
\usepackage{amsmath}
\usepackage{amssymb}
\usepackage{subfig}
\usepackage{wrapfig}
\usepackage{enumitem}
\usepackage[ruled,vlined]{algorithm2e}
\usepackage{booktabs}
\usepackage{xcolor}
\usepackage{diagbox}

\usepackage{color}

\ifCLASSOPTIONcompsoc
  \usepackage[nocompress]{cite}
\else
  \usepackage{cite}
\fi
\ifCLASSINFOpdf
\else
\fi
\begin{document}
\title{PCDNF: Revisiting Learning-based Point Cloud Denoising via Joint Normal Filtering}
%\title{Point Cloud Denoising via Joint Normal Filtering}
% use \thanks{} to gain access to the first footnote areap

\author{Zheng~Liu,
        Yaowu~Zhao,
        Sijing~Zhan,
        Yuanyuan~Liu,
        Renjie~Chen$^{\dagger}$,
        Ying~He
\IEEEcompsocitemizethanks{
\IEEEcompsocthanksitem Z. Liu, Y. Zhao, S. Zhan, and Y. Liu are with School of Computer Science, China University of Geosciences (Wuhan).
\IEEEcompsocthanksitem R. Chen is with School of Mathematical Sciences, University of Science and Technology of China.
\IEEEcompsocthanksitem Y. He is with School of Computer Science and Engineering, Nanyang Technological University.
}
\thanks{$^{\dagger}$Corresponding author. E-mail:renjiec@ustc.edu.cn.}}
\IEEEtitleabstractindextext{
\begin{abstract}
Point cloud denoising is a fundamental and challenging problem in geometry processing. Existing methods typically involve direct denoising of noisy input or filtering raw normals followed by point position updates. Recognizing the crucial relationship between point cloud denoising and normal filtering, we re-examine this problem from a multitask perspective and propose an end-to-end network called PCDNF for joint normal filtering-based point cloud denoising. We introduce an auxiliary normal filtering task to enhance the network's ability to remove noise while preserving geometric features more accurately. Our network incorporates two novel modules. First,  we design a shape-aware selector to improve noise removal performance by constructing latent tangent space representations for specific points, taking into account learned point and normal features as well as geometric priors. Second, we develop a feature refinement module to fuse point and normal features, capitalizing on the strengths of point features in describing geometric details and normal features in representing geometric structures, such as sharp edges and corners. This combination overcomes the limitations of each feature type and better recovers geometric information.
Extensive evaluations, comparisons, and ablation studies demonstrate that the proposed method outperforms state-of-the-art approaches in both point cloud denoising and normal filtering.
\end{abstract}
}
\maketitle

\IEEEdisplaynontitleabstractindextext

% \fi
\IEEEpeerreviewmaketitle

\IEEEraisesectionheading{\section{Introduction}\label{sec:introduction}}
\IEEEPARstart{P}{oint} clouds are widely used in various fields, including computer graphics, 3D computer vision, photogrammetry, autonomous driving, simultaneous localization, and mapping (SLAM), among others.
With the rapid development of modern 3D digital acquisition devices, such as LiDAR and depth cameras, more and more 3D models are routinely obtained and stored as point clouds in shape repositories. However, due to physical measurement and reconstruction errors, the acquired point clouds are inevitably corrupted by noise   \cite{wei2021geodualcnn}. Noise not only degrades the visual quality of 3D models but also causes unexpected problems  in downstream applications \cite{chen2022repcd,zhang2020pointfilter}.
Therefore, point cloud denoising is highly desired and often considered the first step in geometry processing.
Due to the high-frequency nature of both noise and geometric features, it is challenging to distinguish and recover features while removing noise from point clouds. Point cloud denoising has been a topic of extensive research in the past two decades. While significant progress has been made, traditional denoising methods ~\cite{oztireli2009feature,lu2017gpf,chen2019multi,chen2019structure,lu2020low,liu2020feature,Liu2022MeshTGV} typically require numerous parameters and tedious parameter tuning. The tuning process is not only time-consuming but also crucial for achieving promising results.

The success of deep neural networks in image processing has recently led to the adoption of data-driven approaches for various tasks in point cloud processing, including denoising. Deep learning methods are automatic and eliminate  the need for parameter tuning, making them suitable for a wider range of 3D models than traditional methods. Existing deep denoising methods can generally be classified into one- and two-stage methods. 

One-stage methods, such as PointCleanNet \cite{rakotosaona2020pointcleannet}, Pointfilter\cite{zhang2020pointfilter}, Re\textsc{PCD}-\textsc{N}et\cite{chen2022repcd}, typically use point feature representations to regress a displacement per noisy point and adjust its position to the ground-truth directly. Due to a lack of consideration of normal information, PointCleanNet \cite{rakotosaona2020pointcleannet} and Re\textsc{PCD}-\textsc{N}et\cite{chen2022repcd} blur sharp features.
Pointfilter \cite{zhang2020pointfilter}, which incorporates normal information in its loss function, can better preserve sharp features but may oversmooth geometric details.

Two-stage methods, which filter normals followed by updating their locations, have gained widespread attention due to their ability to incorporate local geometry information \cite{lu2020deep,wei2021geodualcnn}.
The main difference among the two-stage methods is in their normal filtering networks.
Similar to Pointfilter\cite{zhang2020pointfilter}, small-scale geometric features including fine details may get blurred when using only the learned normal-based features to update point positions.
To recover detailed features accurately, the method in \cite{lu2020deep} relies on the additional feature detection network, while GeoDualCNN \cite{wei2021geodualcnn} needs the guidance of geometry expertise and a feature-preserving position updating algorithm.

The aforementioned learning-based methods either directly smooth noisy points or filter raw normals followed by updating point positions, making them unsuitable for joint denoising and normal filtering. However, denoising and normal filtering tasks are intrinsically intertwined, influencing and benefiting each other. If better normals can be estimated from the noisy point cloud, the performance of the denoising task can be significantly improved, and vice versa. To date, no existing work can perform denoising and normal filtering tasks jointly.

This observation has led us to develop a point-normal feature interaction network within a joint task paradigm, which can be regarded as a special category of multitask learning. For the rest of this paper, we refer to our multitask learning specifically to joint task learning. Unlike previous work, we present a unified architecture for jointly learning point cloud denoising and normal filtering. The proposed network comprises two branches — one for point cloud denoising and the other for normal filtering — that can mutually benefit each other. The combined technique leverages the best properties of each task while attempting to overcome their respective weaknesses.

Specifically, the proposed network consists of four modules: the multiscale feature extractor, shape-aware selector, feature refinement, and the decoder. The feature extractor utilizes DGCNN \cite{wang2019dynamic} as the backbone for learning point and normal feature representations in a multiscale manner. Next, the learned point and normal features, combined with geometric priors, are fed into the shape-aware selector to construct the latent tangent space for the specific point, which helps reduce the negative impact of points not within the tangent space.

Following this, we design a feature refinement module consisting of two units — feature augmentation and fusion — to enhance the network's ability to accurately recover geometric features. The feature augmentation unit aggregates neighboring features to obtain richer point and normal feature representations, while the feature fusion unit integrates the augmented point and normal features to better preserve both structural and detailed geometric features. Finally, the integrated features are fed into the decoder to predict the denoised point coordinates and filtered normals.

To summarize, the main contributions of this work include the following:
\begin{itemize}
  \item We introduce  a novel architecture for jointly learning point cloud denoising and normal filtering. To the best of our knowledge, this is the first end-to-end framework for point cloud denoising that adopts a multitask perspective.
  \item We develop a shape-aware selecting module to enhance denoising performance by reducing the adverse effects of neighboring points outside the latent tangent space of the specific point. This module represents the tangent space using learned point and normal features combined with geometric priors.
  \item We design a feature refinement module to boost the network's ability for preserving geometric features. This module first expands the receptive fields of learned features and subsequently integrates the learned point and normal features to better recover various geometric features, such as structural and detailed features.
  \item  Qualitative and quantitative experiments on both synthetic and scanned data demonstrate that our network outperforms state-of-the-art methods in both denoising and normal filtering tasks.
\end{itemize}

\section{Related work} \label{sec:relatedWork}
\subsection{Point Cloud Denoising}
As a fundamental geometry processing problem, point cloud denoising has garnered significant attention over the past few decades. Given the vast amount of literature on denoising techniques, providing an exhaustive review is beyond the scope of this paper. We refer interested readers to \cite{zhou2022point} for a comprehensive review. In this section, we briefly discuss traditional methods before focusing on  recent learning-based techniques.

\textbf{Traditional methods}. Moving Least Squares (MLS)-related methods \cite{alexa2003computing,amenta2004defining,fleishman2005robust,oztireli2009feature} project the input point set onto the approximated underlying surface iteratively. Originally designed for reconstructing noise-free surfaces, these classical methods assume piecewise smooth priors for the underlying surface, which can result in the smoothing of geometric features. Later on, Lipman et al. \cite{lipman2007parameterization} proposed the pioneering Locally Optimal Projection (LOP) method, which has been proven successful for point cloud consolidation. LOP and its variants \cite{huang2009consolidation, huang2013edge, preiner2014continuous, lu2017gpf} aim to generate a point set that describes the underlying surface while maintaining a uniform distribution. Despite their ability to robustly remove noise and produce uniformly sampled results, LOP-related methods struggle to preserve geometric features in the presence of large noise. Optimization-based methods, on the other hand, formulate the denoising process as optimization problems with suitable priors. Among them, sparse optimization methods, such as \cite{avron2010, sun2015denoising, mattei2017point, liu2020feature}, effectively preserve geometric features (particularly sharp ones) by leveraging the sparsity of geometric features over underlying surfaces. Recently, low-rank and dictionary learning techniques \cite{digne2017sparse, chen2019multi, lu2020low, wang2022rethinking, sun2022structure} have gained attention due to their ability to preserve structural repetition on underlying surfaces by exploiting self-similarity characteristics. Hu et al. \cite{huweifeaturegraph} proposed a feature graph learning approach and employed it for point cloud denoising with points' positions and normals as features, leading to impressive denoising results. Although optimization-based methods excel in preserving certain geometric features, their reliance on geometric priors may hinder their performance in preserving other types of features. In general, traditional methods involve complex computations or optimization problems and necessitate a tedious trial-and-error process to achieve satisfactory results.

\textbf{Learning-based methods}.
Recently, with the development of neural networks\cite{qi2017pointnet,qi2017pointnet++,wang2019dynamic,liu2021deep,Lu_IterativePFN}, deep learning techniques have been introduced into point cloud denoising extensively and achieved impressive results.
Roveri et al. \cite{roveri2018pointpronets} proposed PointProNet, a fully differentiable denoising architecture based on 2D CNN, which converts unordered points to regularly sampled height maps.
EC-Net\cite{yu2018ec} and DMRDenoise\cite{luo2020differentiable} mainly focus on upsampling and consolidating techniques over point clouds. EC-Net designs an edge-aware consolidation network to denoise point clouds. This method preserves sharp geometric features but retains noise to some extent. Although DMRDenoise can remove noise successfully during the downsampling stage, it may blur geometric features. TotalDenoising, developed by Hermosilla et al. \cite{hermosilla2019total}, introduces a spatial prior that steers converge to underlying surfaces without supervision. However, as an unsupervised method, TotalDenoising is sensitive to large noise and may suffer shrinkage artifacts. GPDNet \cite{pistilli2020learning} uses the graph-convolutional neural network for denoising. Luo and Hu \cite{luo2021score} proposed a paradigm of denoising by exploiting the distribution model of noisy point clouds. Most of the above learning-based methods cannot effectively preserve geometric features, especially sharp features. Moreover, the noise removal performance of these methods decreases evidently as the noise level increases.

More recently, some techniques belonging to the two-stage paradigm \cite{lu2020deep,wei2021geodualcnn} - normal filtering followed by updating point coordinates - have been developed for feature-preserving denoising. Lu et al. \cite{lu2020deep} classified the noisy input as feature points and non-feature points and then predicted multi-normal on the feature points to preserve sharp features.
Wei et al. \cite{wei2021geodualcnn} proposed a geometry-supporting dual convolutional neural network (GeoDualCNN) to filter normals and then updated point coordinates to match the filtered normals. Although GeoDualCNN can produce promising results, the requirement of computing the extra homogeneous neighborhood for each point limits its end-to-end applicability. Luo and Hu \cite{luo2021score} proposed to first estimate the gradient for the noisy point cloud and then perform denoising via gradient ascent. Chen et al. \cite{huwei_deep_resampling} proposed a global and continuous gradient field model, which can resample degraded point clouds via gradient ascent with the introduced graph Laplacian regularizer.

Compared to these two-stage methods, an alternative paradigm involves  developing networks that directly predict displacements of the noisy point cloud and then apply the predicted displacements to reposition point coordinates. Several methods have been developed following this paradigm \cite{rakotosaona2020pointcleannet, zhang2020pointfilter, chen2022repcd}. PointCleanNet \cite{rakotosaona2020pointcleannet} first excludes outliers and then predicts displacement vectors for the remaining noisy points. This method may retain extra noise on the denoised results and tends to smooth sharp features to varying degrees. To address this issue, Pointfilter \cite{zhang2020pointfilter} introduces a loss function preserving features with an encoder-decoder framework. Although Pointfilter can preserve sharp features, it cannot recover small-scale geometric features well. Later, Chen et al. \cite{chen2022repcd} proposed a feature-aware recurrent architecture to learn more representative features for recovering multiscale geometric features effectively. However, in the case of large noise, their method seems to be difficult to keep a balance between noisy removal and geometric feature recovery. More recently, Edirimuni et al. \cite{Lu_contrastive} introduced a contrastive learning framework to generate effective patch-wise representations with noise corruption as augmentation, which can infer both  displacements and normals simultaneously.

\subsection{Point Cloud Normal Filtering}
Point normals, which indicate the orientation of the scanned surface, are essential signals that have been widely applied in various practical problems, such as surface reconstruction \cite{kazhdan2013screened,hou2022iterative}, 3D descriptors \cite{rusu2009fast}, and registration \cite{zhang2021fast}. However, accurately estimating point normals is challenging, as captured point clouds are inevitably corrupted by noise and outliers. Consequently, normal filtering has been extensively studied over the past decades. Due to the vast amount of literature on normal filtering, we only review the learning-based methods related to our work.

Since the pioneering work of Qi et al. \cite{qi2017pointnet}, numerous studies have extended and applied the PointNet architecture to point cloud processing problems. PCPNet \cite{guerrero2018pcpnet} was the first to apply the PointNet architecture for estimating normals from noisy point clouds, addressing the normal filtering task.
Zhou et al. \cite{zhou2020normal} proposed a plane constraint mechanism to divide neighborhood points into main plane points and error points, using only the learned features of the main plane points to regress normals. Their method is robust to noise and neighborhood scales. To overcome the oversmoothing artifacts, Nesti-Net \cite{ben2019nesti} introduces mixtures of experts to predict the optimal neighborhood scale instead of simply concatenating multiple scales together. Their multiscale strategy can improve performance effectively but leads to evident time consumption.
Zhou et al. \cite{zhou2022fast} proposed a normal filter based on multipatch stitching.
Thanks to their patch-level architecture, their method can reduce computational costs and improve the robustness of noise removal.
Instead of directly predicting normals from the learned features, some methods \cite{ben2020deepfit,zhu2021adafit,zhang2022geometry} estimate the normal for a specific point by fitting a local underlying surface through its neighboring points and then compute the normal from the fitting surface.
These methods are based on the weighted least squares surface fitting of the local geometric neighborhood, which can improve the generalization ability of their networks on real scanning data \cite{zhu2021adafit}.
Cao et al. \cite{cao2021latent} learned a latent tangent space representation with a lightweight network and then utilized a differentiable RANSAC to estimate normals of the underlying surface.
Zhou et al. \cite{zhou2022refine} deployed a multi-feature scheme to capture geometric information from multiple feature representations and then updated normals in a refinement system.
Unlike these existing methods, we present a unified architecture for jointly learning point cloud denoising and normal filtering. The combined technique is able to apply the best properties of each of the two tasks, and try to overcome the weakness of both. Thus, it performs well in preserving geometric features and removing noise, and at the same time avoids the artifacts in the results.

\section{Method} \label{sec:method}
This section starts with an overview of our framework of point cloud denoising with joint normal filtering. Then, we present our network architecture, followed by elaborating on each module of the network. Finally, an end-to-end joint loss function is introduced.

\begin{figure*}
\begin{center}
\includegraphics[scale=0.54]{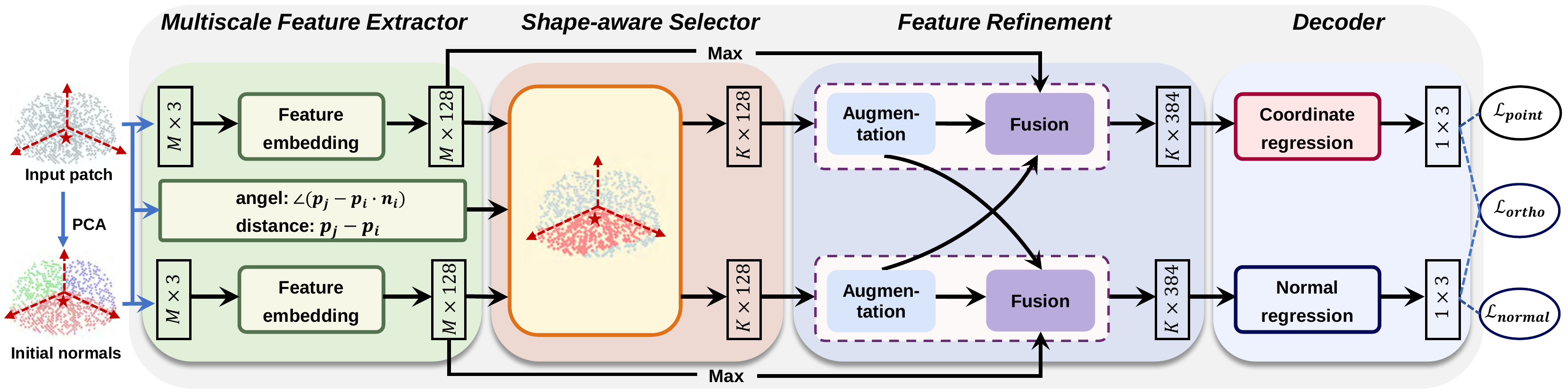}
\end{center}
   \caption{An overview of our joint point cloud denoising and normal filtering network, coined as PCDNF. PCDNF consists of four main modules: the multiscale feature extractor, shape-aware selector, feature refinement (including feature augmentation and fusion units), and the decoder. Given a noisy patch of a point (along with the corresponding raw normals of the patch), PCDNF is capable of concurrently predicting the coordinate and filtered normal of the specific point.
   }
\label{fig:NetworkArchitecture}
\end{figure*}

\subsection{Problem Statement}
Point cloud denoising is nontrivial due to the ill-posed nature of the problem. Many learning-based methods \cite{rakotosaona2020pointcleannet,zhang2020pointfilter,chen2022repcd} cast the noise as pointwise residuals (i.e., a displacement vector per input point), and try to predict the residual vectors in order to smooth the input point cloud. However, given only the point positions, it is still challenging to achieve satisfactory denoising results while preserving geometric features. Note that, high-quality normals can improve denoising performance; conversely, accurate normals can be computed from high-quality point clouds. Thus, as an alternative to directly predict the displacement vectors from the noisy input, we propose to perform point cloud denoising by combining position correction and normal filtering, which is stated as follows
\begin{equation} \label{eq:DenoisingAndFilteringModel}
(\widehat{\mathrm{D}}, \widehat{\mathrm{N}}) = \mathcal{F}(\mathrm{P}, \mathrm{N}), \ \ \ \mathrm{P} = \widehat{\mathrm{P}} + \widehat{\mathrm{D}},
\end{equation}
where $\mathrm{P}$ and $\widehat{\mathrm{P}}$ are the noisy point cloud and its corresponding denoised point cloud, $\widehat{\mathrm{D}}$ denotes the predicted displacement vectors, $\mathrm{N}$  denotes the raw normals of the noisy input and $\widehat{\mathrm{N}}$ denotes the corresponding predicted normals.
Our method aims to learn a mapping $\mathcal{F}:(\mathrm{P}, \mathrm{N}) \rightarrow (\widehat{\mathrm{D}}, \widehat{\mathrm{N}})$ for predicting displacement vectors and filtered normals simultaneously. Then, the denoised point cloud can be derived from \eqref{eq:DenoisingAndFilteringModel} straightforwardly.

\begin{figure*}[thb]
	\centering
	\subfloat[Feature embedding]{\label{fig:featureembedding}\includegraphics[width=0.5\textwidth]{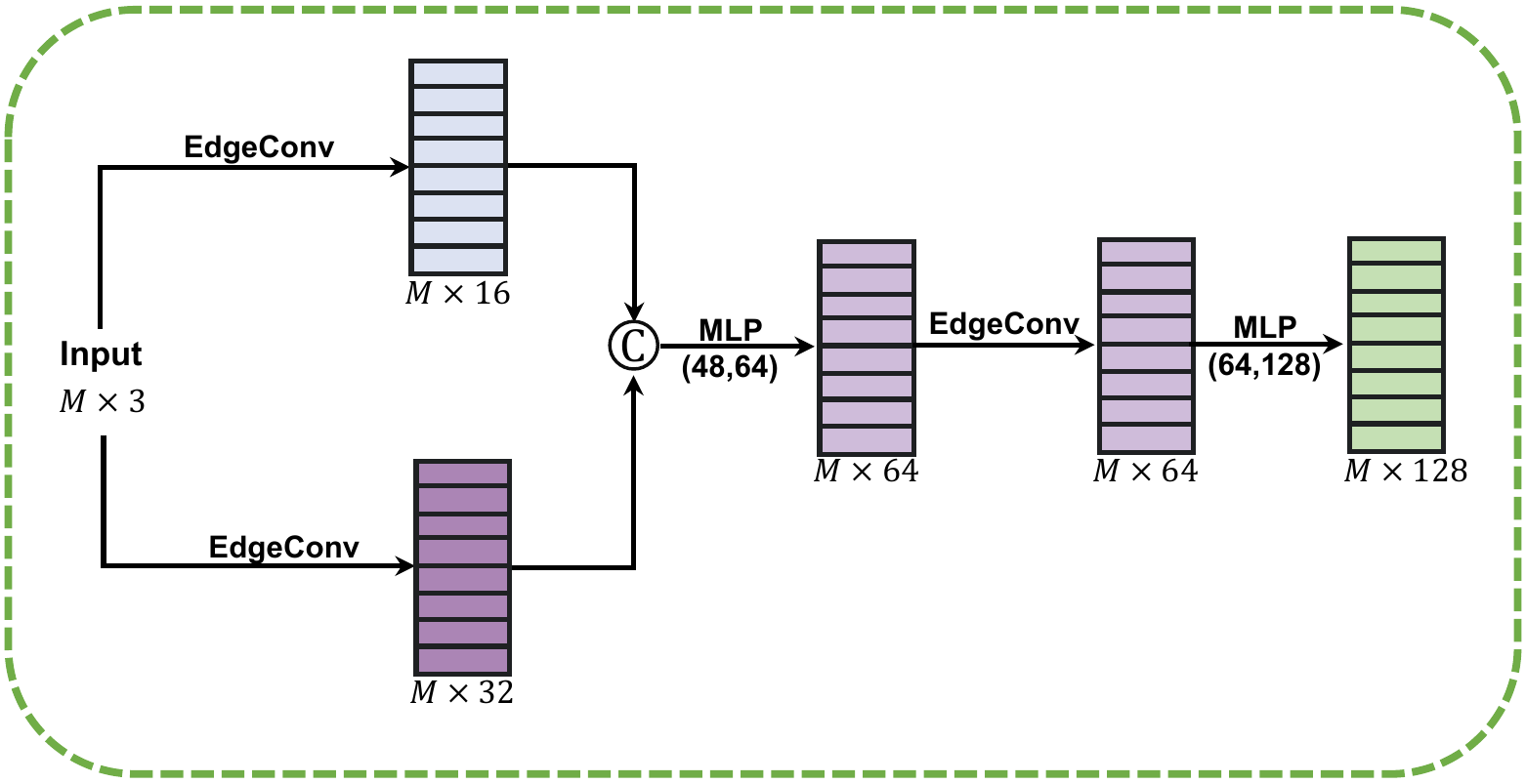}}
	\subfloat[Coordinate regression]{\label{fig:pointregression}\includegraphics[width=0.5\textwidth]{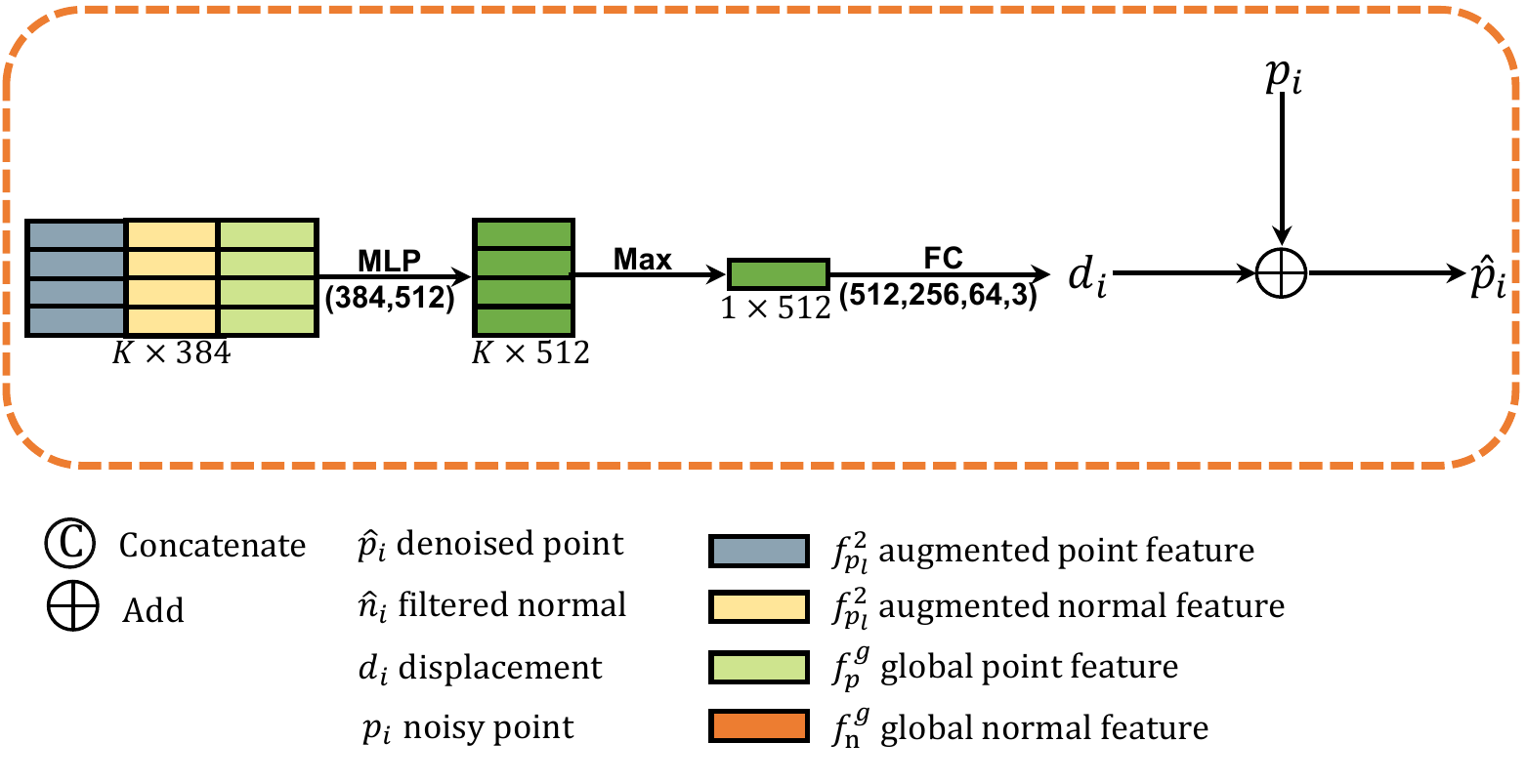}}\\
	\subfloat[Normal regression]{\label{fig:normalregression}\includegraphics[width=1\textwidth]{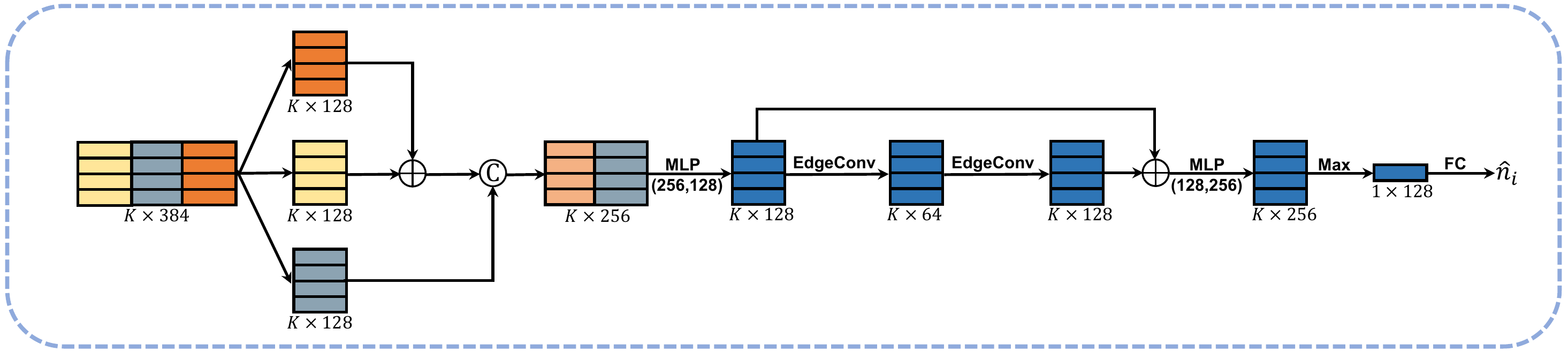}}
	\caption{
 % \color{red}
 {Illustration of (a) feature embedding and decoder consisting of (b) coordinate regression and (c) normal regression.
	\label{fig:featureNetwork}}}
\end{figure*}

\subsection{Network Architecture}
Based on the problem statement \eqref{eq:DenoisingAndFilteringModel}, we design a multitask network, dubbed PCDNF, for joint point cloud denoising and normal filtering. Fig. \ref{fig:NetworkArchitecture} shows the architecture of PCDNF. Our network consists of four modules: the multiscale feature extractor, shape-aware selector, feature refinement, and decoder.

Specifically, given a noisy patch and its corresponding raw normals, the multiscale feature extractor embeds the inputs into the coarse representations of point and normal features. Then, the shape-aware selector selects the points highly related to the specific point in terms of geometric information and coarse representations. These selected points form a latent tangent space of the specific point, making noise removal more effective. The feature refinement module first encodes local spatial information to augment the similarity representations and then fuses the augmented representations for better geometric feature preservation. Finally, the coordinate and filtered normal of the specific point can be predicted by the coordinate and normal regressors of the decoder.

\subsubsection{Multiscale feature extractor}
Given a point $p_i$ of the noisy point cloud $\mathrm{P}$,
a patch centered at $p_i$ is defined as
\begin{equation*}\label{eq:PatchDefinition}
  \mathrm{P}_i = \{ p_j |\  \| p_j - p_i \| < r  \} \in \mathbb{R}^{M \times 3}.
\end{equation*}
$M$ is the number of points within the patch and $r$ is the patch radius.
The corresponding raw normals of $\mathrm{P}_i$ can be denoted as $\mathrm{N}_i = \{ n_j \} \in \mathbb{R}^{M \times 3}$. It is known that the features learned from a single-scale receptive field cannot faithfully describe the local shape of the underlying surface. To address this issue, we propose a multiscale feature extractor using a series of EdgeConv operations \cite{wang2019dynamic}, which can learn multiscale discriminative representations in both the Euclidean and the feature spaces.

Given an input patch with coordinates $\mathrm{P}_i$ and normals $\mathrm{N}_i$, our multiscale feature extractor  learns a coarse point feature $\mathrm{F}_{\mathrm{P}_i}=\{f_{p_j} |  \ \forall p_j \in \mathrm{P}_i\}\in \mathbb{R}^{M\times 128}$ from $\mathrm{P}_i$, and a coarse normal feature $\mathrm{F}_{\mathrm{N}_i}=\{f_{n_j}|  \ \forall n_j \in \mathrm{N}_i\}\in \mathbb{R}^{M\times 128}$ from $\mathrm{N}_i$.
Fig. \ref{fig:featureembedding} provides details on our feature embeddings.
We elaborate on the process of extracting the coarse point feature $\mathrm{F}_{\mathrm{P}_i}$. The normal feature embedding is done similarly.
For any point $p_j \in \mathrm{P}_i$, to extract its pointwise feature $f_{p_j}$, we first  construct two k-nearest neighbors (kNN) graphs of different sizes in order to capture multiscale geometric information around $p_j$.
Then, we compute the features of $p_j$ using the EdgeConv operation \cite{wang2019dynamic} with graph sizes $k_1,k_2$ in the $1$-st layer as
\begin{equation*}
	\begin{aligned}
		f^{1,k_1}_{p_j}= \max_{q \in \mathcal{N}_{k_1}(p_j)} \mathrm{h}_{\theta}(p_j, q - p_j), \\
        f^{1,k_2}_{p_j}= \max_{q \in \mathcal{N}_{k_2}(p_j)} \mathrm{h}_{\theta}(p_j, q - p_j),
	\end{aligned}
\end{equation*}
where $\mathcal{N}_{k_1}(p_j)$ and $\mathcal{N}_{k_2}(p_j)$ denote the neighbors of point $p_j$ in the graphs with sizes $k_1$ and $k_2$, $\mathrm{h}_{\theta}$ denotes the multi-layer perception (MLP) parameterized by $\theta$,
$\max$ denotes the max pooling operation.
To make the pointwise feature more discriminative, we concatenate features $f^{1,k_1}_{p_j}$ and $f^{1,k_2}_{p_j}$, followed by MLP as
\begin{equation*}
  f^{1}_{p_j}=  \mathrm{MLP}({\mathrm{concat}(f^{1,k_1}_{p_j},f^{1,k_2}_{p_j})}),
\end{equation*}
where $\mathrm{concat}$ is the concatenation operation.
To further enlarge receptive fields and object relations in the feature space, we perform EdgeConv in the feature space followed by MLP as
\begin{equation*}
  f_{p_j}= \mathrm{MLP}\left(\max_{q \in \mathcal{N}_{k_3}(p_j)} \mathrm{h}_{\theta}(f^{1}_{p_j}, q - f^{1}_{p_j})  \right),
\end{equation*}
where $f_{p_j}$ is the pointwise feature of $p_j$ with graph size $k_3$.
Thus, the coarse point feature $\mathrm{F}_{\mathrm{P}_i}$ of patch $\mathrm{P}_i$ can be derived easily. Similarly, we can learn the pointwise normal feature $f_{n_j}$, and obtain the coarse normal feature $\mathrm{F}_{\mathrm{N}_i}$ of the patch.

\begin{figure}[thb]
  \centering
  \subfloat[]{\label{fig:2DPatch-a}\includegraphics[width=0.24\textwidth]{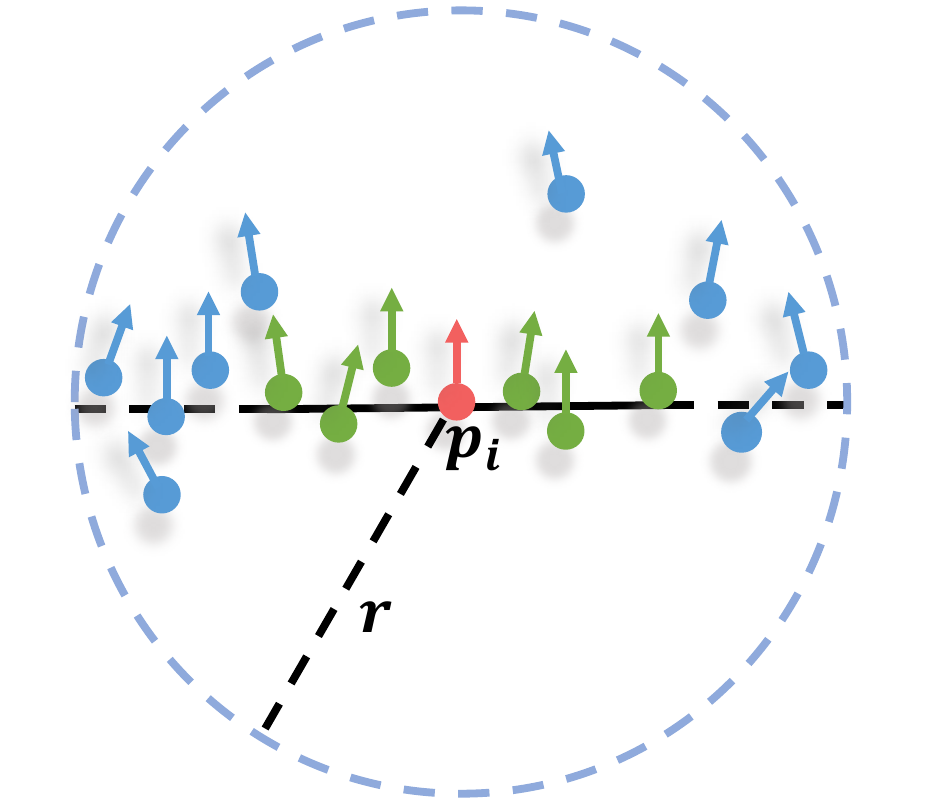}}
  \subfloat[]{\label{fig:2DPatch-b}\includegraphics[width=0.24\textwidth]{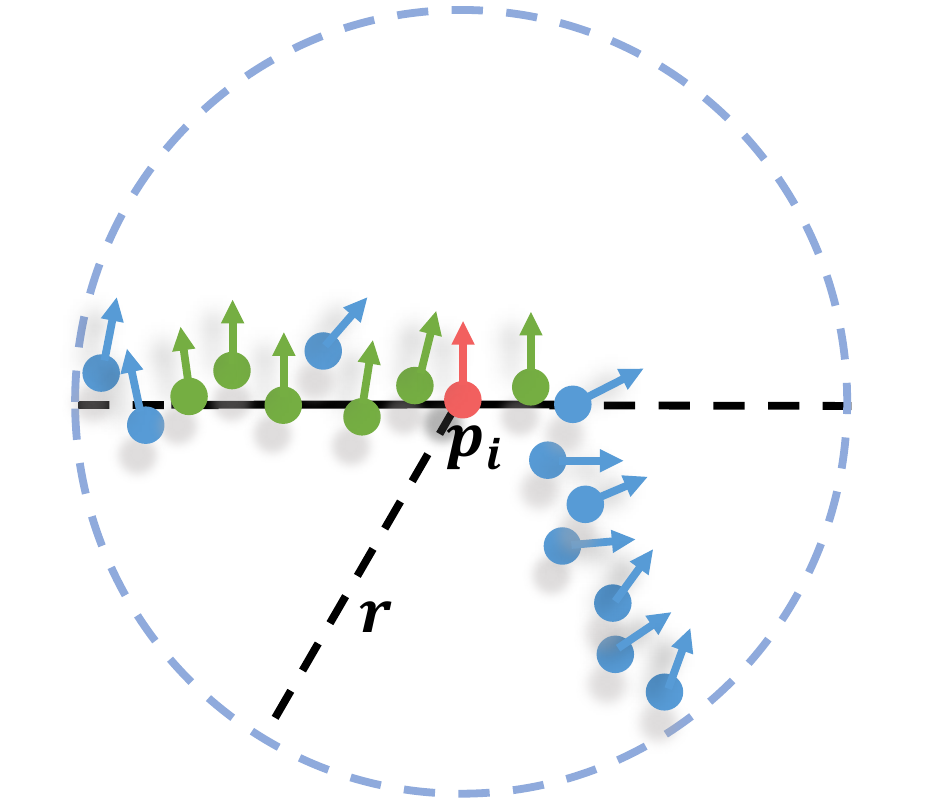}}
  \caption{Illustration for selecting similar points to the specific point $p_i$ plotted in red. The selected similar points in the same latent space of $p_i$ are plotted in green, and the negative points are plotted in blue. The solid line is the latent tangent space of $p_i$. The dashed line is the patch radius.
  \label{fig:2DPatch}}
\end{figure}

\begin{figure}[thb]
  \centering
  \includegraphics[width=0.5\textwidth]{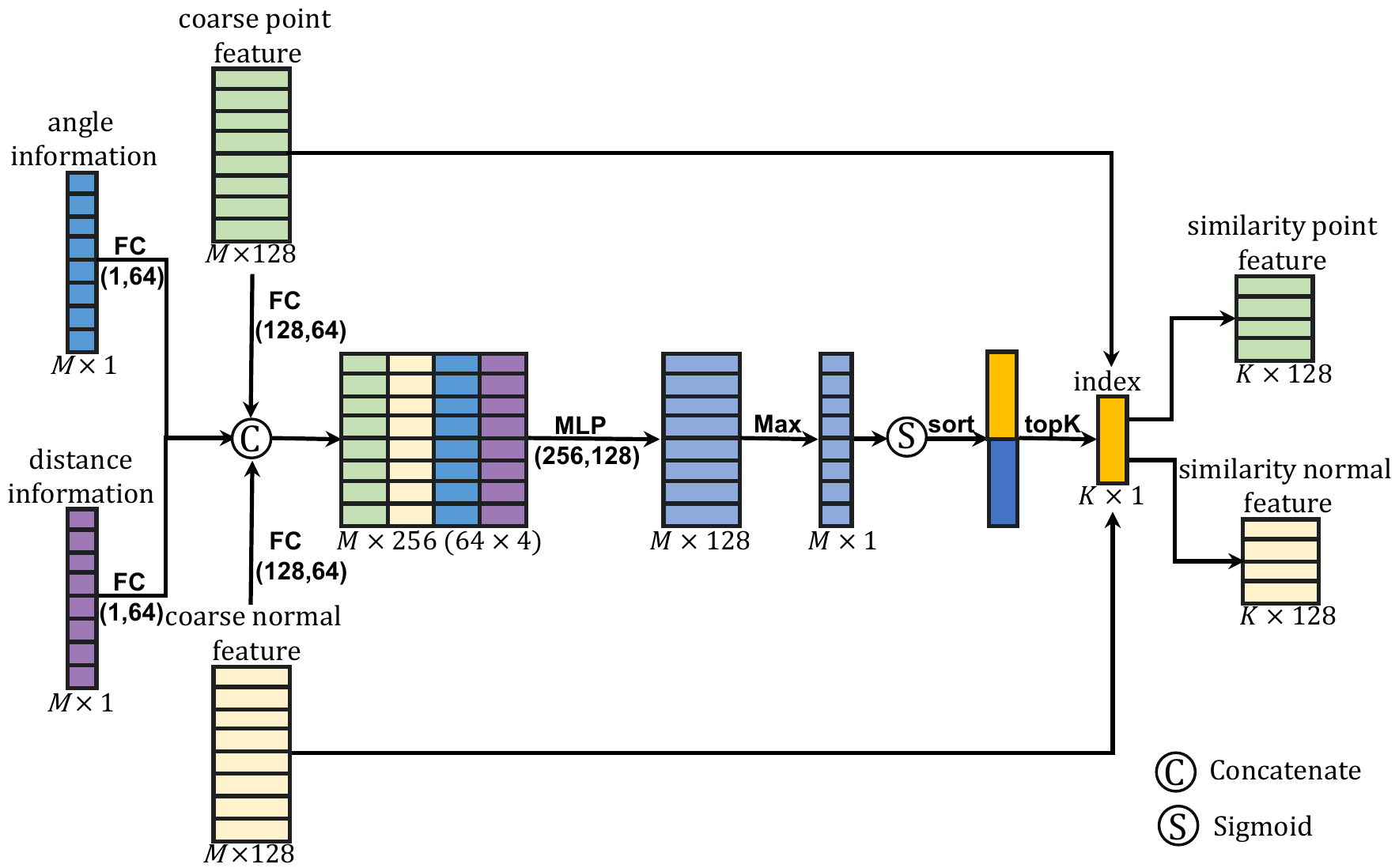}
  \caption{
  %\color{red}
  {Illustration of the shape-aware selector.}
  \label{fig:simliaritySelectionModule}}
\end{figure}

\subsubsection{Shape-aware selector}
The extracted coarse representations $\mathrm{F}_{\mathrm{P}_i}$ and $\mathrm{F}_{\mathrm{N}_i}$ for point $p_i$ can roughly describe the overall shape of patch $\mathrm{P}_i$. However, outliers, different scales and types of noise, and sampling anisotropy may be contained in the noisy patch. These defects of the patch inevitably hurt network performance and lead to unsatisfactory denoising results.
For example, suppose the patch of $p_i$ contains outliers and heavy-noise points; see Fig. \ref{fig:2DPatch-a}. In this case, these points negatively influence the representations of the patch, which may degrade denoising results, causing problems such as residual noise or shape collapsing. Furthermore, if the patch contains sharp features, the neighbors outside the same tangent space of $p_i$ have negative influences on feature representations of the patch, which may blur geometric features in the denoising results; see Fig. \ref{fig:2DPatch-b}.

In order to address the above problems, we design a shape-aware selector to choose points  similar to the specific point within the patch, which can greatly reduce the influences of negative points.
Fig. \ref{fig:simliaritySelectionModule} shows the detailed structure of the shape-aware selector. As we can see, the selector leverages the guidance of four types of information, including the coarse point and normal features learned by our feature extractor and two geometry information (distance and angle information). Specifically, we first apply fully connected layers to extract the feature from the four types of information, respectively. Then, we calculate the similarity score between  point $p_j$ and $p_i$ using the following score function:
\begin{equation} \label{eq:scoreFunction}
	\begin{split}
		  w_{p_j} = \mathrm{score}\bigl(\mathrm{concat}(&\mathrm{FC}(ang_j), \mathrm{FC}(dist_j),
		\mathrm{FC}(f_{p_j}), \\
		&\mathrm{FC}(f_{n_j}))\bigr), \ \ \forall p_j \in \mathrm{P}_i, \forall n_j \in \mathrm{N}_i,
	\end{split}
\end{equation}
where $\mathrm{FC}$ denotes fully connected layers.
$ang_j$ and $dist_j$ represent angle and distance information of $p_j$ defined as
\begin{equation*}
  ang_j = \angle(p_j - p_i, n_j), \  dist_j = \mathrm{exp}(-\|p_j - p_i \|^2).
\end{equation*}
The implementation of score function \eqref{eq:scoreFunction} is shown in Fig.  \ref{fig:simliaritySelectionModule}. It produces a similarity vector $W_{\mathrm{P}_i}=\{w_{p_j}\} \in \mathbb{R}^{M \times 1}$, recording the degree of similarity between point $p_j$ and $p_i$.
Then, the points within the patch that have top-K scores are retained, while the others are discarded as follows:
\begin{equation*}\label{eq:topk}
  \mathrm{P}^K_i = \{p_l | l \in \mathrm{topK}(W_{\mathrm{P}_i}) \} \in \mathbb{R}^{K \times 3},
\end{equation*}
where $\mathrm{topK}$  is the function that extracts the indices of the $K$ largest elements of the given input, and $\mathrm{P}^K_i$ is the similarity point set to the specific point.
As a result, we can obtain the similarity representation for point and normal vectors from the similarity point set within the patch as
\begin{equation*}\label{eq:similarFeatures}
  \mathrm{F}_{\mathrm{P}_i}^1 = \{f_{p_l} \}  \in  \mathbb{R}^{K \times 128}, \ \mathrm{F}_{\mathrm{N}_i}^1 = \{f_{n_l} \}  \in  \mathbb{R}^{K \times 128},
  %\forall p_l \!\in\! \mathrm{P}^K_i,
\end{equation*}
where $f_{p_l}$ and $f_{n_l}$ are pointwise coordinate and normal features of point $p_l \in \mathrm{P}^K_i$.
\begin{figure}[thb]
  \centering
  \includegraphics[width=1\linewidth]{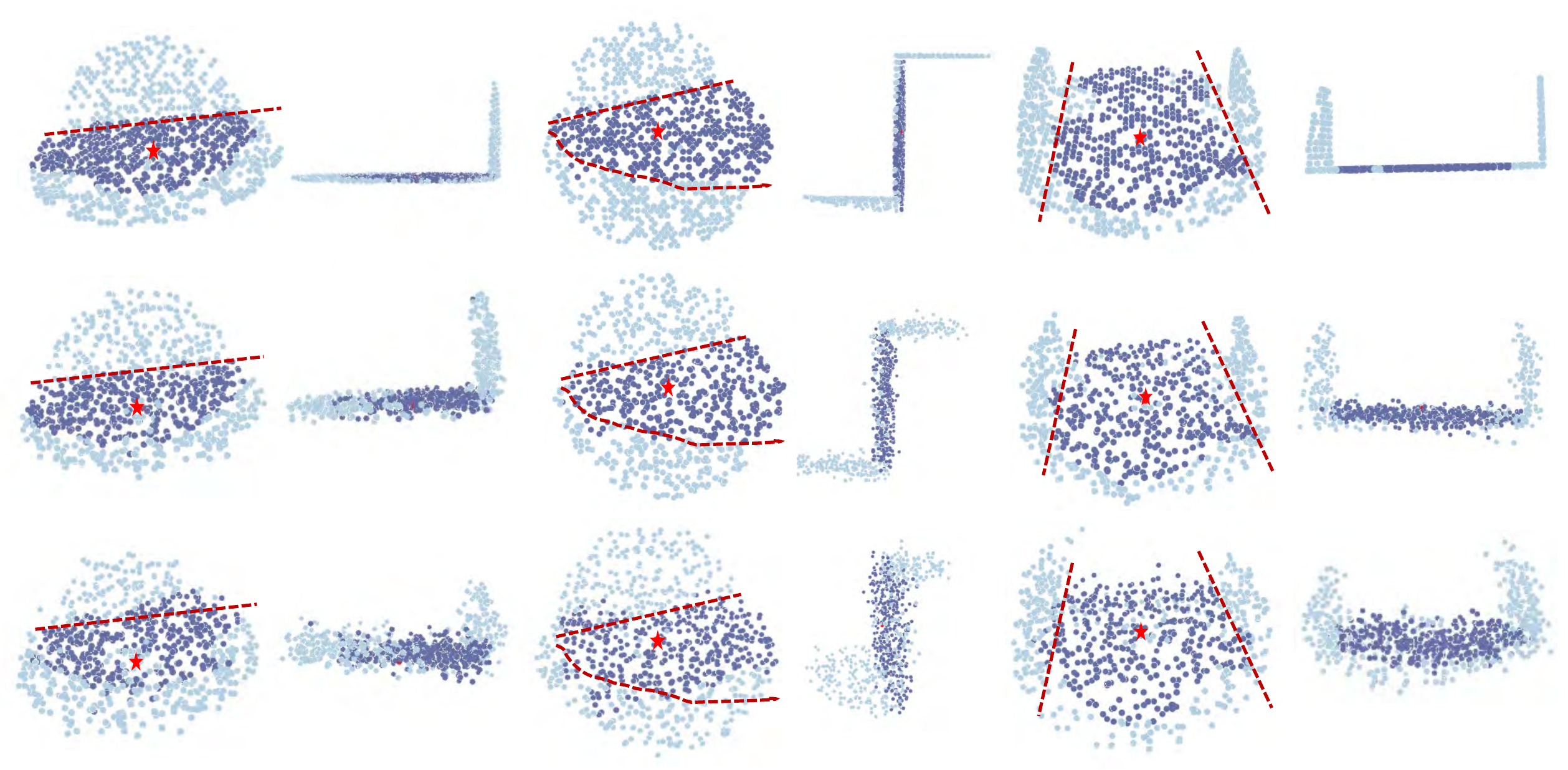}
  \caption{Illustration of  similar points selected  from three different shaped patches. From top to bottom: patches corrupted with 0\%, 0.25\%, 0.5\% noise, respectively.
  From left to right: for each patch, we show the front and side views of the patch to demonstrate the selected  points (plotted in dark blue) similar to the specific point (plotted in red). The negative points to the specific point are plotted in light blue.
  }
  \label{fig:PatchesAndPointsets}
\end{figure}
Fig. \ref{fig:PatchesAndPointsets} shows  examples of selecting similar points from different shapes of patches using our shape-aware selector. Consistent with our intuition, our selector is robust against noise and can choose similar points distributed on the tangent space of the specific point.

{\textbf{Remark 1}. We provide an intuitive interpretation of the shape-aware selector. To effectively preserve geometric features for denoising and normal filtering tasks, our selector chooses neighboring points that do not span geometric features (residing on the same tangent space). Our method relies solely on the prior information and feature-preserving loss function for the two related tasks to ensure the co-planarity of the selector, eliminating the need to label these co-planar points. In future work, we plan to investigate the use of the selector for weakly supervised plane segmentation.}

\begin{figure}[thb]
  \centering
  \includegraphics[width=0.5\textwidth]{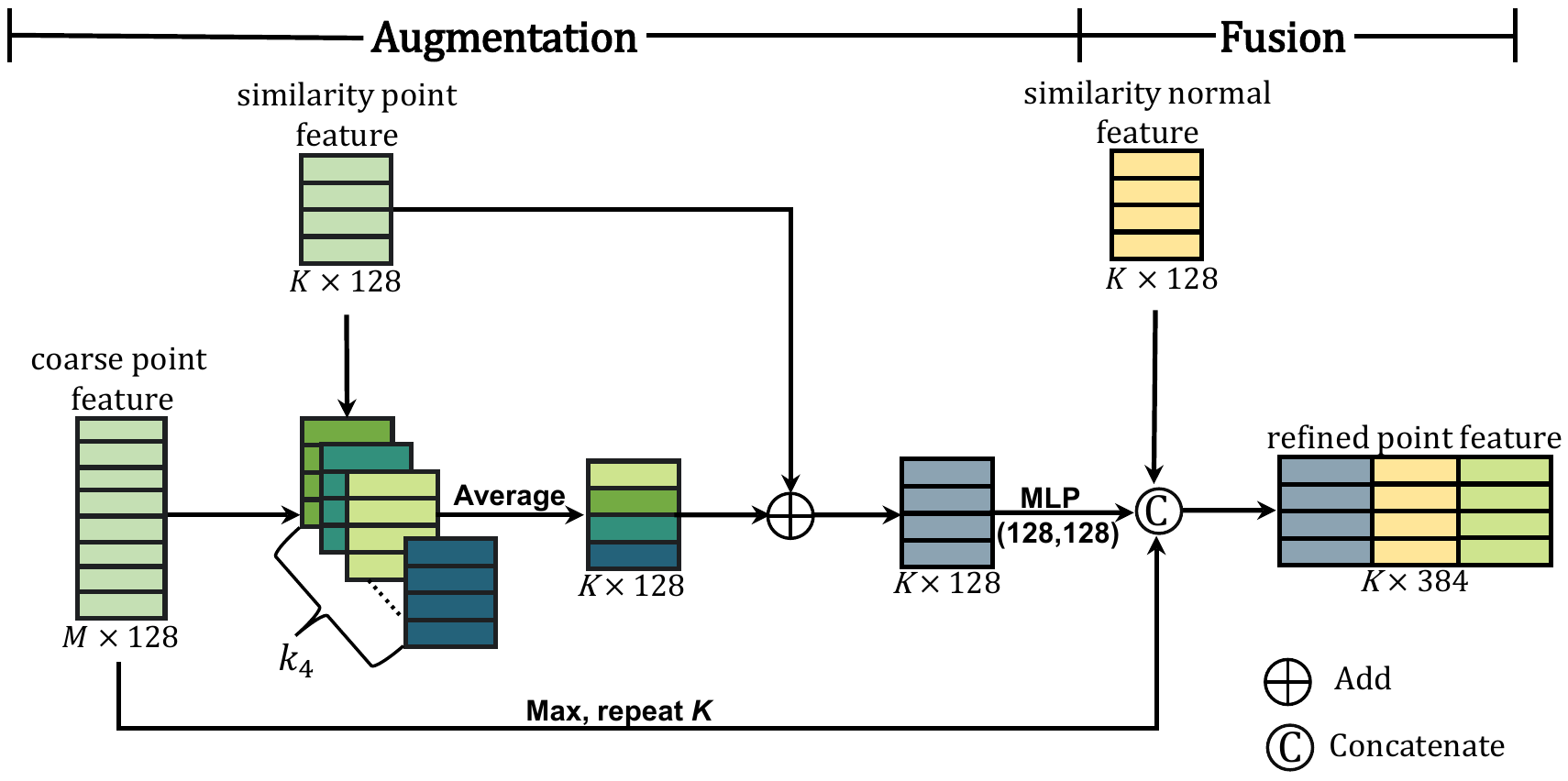}
  \caption{
  %\color{red}
  {Illustration of the feature refinement module consisting of feature augmentation and fusion units, using point cloud denoising network branch as an example.}
  \label{fig:augmentationAndFusionModule}}
\end{figure}

\subsubsection{Feature refinement} \label{subsec:featureRefinement}
Due to the unavoidable trade-off between noise removal and preservation of geometric features, we can remove the noise effectively but inevitably blur some small-scale geometric features through the previous two modules (the multiscale feature extractor and shape-aware selector). To address this issue, we propose a feature refinement module to facilitate the feature-preserving ability of the overall network. Fig. \ref{fig:augmentationAndFusionModule} shows the proposed module consisting of two units: feature augmentation and feature fusion, which will be detailed in the following.

\begin{figure}[thb]
	\centering
	\subfloat[]{\label{fig:PointAug-a}\includegraphics[width=0.25\textwidth]{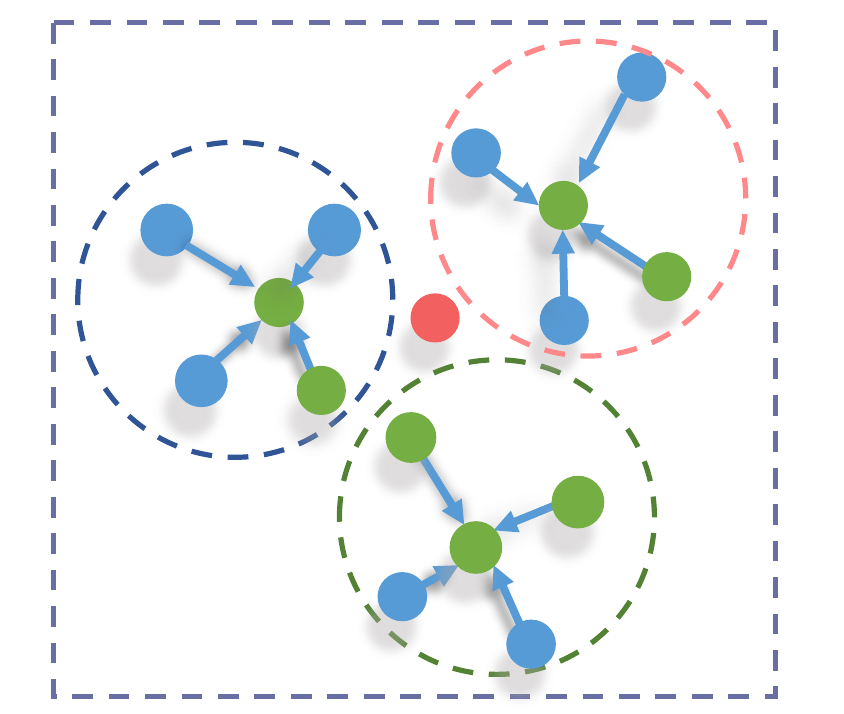}}
	\subfloat[]{\label{fig:PointAug-a}\includegraphics[width=0.25\textwidth]{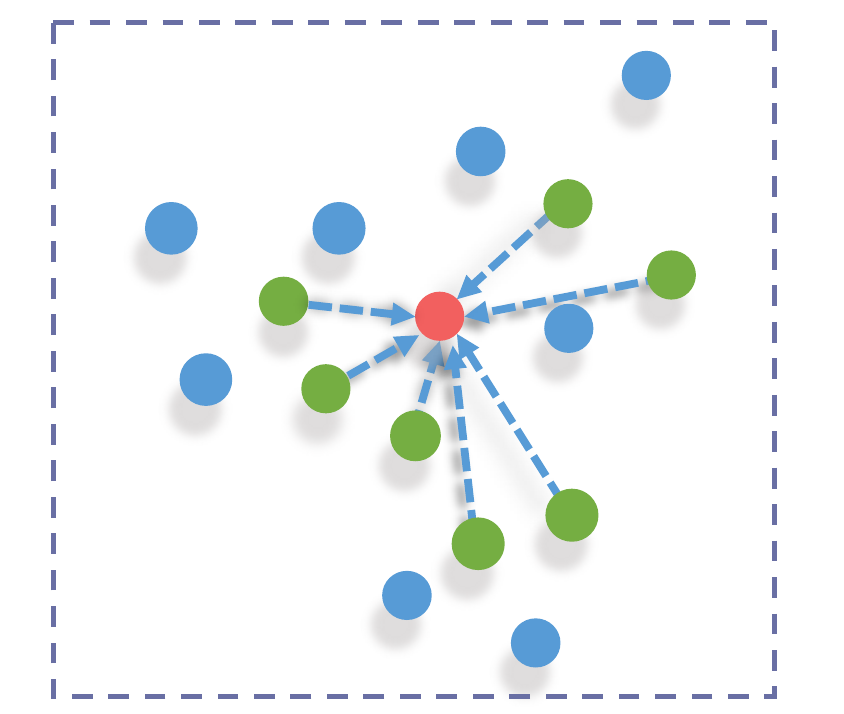}}
	\caption{(a) Illustration of the feature augmentation unit, which enlarges receptive fields (dotted circles) of similar points plotted in green. Thus, the specific point (plotted in red) can obtain a more representative feature from those augmented features of similar points, illustrated in (b).}
    \label{fig:PointAug}
\end{figure}

\begin{figure}[thb]
	\centering
	\subfloat[]{\label{fig:featureAugmentation-a}\includegraphics[width=0.16\textwidth]{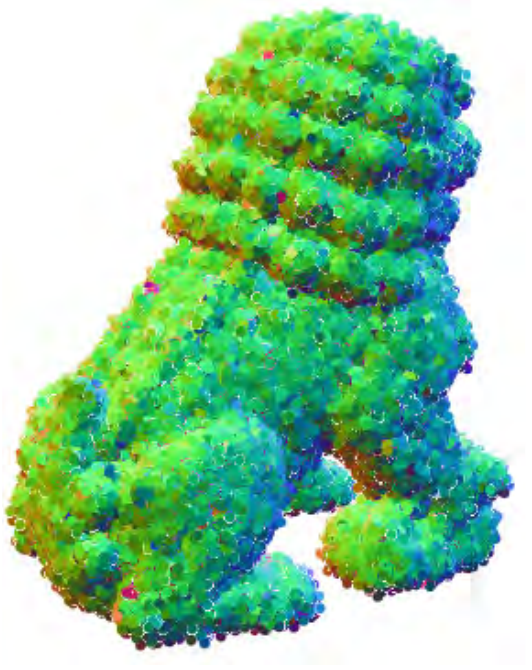}}
	\subfloat[]{\label{fig:featureAugmentation-b}\includegraphics[width=0.16\textwidth]{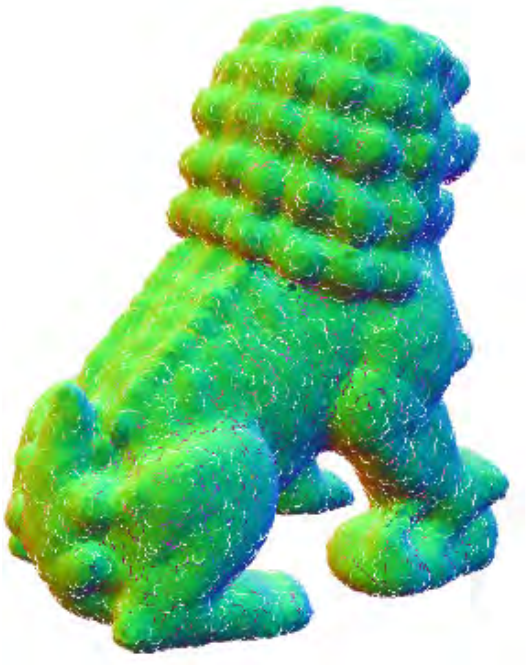}}
	\subfloat[]{\label{fig:featureAugmentation-c}\includegraphics[width=0.16\textwidth]{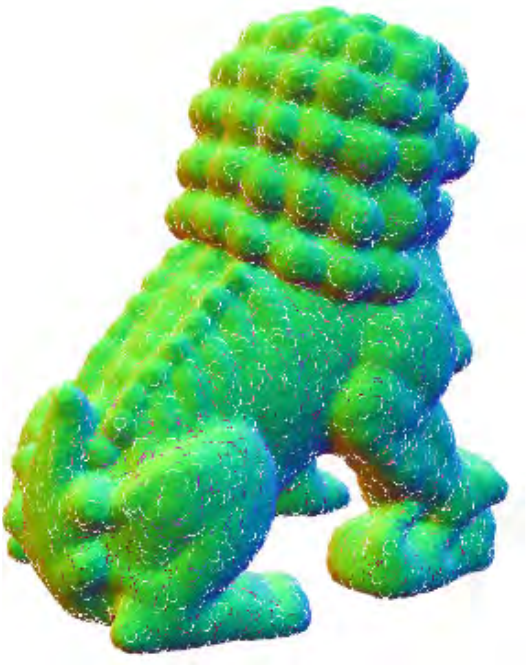}}
	\caption{(a) Noisy input. (b) Denoising result without the feature augmentation unit. (c) Denoising result with the feature augmentation unit. \label{fig:featureAugmentation}}
\end{figure}

\noindent \textbf{Feature augmentation.} After the stage of similar point selection, we sequentially augment the learned features of similar points by enlarging the receptive fields for discovering more locally geometric details; see Fig. \ref{fig:PointAug} for illustration.
To learn more representative features, for each similar point (to the specific point within the patch), we first search KNN neighbors of it and gather the features of the neighbors according to the similarity scores. Then, the augmented point and normal features of each similar point are computed as
\begin{equation*}
	\begin{aligned}
	   f^2_{p_l} & = \mathrm{MLP}\left(f_{p_l}+ \biggl(\frac{1}{k_4} \sum\limits_{{q} \in \mathcal{N}_{k_4}(p_l)} w_{q} f_{q} \biggr)\right),\\
       f^2_{n_l} & = \mathrm{MLP}\left( f_{n_l} + \biggl(\frac{1}{k_4} \sum\limits_{{q} \in \mathcal{N}_{k_4}(p_l)} w_{q} f_{n_q} \biggr)\right),
	\end{aligned}
\end{equation*}
where $k_4$ is the neighboring number of $p_l$, and $w_q$ is the similarity score obtained with \eqref{eq:scoreFunction}.
Fig. \ref{fig:featureAugmentation} shows the capability of the feature augmentation unit.
Fig. \ref{fig:featureAugmentation-c} shows that the proposed unit plays a key role in recovering local geometric features. Without this  unit, some geometric details and shallow structures are blurred in the result; see Fig. \ref{fig:featureAugmentation-b}.

\begin{figure}[thb]
  \centering
  \subfloat[]{\label{fig:featureFusion-a}\includegraphics[width=0.16\textwidth]{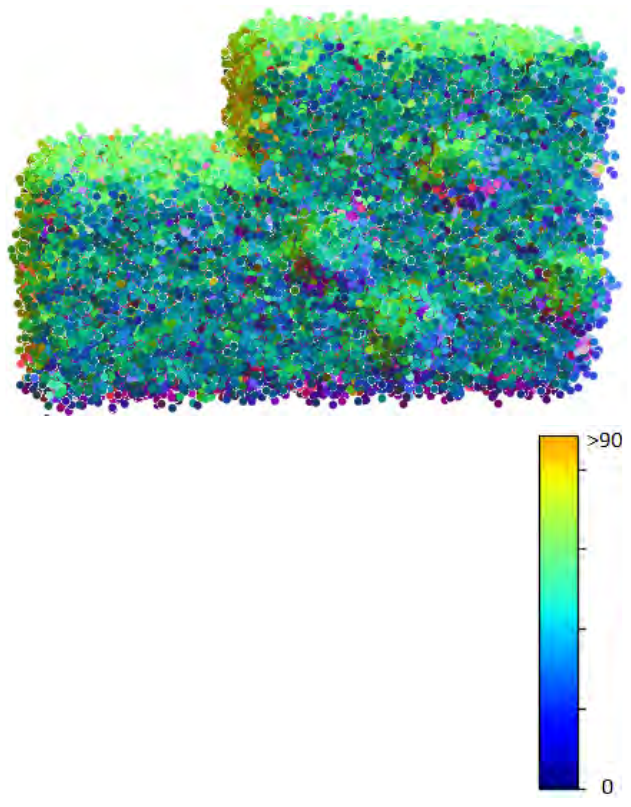}}
  \subfloat[]{\label{fig:featureFusion-b}\includegraphics[width=0.16\textwidth]{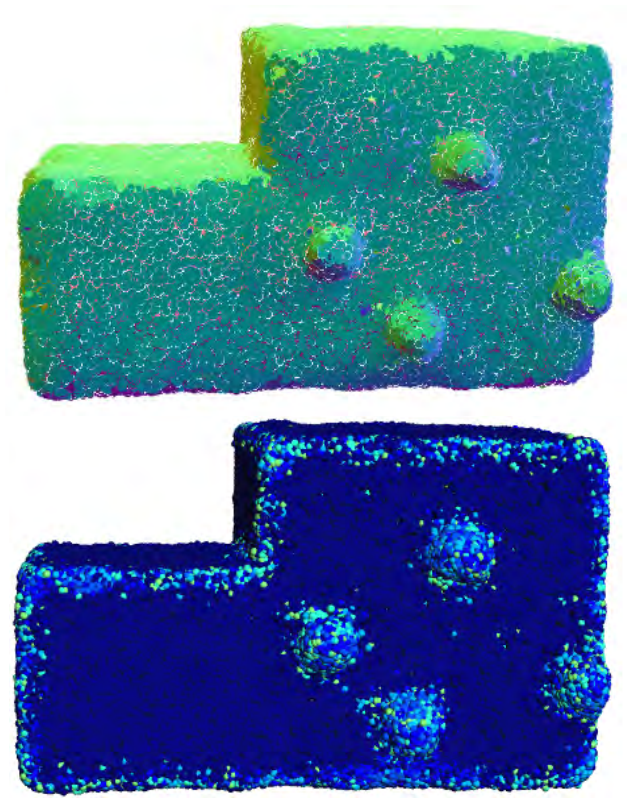}}
  \subfloat[]{\label{fig:featureFusion-c}\includegraphics[width=0.16\textwidth]{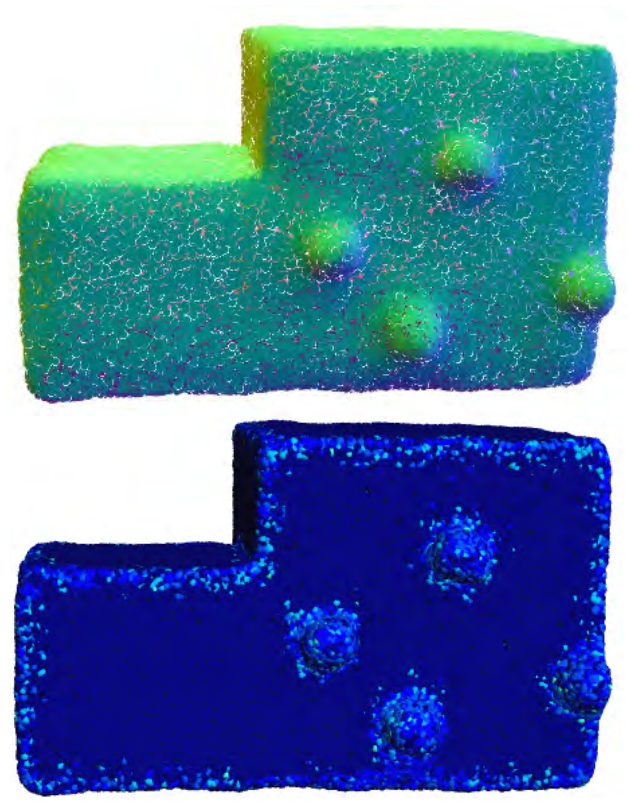}}
  \caption{Illustration of the feature fusion unit. (a) Noisy input. (b) Denoising and normal filtering results without the feature fusion unit. (c) Denoising and normal filtering results with the feature fusion unit. The second row visualizes the normal error maps, measured as the angular differences between the filtered normals and the ground truth.  \label{fig:featureFusion}}
\end{figure}

\noindent \textbf{Feature fusion.} The feature fusion unit is motivated by the following observation.
The point features are more instrumental in recovering local geometric features, while the normal features are more suitable for recovering sharp edges, corners, and smooth transition regions. Thus, the fusion of point and normal features can help better recover various types of geometric features.
In addition, the global feature can better describe the whole shape of the patch. Specifically, for each similar point within the patch, we concatenate the three types of features together (including point and normal features and the global feature) to obtain the fused features as
\begin{equation*}
	\begin{aligned}
	   f^3_{p_l} =\mathrm{concat}( f^2_{p_l},f^2_{n_l},f^g_{\mathrm{P}_i}),\\
       f^3_{n_l} =\mathrm{concat}( f^2_{n_l}, f^2_{p_l},f^g_{\mathrm{N}_i}),
	\end{aligned}
\end{equation*}
where $f^3_{p_l}$, $f^3_{n_l}$ are the refined point and normal features of the similar point.
$f^g_{\mathrm{P}_i}$ and $f^g_{\mathrm{N}_i}$ are global features of the patch which are obtained by feeding the coarse features $\mathrm{F}_{\mathrm{p}_i}$ and $\mathrm{F}_{\mathrm{N}_i}$ into the max-pooling operator, respectively.
We validate the effectiveness of our feature fusion unit in Fig. \ref{fig:featureFusion}.
As Figs. \ref{fig:featureFusion-b} and \ref{fig:featureFusion-c} show,
incorporating normal features into point features allows the denoising task to better preserve sharp geometric features and smooth regions while incorporating point features into normal features can yield a more satisfactory normal filtering result, see the normal error maps in the second row of Figs. \ref{fig:featureFusion-b} and \ref{fig:featureFusion-c}.

\subsubsection{Decoder}
Given a point $p_i$, the feature refinement module generates refined point and normal features of the patch $\mathrm{P}_i$, $\mathrm{F}_{\mathrm{P}_i}^3 \!=\! \{f^3_{p_l} \} \!\in\! \mathbb{R}^{{K} \times 384}$ and $\mathrm{F}_{\mathrm{N}_i}^3 \!=\! \{f^3_{n_l} \} \!\in\! \mathbb{R}^{{K} \times 384}$, which are inputs to our decoder.
Our decoder includes two regressors: coordinate and normal regression.
For coordinate regression, we apply MLP and FC layers to predict the displacement vector from the refined point feature $\mathrm{F}_{\mathrm{P}_i}^3$, and then obtain the denoised point by adding the predicted displacement vector to the original coordinate; see Fig. \ref{fig:pointregression}.
For normal regression, we use ResNet-like operations \cite{he2016deep} to obtain the filtered normal from the refined normal feature $\mathrm{F}_{\mathrm{N}_i}^3$; see the details in Fig. \ref{fig:normalregression}.

\subsection{Training Losses}
To train our network in an end-to-end manner, we design three types of losses as our optimization objectives: a point-denoise loss, a normal-filter loss, and an orthogonality loss.
The joint loss function is formulated as
\begin{equation}
\mathcal{L}=\lambda_{1} \mathcal{L}_\text{point} + \lambda _{2} \mathcal{L}_\text{normal} + \lambda_{3}\mathcal{L}_\text{ortho},
\label{eq16}
\end{equation}
where $\lambda_{1}$, $\lambda_{2}$, and $\lambda_{3}$ are three hyperparameters that balance the importance of each term. We empirically set $\lambda_{1}=100$, $\lambda_{2}=10$ and $\lambda_{3}=10$ for training.

\noindent \textbf{Point-denoise loss.} To ensure that the denoising result can approximate the underlying surface while preserving sharp features well, we apply a bilateral mechanism proposed in \cite{zhang2020pointfilter} to compute the project distance between the denoised point and its neighboring points within the ground-truth patch.
To further improve the distribution of the denoising result, we follow \cite{lipman2007parameterization,zhang2020pointfilter,chen2022repcd} and adopt a repulsive term to penalize those points that are too close to each other.
Thus, our two-term point denoising loss is defined as follows:
\begin{equation*}
\begin{aligned}
\mathcal{L}_{\text {point}} &=\alpha \frac{\sum_{\bar{p}_{j} \in \overline{\mathrm{P}}_{i}}\left|\left(\hat{p}_{i}-\bar{p}_{j}\right) \cdot \bar{n}_{j}\right| \cdot \phi\left(\left\|\hat{p}_{i}-\bar{p}_{j}\right\|\right) \theta\left(\bar{n}_{i}, \bar{n}_{j}\right)}{\sum_{\bar{p}_{j} \in \overline{\mathrm{P}}_{i}} \phi\left(\left\|\hat{p}_{i}-\bar{p}_{j}\right\|\right) \theta\left(\bar{n}_{i}, \bar{n}_{j}\right)} \\
&+(1-\alpha) \max _{\bar{p}_{j} \in \overline{\mathrm{P}}_{i}}\left\|\hat{p}_{i}-\bar{p}_{j}\right\|,
\end{aligned}
\end{equation*}
where $\alpha$ is a parameter that balances the denoising and uniform distribution terms. $\alpha$ is set to 0.97 referring to \cite{zhang2020pointfilter}.
$\mathrm{\overline{P}}_{i}$ is the ground-truth patch centered at the denoised point $\hat{p}_{i}$, ${\bar{n}_{i}}$ is the ground-truth normal of the clean point which is nearest to the denoised point $\hat{p}_{i}$, ${\bar{n}_{j}}$ is the ground-truth normal of $\bar{p}_{j}$.
$\phi$ and $\theta$ are two monotonically decreasing functions in terms of distance and normal deviation.
$\phi$ is the Gaussian function, and $\theta$ is given as $\theta(\bar{n}_i, \bar{n}_j) = \exp(-\frac{1-\bar{n}_i\cdot \bar{n}_j}{1-\cos(15^{\circ})})$.

\noindent \textbf{Normal-filter loss.} For the denoised point $\hat{p}_{i}$, we simply use the Euclidean distance between its filtered normal $\hat{n}_i$ and the ground truth $\bar{n}_{i}$ as our normal-filter loss:
\begin{equation*}
\mathcal{L}_{\text{normal}}= {\left\| \hat{n}_{i}-\bar{n}_{i} \right\|}^{2}.
\end{equation*}
\noindent\textbf{Orthogonality loss.} To encourage the denoised point to move in the direction of its filtered normal while ensuring the orthogonality between the filtered normal and the edges connecting the denoised point and its neighboring points, we propose the following orthogonality loss that can constrain the denoising and normal filtering branches at the same time:
\begin{equation*}
{\mathcal{L}_\text{ortho}}=\sum_{\bar{p}_{j} \in \overline{\mathrm{P}}_{i} } {\left( w_{p_j} \left|(\hat{p}_{i} -\bar{p}_{j})\cdot \hat{n}_{i} \right| \right)}^{2}.
\end{equation*}

%\textcolor{red}
{\textbf{Remark 2}. Our method employs several mechanisms and modules, including multitasking and a shape-aware selecting module, to achieve high-quality results. 
Even in the presence of high levels of noise, our method can progressively improve the filtered normal (used in the orthogonality loss), and make it to be consistent with the ground-truth normal (used in the point-denoise loss). Our multitask network can be iteratively performed to enhance the results (denoising and normal filtering) gradually. The selecting module reduces the negative impact of large noise, resulting in high-quality results. Additionally, our normal-filter loss optimizes the filtered normal to closely match the ground-truth normal.}

\section{Experiments and Discussions} \label{sec:experiments}
We evaluate our method for denoising and normal filtering tasks visually and numerically and show its superiority compared to the SOTA denoising and normal filtering methods. We further modify our method and conduct ablation studies to validate the effectiveness of each design choice made in our method.

\subsection{Experimental Settings} \label{subsec:experimentalSettings}
\textbf{Dataset.} To train our end-to-end network for denoising and normal filtering tasks, we adopt the dataset provided by \cite{zhang2020pointfilter}.
The training set contains 11 CAD and 11 non-CAD models (clean data without noise).
The ground truth point clouds with normal information can be obtained by randomly sampling the clean models.
The number of sample points for each model is set uniformly as 100K. For these ground truth point clouds, we corrupt each of them by Gaussian noise with standard deviations of 0.25\%, 0.5\%, 1\%, 1.5\%, and 2.5\% to the diameter of the bounding box.
Thus, the final training set contains 110 noisy point clouds (normals are estimated with PCA) and the corresponding 22 ground truth point clouds (with normals).

To conduct comparisons more effectively for both tasks, we construct a test set consisting of synthetic point clouds and raw scanned data.
For synthetic data, we rely on the
synthetic dataset released in \cite{zhang2020pointfilter} including three categories:
simple, medium, and complicated, in which there are 7, 6,
and 7 clean models, respectively. The raw scanned data will be introduced in the following experimental section.
Again, the number of sampled points for each model is unified to 100K.
Each of the clean point clouds is perturbed by Gaussian noise with standard deviations of 0.25\%, 0.5\%, 1\%, 1.5\%, and 2.5\% to the diagonal of the bounding box.

\textbf{Network inference.} Our multitask network excels at producing feature-preserving denoising results and satisfactory normal filtering results simultaneously. Additionally, our method can be iteratively applied to further refine the results when the noise level is high. In this process, the denoised points and filtered normals from previous iterations can serve as inputs to our method for further refinement. It is important to note that this iterative process is only performed during the inference phase.

\textbf{Implementation details.} In our implementation, the patch radius $r$ is set to 5\% of the diagonal length of the bounding box by default.
We also set the number of points within each patch as $M=512$.
If the patch has insufficient points ($<512$), we pad the origin, and if it has sufficient points ($>512$), we randomly select a subset.
Our feature extractors employ three scales, two based on Euclidean distance ($k_1=8,k_2=16$) and one in feature space ($k_3=16$).
For the $\mathrm{topK}$ strategy of the shape-aware selector, we set $K=\frac{1}{2}M=256$.
The neighborhood size in the feature augmentation unit is set to $k_4 = 10$.
Our network is implemented in PyTorch and trained on a single NVIDIA GTX 2080Ti GPU for 45 epochs using the SGD optimizer. The learning rate decreased from 1e-4 to 1e-8.
Upon publication, we will release our source code and pre-trained models on GitHub.

\begin{figure*}[htb]
	\centering
	\subfloat[Noisy]{\label{fig:fandisk-a}\includegraphics[width=0.124\textwidth]{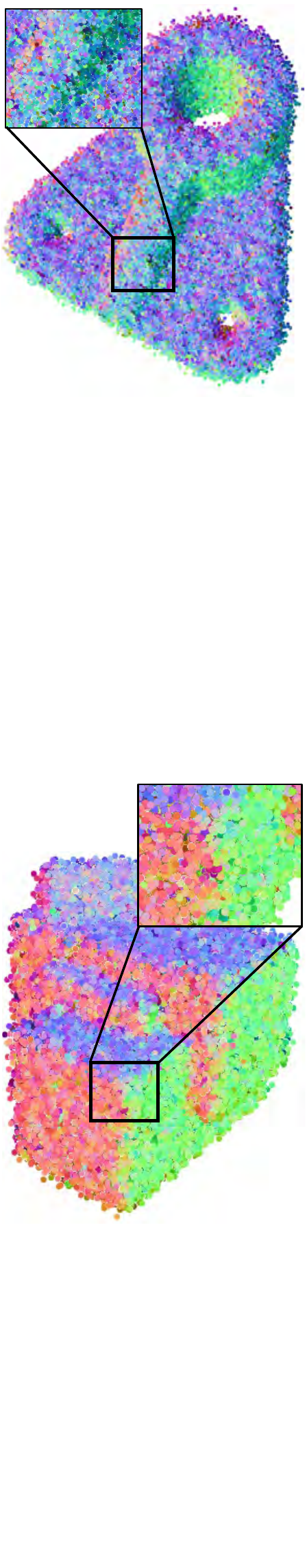}}
	\subfloat[WLOP]{\label{fig:fandisk-b}\includegraphics[width=0.124\textwidth]{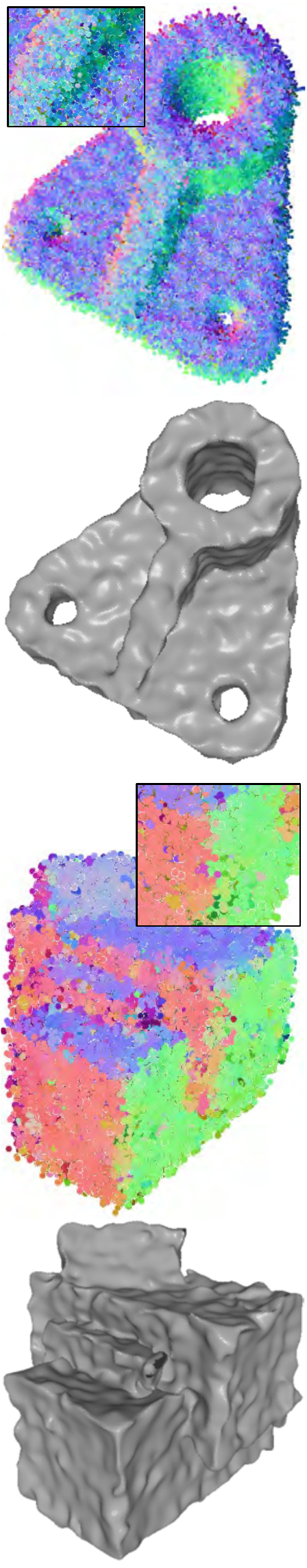}}
	\subfloat[RIMLS]{\label{fig:fandisk-c}\includegraphics[width=0.124\textwidth]{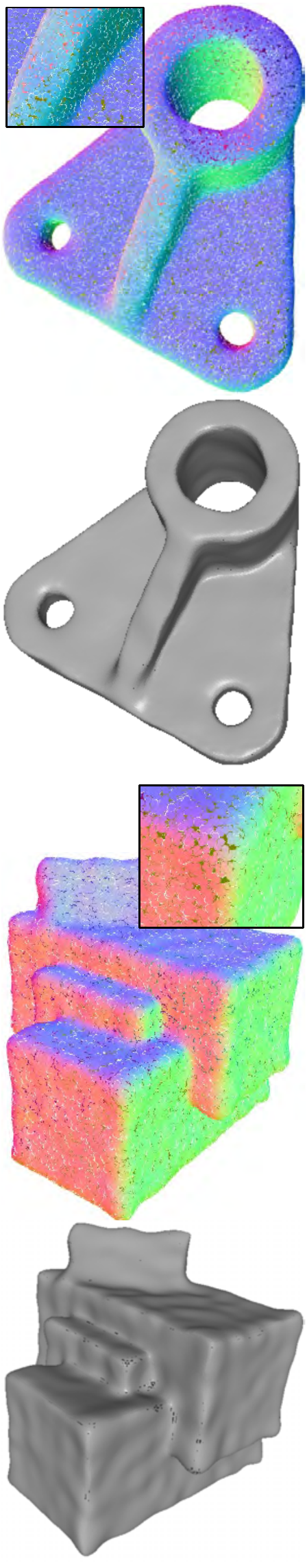}}
	\subfloat[EC-Net]{\label{fig:fandisk-d}\includegraphics[width=0.124\textwidth]{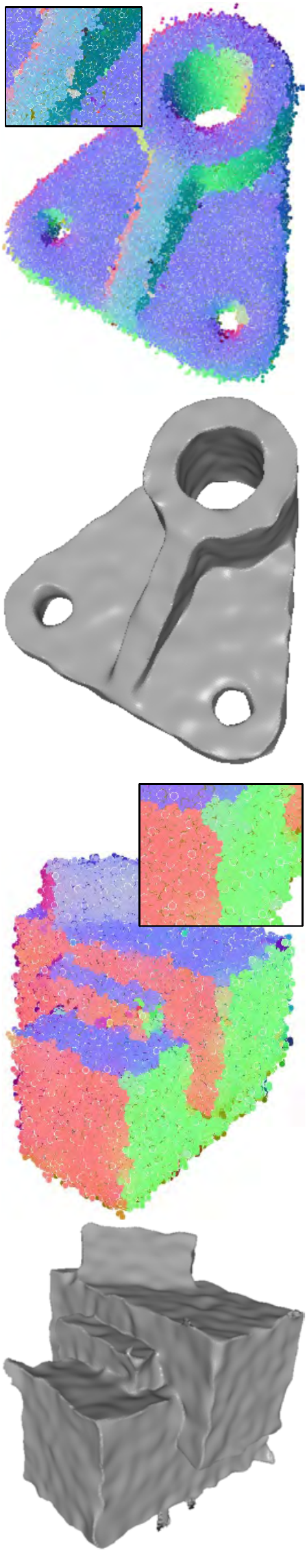}}
	\subfloat[DMR]{\label{fig:fandisk-e}\includegraphics[width=0.124\textwidth]{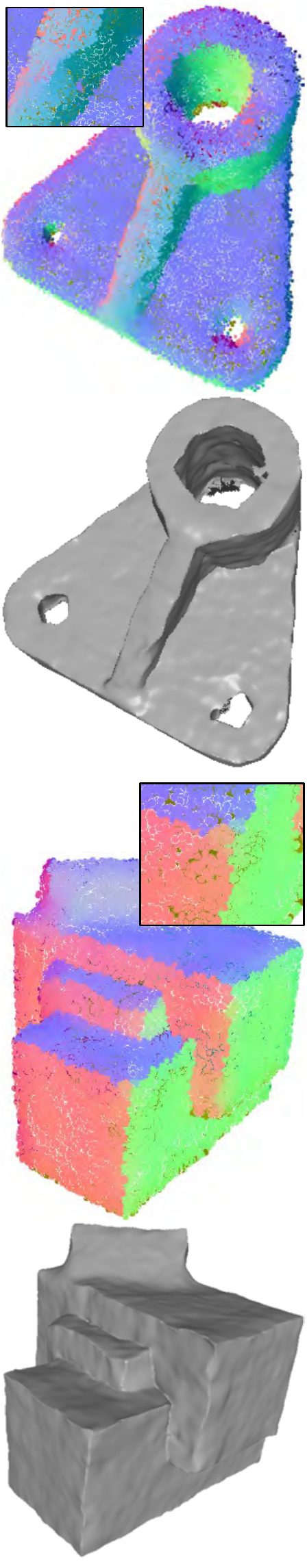}}
	\subfloat[PCN]{\label{fig:fandisk-f}\includegraphics[width=0.124\textwidth]{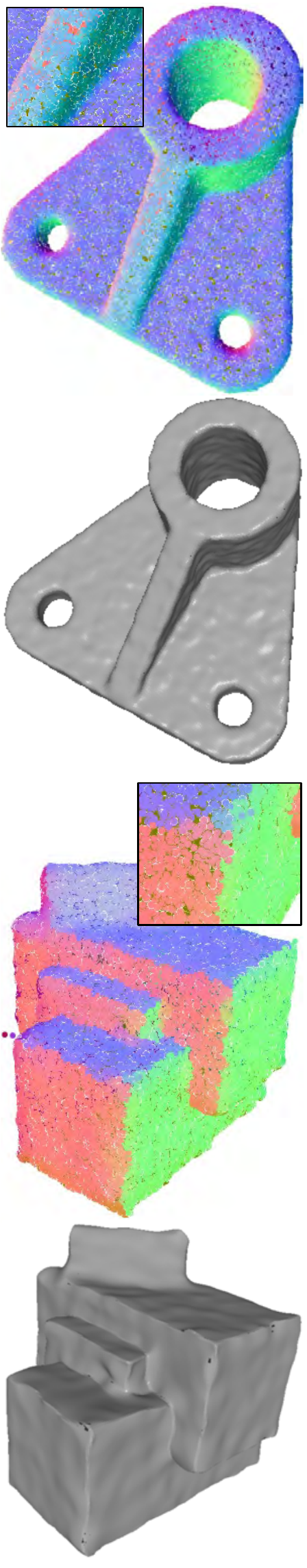}}
	\subfloat[PF]{\label{fig:fandisk-g}\includegraphics[width=0.124\textwidth]{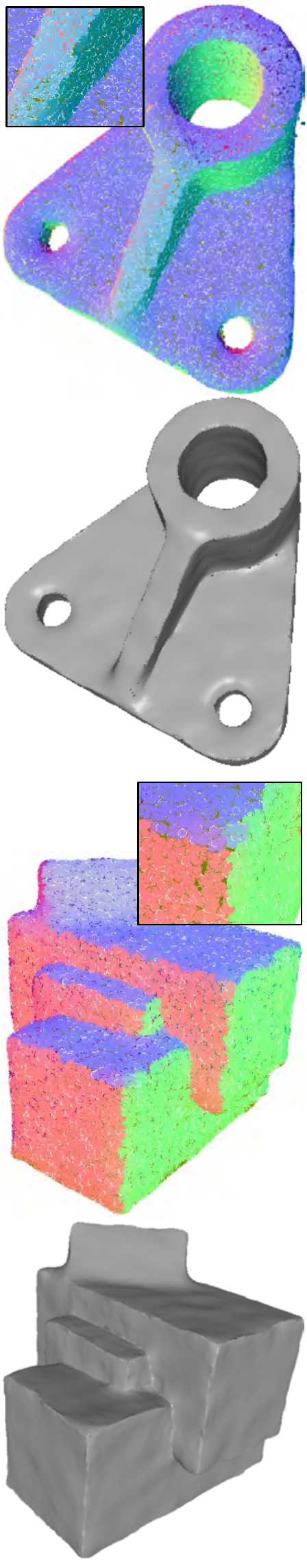}}
	\subfloat[Ours]{\label{fig:fandisk-h}\includegraphics[width=0.124\textwidth]{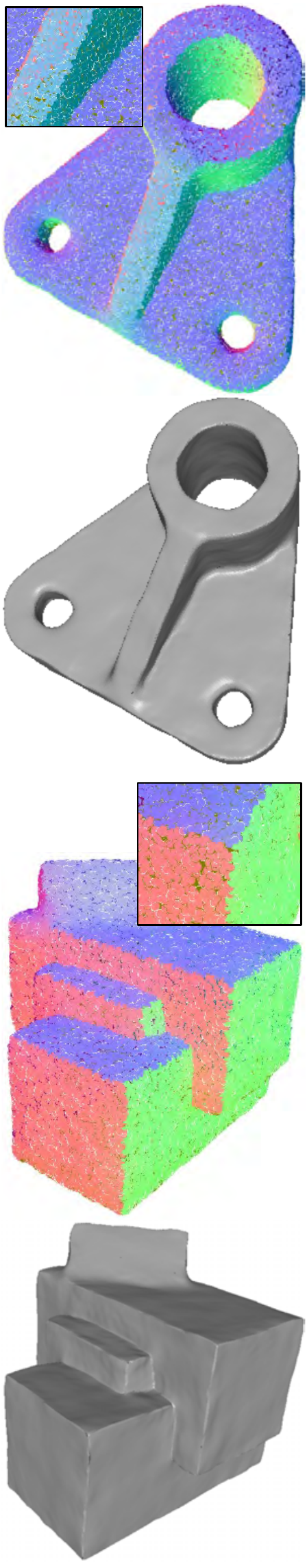}}
	\caption{ Denoising results of synthetic data with CAD models. From left to right: noisy input, results produced by WLOP, RIMLS, EC-Net, DMR, PCN, PF, and our method, respectively. The first and third rows show the denoised results with 1\% noise and their zoomed views. The second and fourth rows show the corresponding surface reconstruction results. Our method better removes noise and preserves sharp features.
	}
	\label{fig:fandisk}
\end{figure*}
\subsection{Experiments for Point Cloud Denoising}
We present visual and numerical comparisons between our method and SOTA denoising methods, including WLOP \cite{huang2009consolidation}, RIMLS \cite{oztireli2009feature}, EC-Net \cite{yu2018ec}, DMRDenoise (DMR) \cite{luo2020differentiable}, PointCleanNet (PCN) \cite{rakotosaona2020pointcleannet}, and Pointfilter (PF) \cite{zhang2020pointfilter}.
For WLOP and RIMLS, we perform the code provided by their authors and carefully tune the parameters to produce denoised results.
For DMR and PCN, we use the code released by their authors to retrain new models over our training set.
For EC-Net and PF, we use the pretrained models provided by the authors.
Sometimes, we reconstruct the denoised results produced by the tested methods for enhancing visual effects via RIMLS (provided by Meshlab for feature-preserving reconstruction).

\textbf{Synthetic data.} Fig. \ref{fig:fandisk} demonstrates comparisons of CAD surfaces including sharp features and smooth regions. The tested CAD surfaces are corrupted by a significant amount of noise. As depicted in Fig. \ref{fig:fandisk}, WLOP and EC-Net exhibit excessive noise in their results when the noise level is high. DMR is effective at removing the noise, but it distorts the overall shapes of the surfaces, as illustrated in Fig. \ref{fig:fandisk-e}.
While RIMLS and PCN recover smooth regions well, they blur sharp features to varying degrees, as shown in Figs. \ref{fig:fandisk-c} and \ref{fig:fandisk-f}.
Additionally, PCN may induce some artifacts in the smooth regions of the results at the tested noise level.
In contrast, our method and PF effectively preserve sharp features. Compared to PF, our method more accurately recovers sharp features, as shown in Figs. \ref{fig:fandisk-g} and \ref{fig:fandisk-h}. As a result, our method is superior to the compared methods in preserving sharp features and smooth regions simultaneously.

\begin{figure*}[htb]
	\centering
	\subfloat[Noisy]{\label{fig:elephant-a}\includegraphics[width=0.125\textwidth]{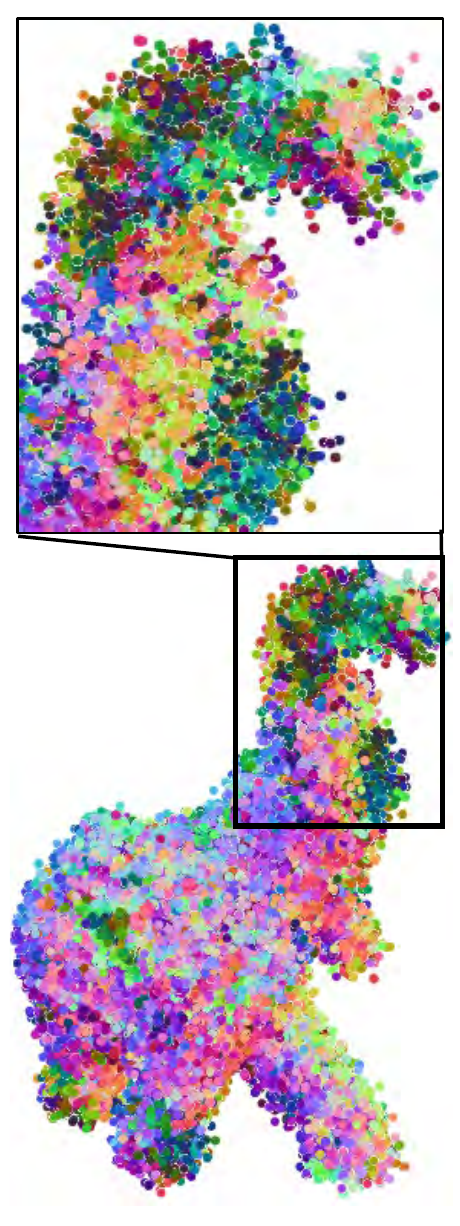}}
	\subfloat[WLOP]{\label{fig:elephant-b}\includegraphics[width=0.125\textwidth]{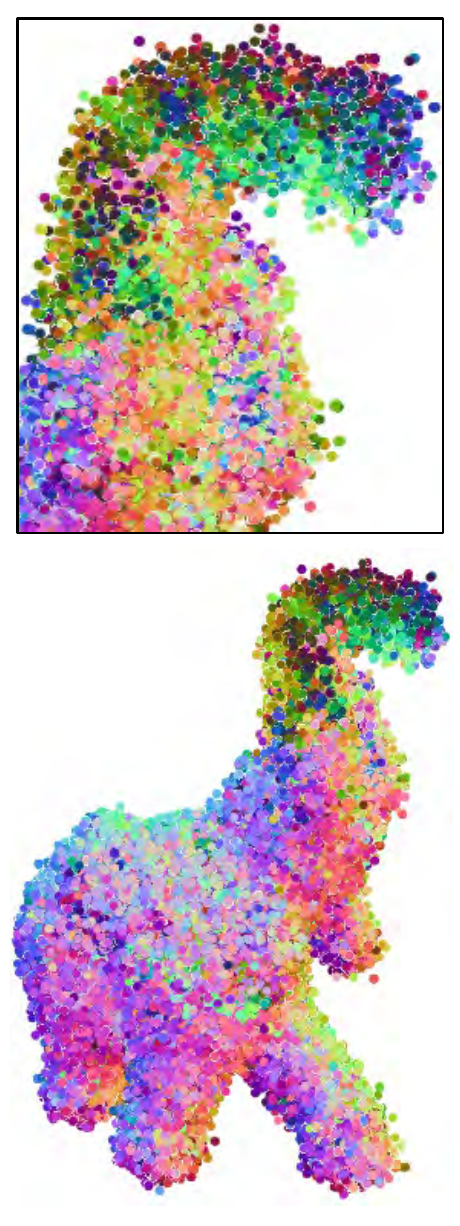}}
	\subfloat[RIMLS]{\label{fig:elephant-c}\includegraphics[width=0.125\textwidth]{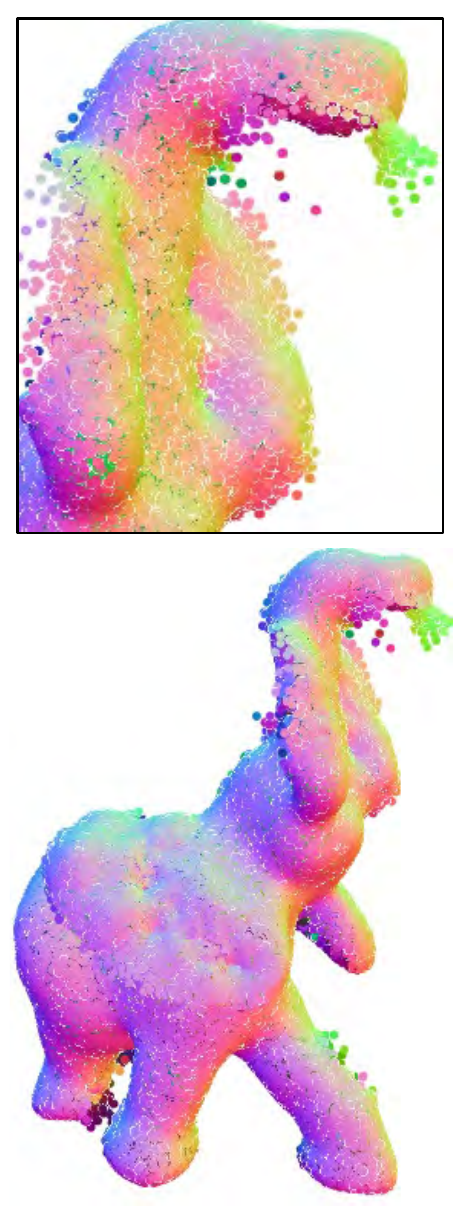}}
	\subfloat[EC-Net]{\label{fig:elephat-d}\includegraphics[width=0.125\textwidth]{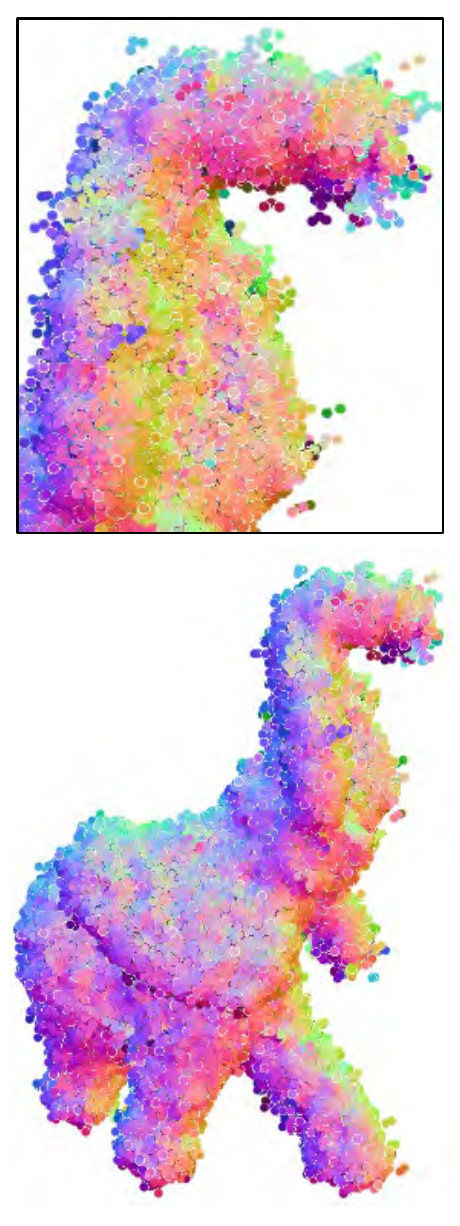}}
	\subfloat[DMR]{\label{fig:elephant-e}\includegraphics[width=0.125\textwidth]{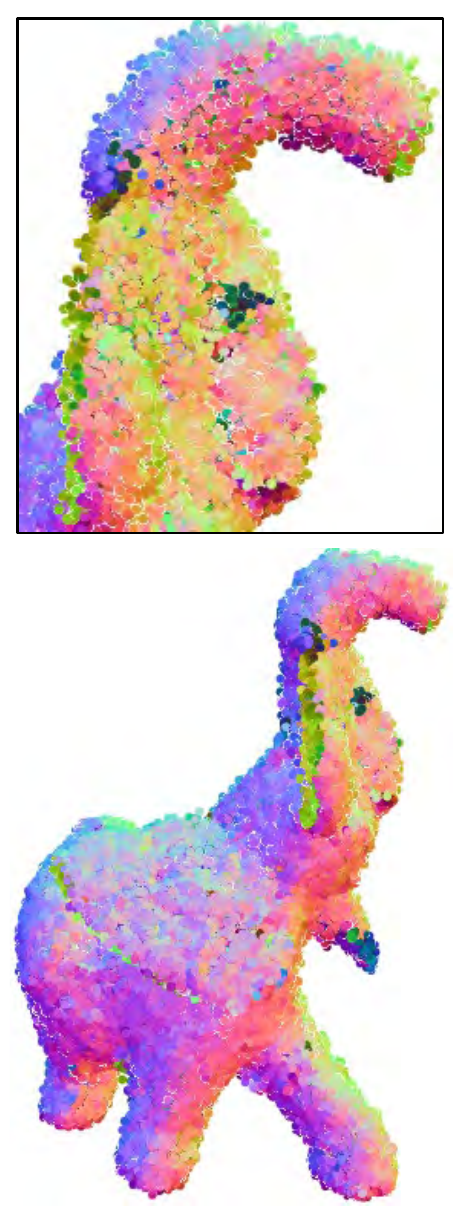}}
	\subfloat[PCN]{\label{fig:elephant-f}\includegraphics[width=0.125\textwidth]{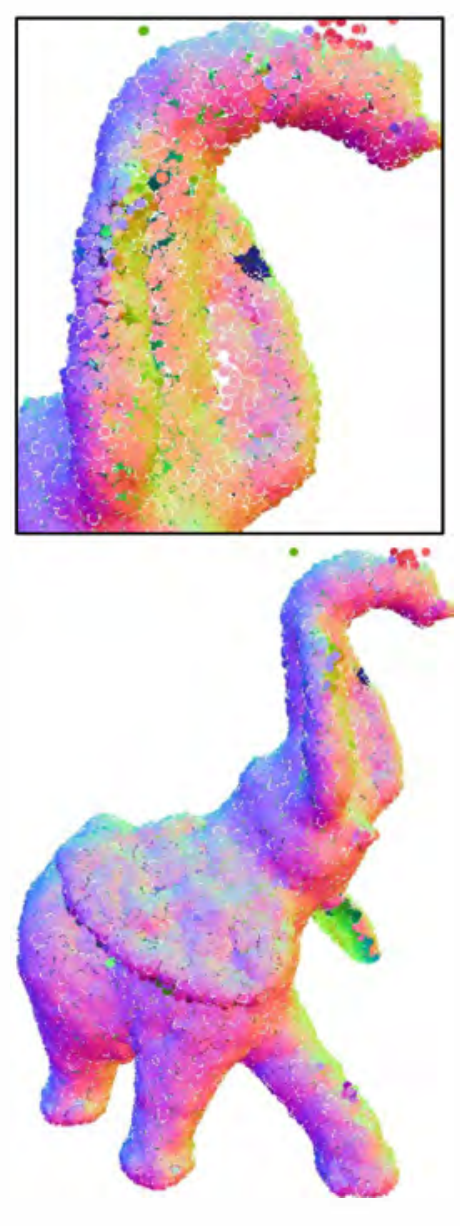}}
	\subfloat[PF]{\label{fig:elephant-g}\includegraphics[width=0.125\textwidth]{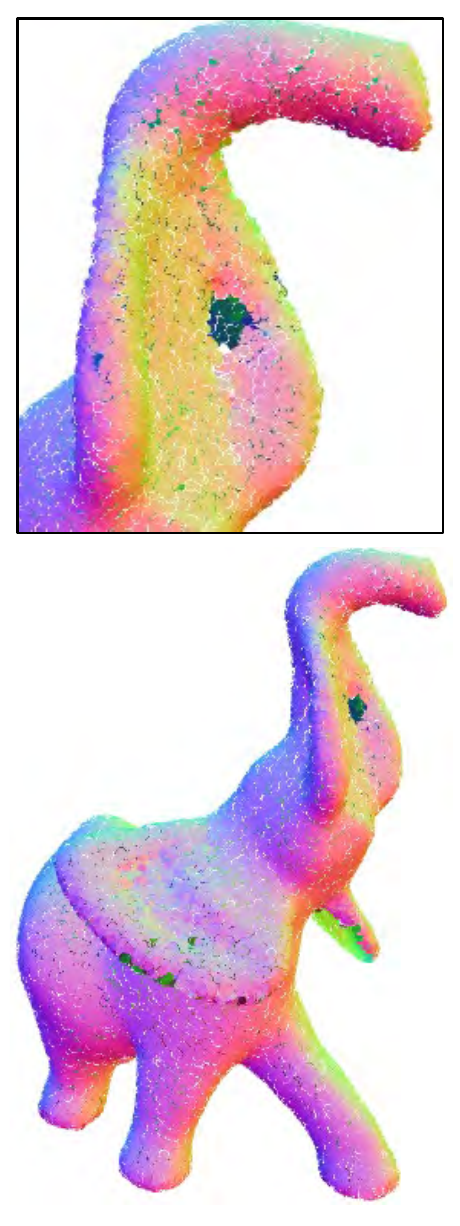}}
	\subfloat[Ours]{\label{fig:elephant-h}\includegraphics[width=0.125\textwidth]{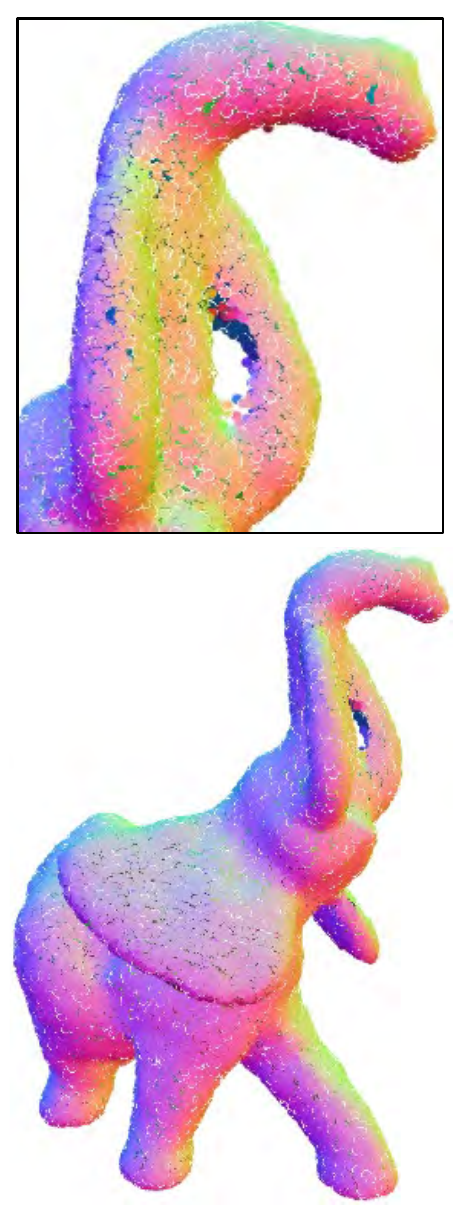}}
	\caption{Denoising results of synthetic data with 1\% non-CAD models. From left to right: noisy input, results produced by WLOP, RIMLS, EC-Net, DMR, PCN, PF, and our method, respectively. The zoomed views, shown in the first row, highlight that our method better preserves detailed features.
	}
	\label{fig:elephant}
\end{figure*}
\begin{figure*}[htb]
	\centering
	\subfloat[Noisy]{\label{fig:bunny_hi-a}\includegraphics[width=0.125\textwidth]{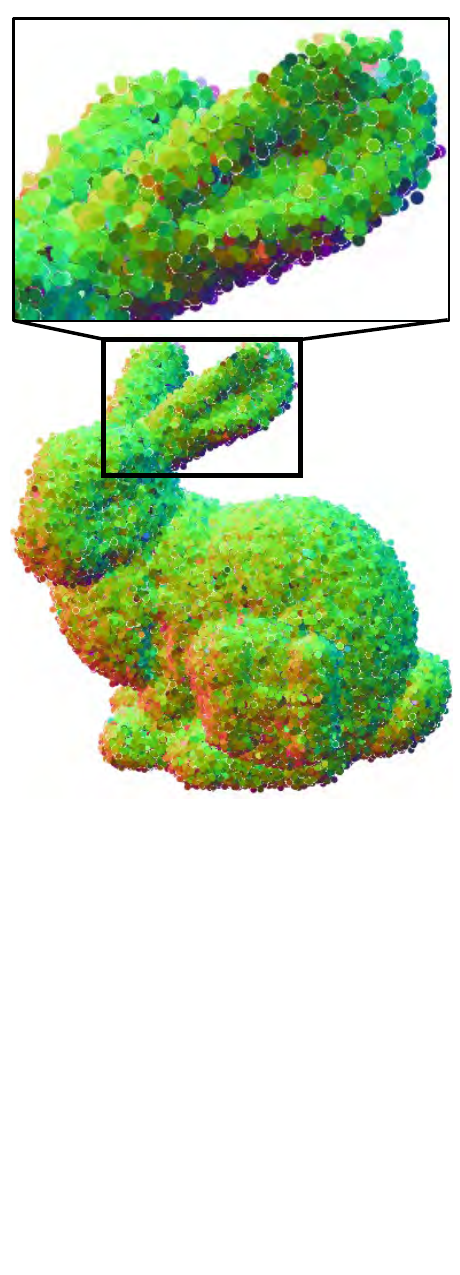}}
	\subfloat[WLOP]{\label{fig:bunny_hi-b}\includegraphics[width=0.125\textwidth]{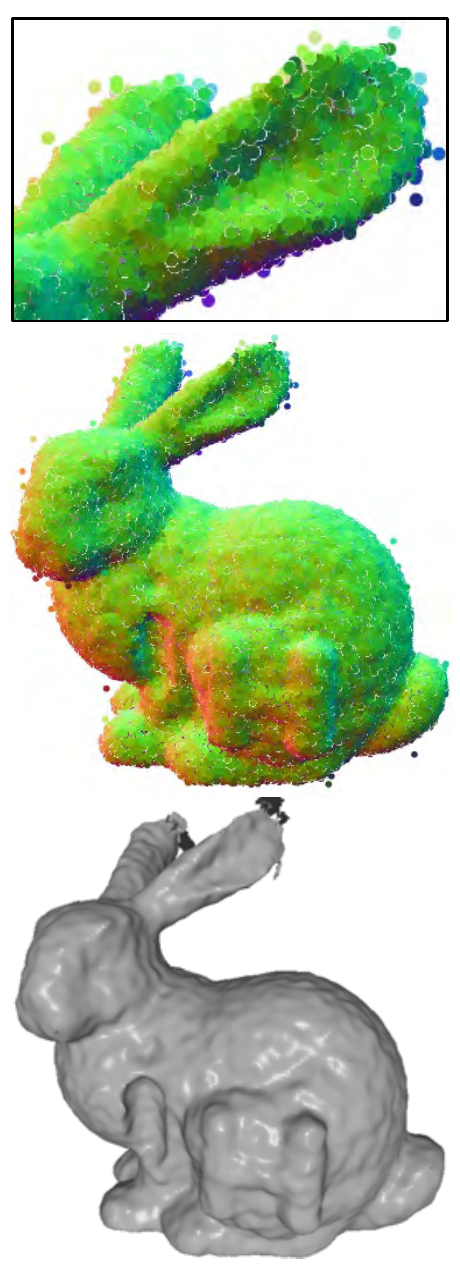}}
	\subfloat[RIMLS]{\label{fig:bunny_hi-c}\includegraphics[width=0.125\textwidth]{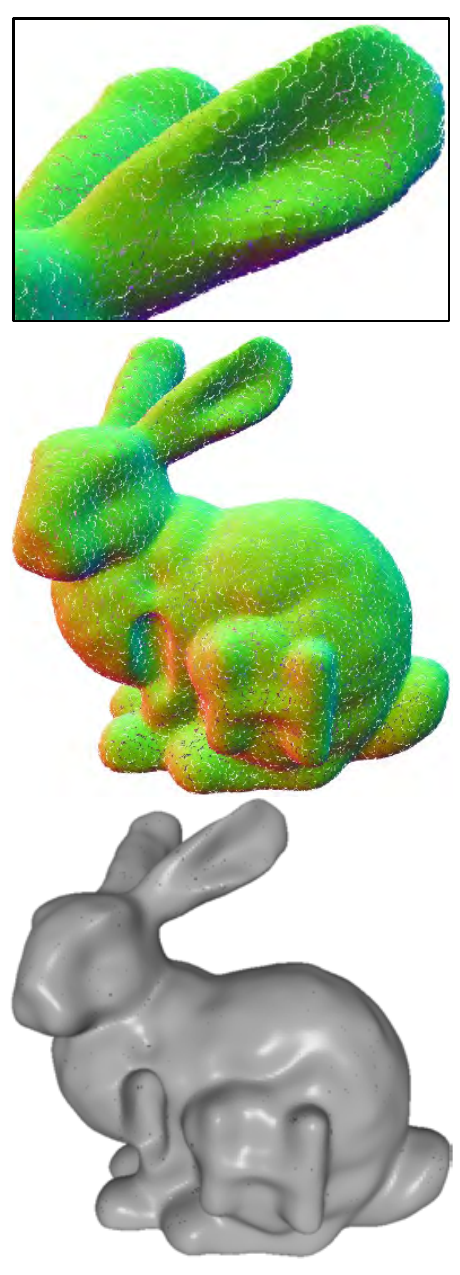}}
	\subfloat[EC-Net]{\label{fig:bunny_hi-d}\includegraphics[width=0.125\textwidth]{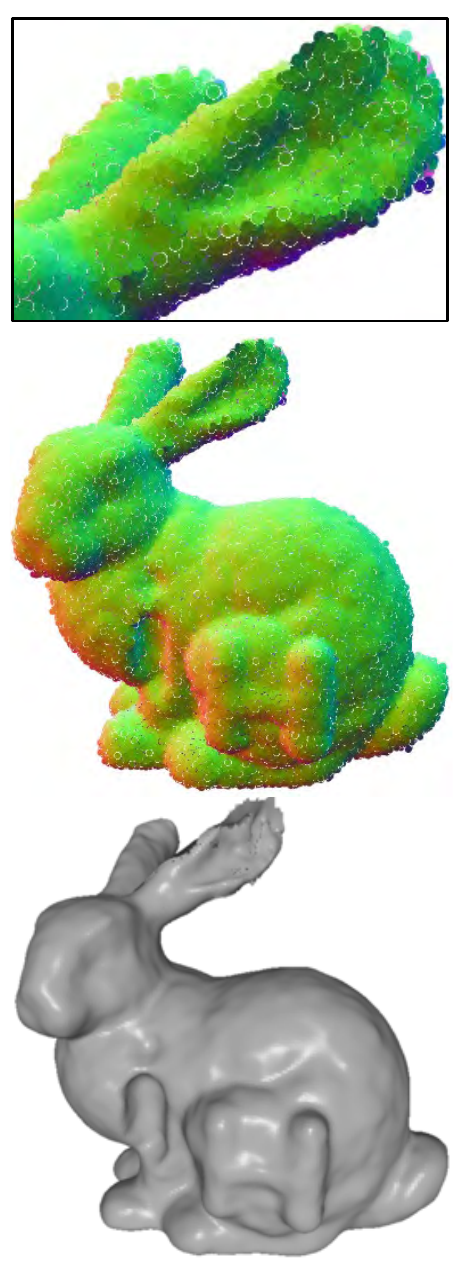}}
	\subfloat[DMR]{\label{fig:bunny_hi-e}\includegraphics[width=0.125\textwidth]{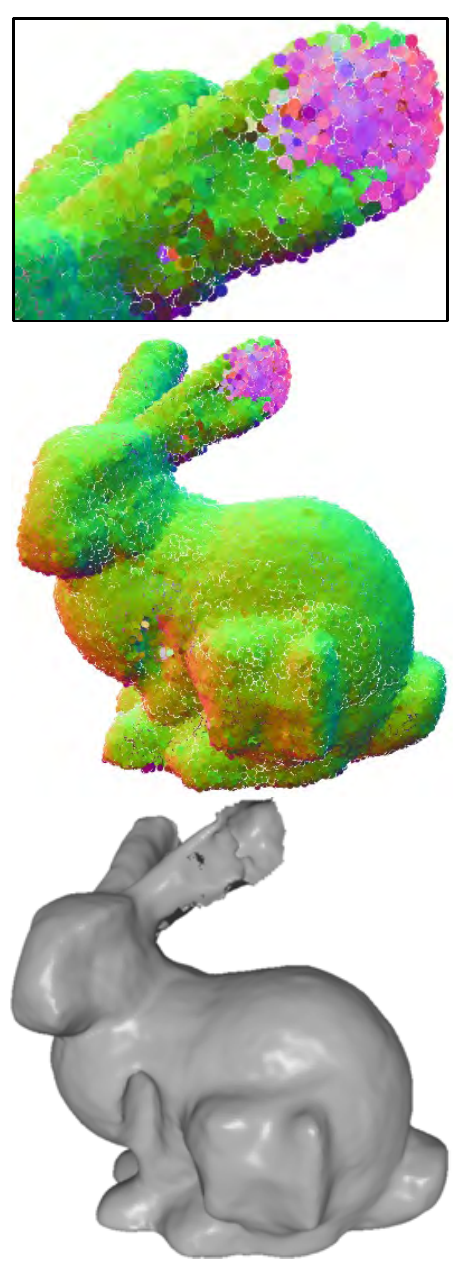}}
	\subfloat[PCN]{\label{fig:bunny_hi-f}\includegraphics[width=0.125\textwidth]{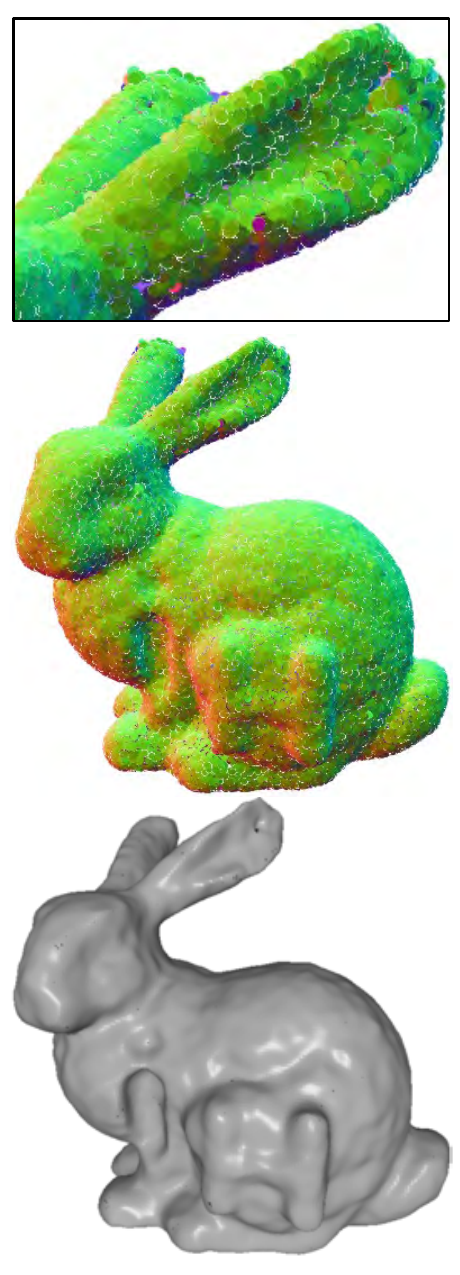}}
	\subfloat[PF]{\label{fig:bunny_hi-g}\includegraphics[width=0.125\textwidth]{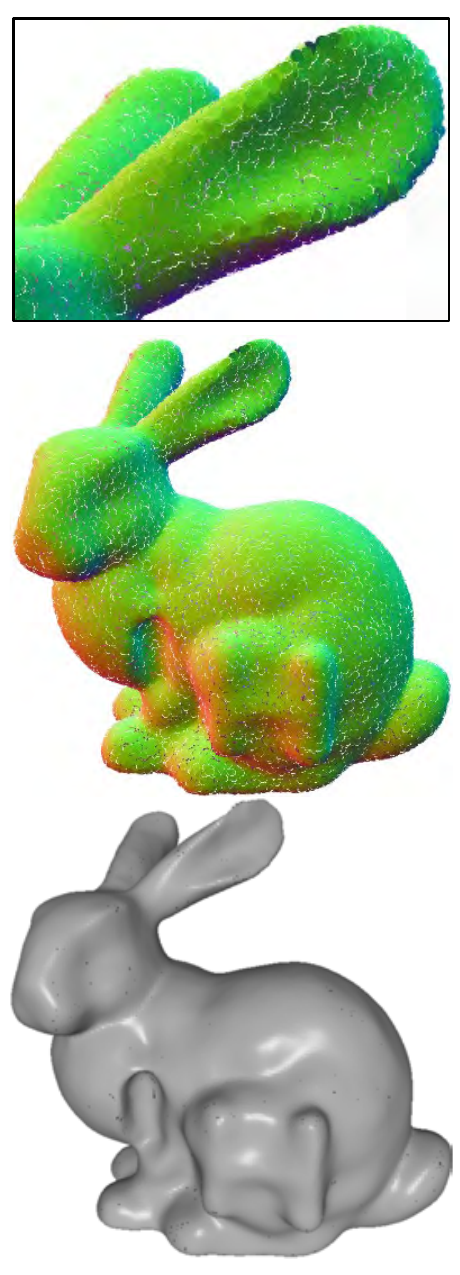}}
	\subfloat[Ours]{\label{fig:bunny_hi-h}\includegraphics[width=0.125\textwidth]{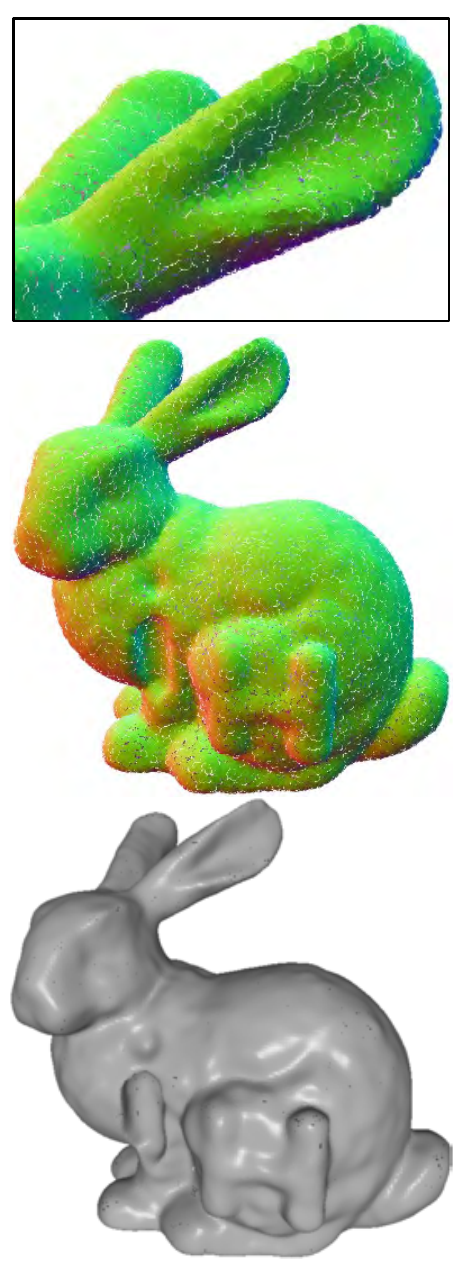}}
	\caption{Denoising results of synthetic data with 0.5\% non-CAD models. From left to right: noisy input, results produced by WLOP, RIMLS, EC-Net, DMR, PCN, PF, and our method, respectively. The zoomed views, shown in the first row, highlight that our method better preserves multiscale features.
	}
	\label{fig:bunny_hi}
\end{figure*}

Fig. \ref{fig:elephant} gives comparisons of a non-CAD surface corrupted by considerable noise. WLOP and EC-Net fail to remove noise entirely in this example.
Although DMR does a good job of noise removal, it causes shape distortion and induces bumps in the result, as shown in Fig. \ref{fig:elephant-e}.
As Figs. \ref{fig:elephant-c} and \ref{fig:elephant-f} show, RIMLS and PCN seem to have difficulty balancing the performance of noise removal and feature recovery.
PF and our method can generate visually better results than the other tested methods.
However, from the zoomed views of \ref{fig:elephant-g} and \ref{fig:elephant-h}, we observe that the trunk of the elephant in the result of PF is swelling; in contrast, our result does not induce this artifact. Therefore, our method visually yields the best result for faithfully preserving geometric features.

Fig. \ref{fig:bunny_hi} shows comparisons of a non-CAD surface with multiscale geometric features. The tested non-CAD surface is corrupted by moderate noise.
Except for WLOP, all the other methods can remove noise effectively.
DMR flattens medium- and small-scale features and induces distortion and swelling in the ear regions of the bunny surface, as depicted in Fig. \ref{fig:bunny_hi-e}.
RIMLS and PF oversmooth small-scale features to varying degrees, and PF makes this situation even worse; as illustrated in Figs. \ref{fig:bunny_hi-c} and \ref{fig:bunny_hi-g}.
Furthermore, although EC-Net and PCN can preserve different levels of geometric features in a better manner, they induce some artifacts in the ear regions of the results.
These artifacts degrade the visual quality of the denoised results and further affect the reconstruction results; see the zoomed views and reconstruction results of Figs. \ref{fig:bunny_hi-d} and \ref{fig:bunny_hi-f}.
In contrast, our method outperforms the other methods in terms of recovering most geometric features and preventing inducing annoying artifacts.

\begin{figure*}[htb]
	\centering
	\subfloat[Noisy]{\label{fig:kinect-a}\includegraphics[width=0.124\textwidth]{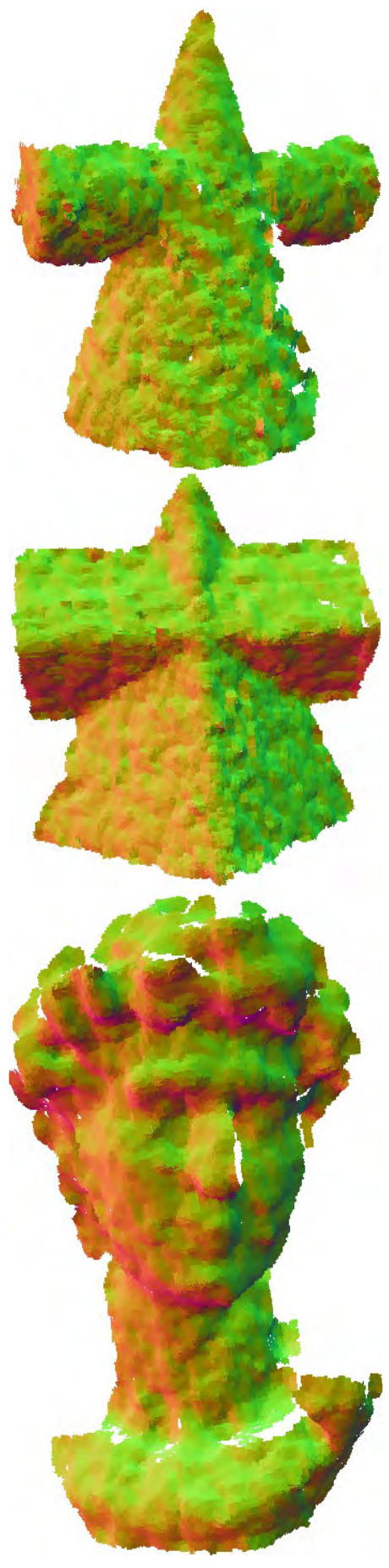}}
	\subfloat[WLOP]{\label{fig:kinect-b}\includegraphics[width=0.124\textwidth]{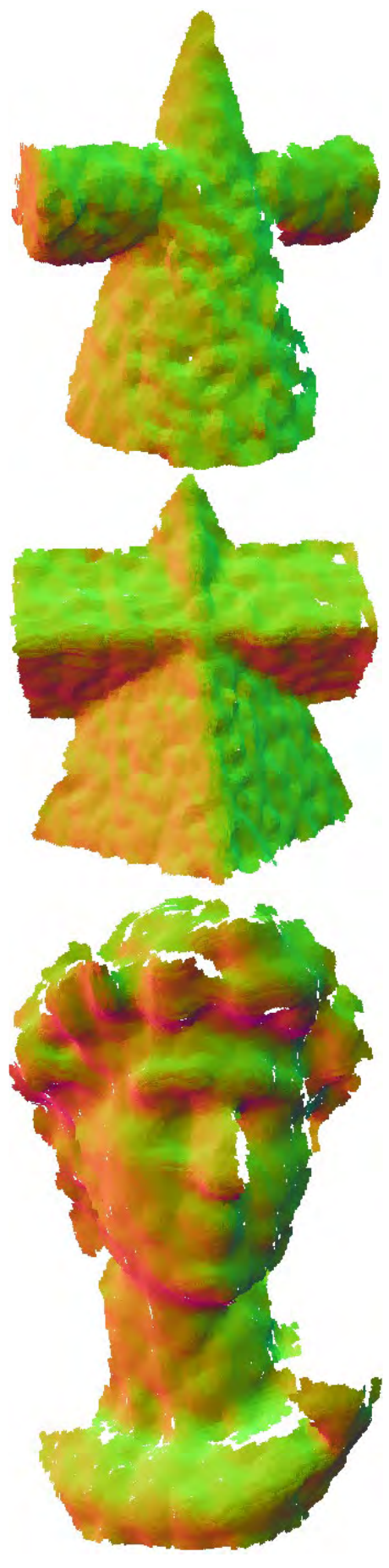}}
	\subfloat[RIMLS]{\label{fig:kinect-c}\includegraphics[width=0.124\textwidth]{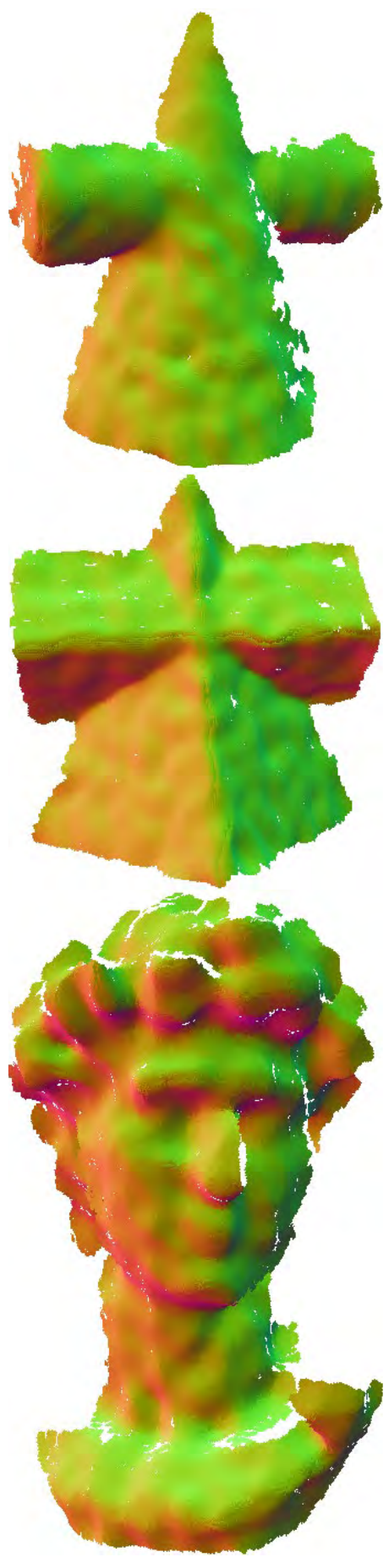}}
	\subfloat[EC-Net]{\label{fig:kinect-d}\includegraphics[width=0.124\textwidth]{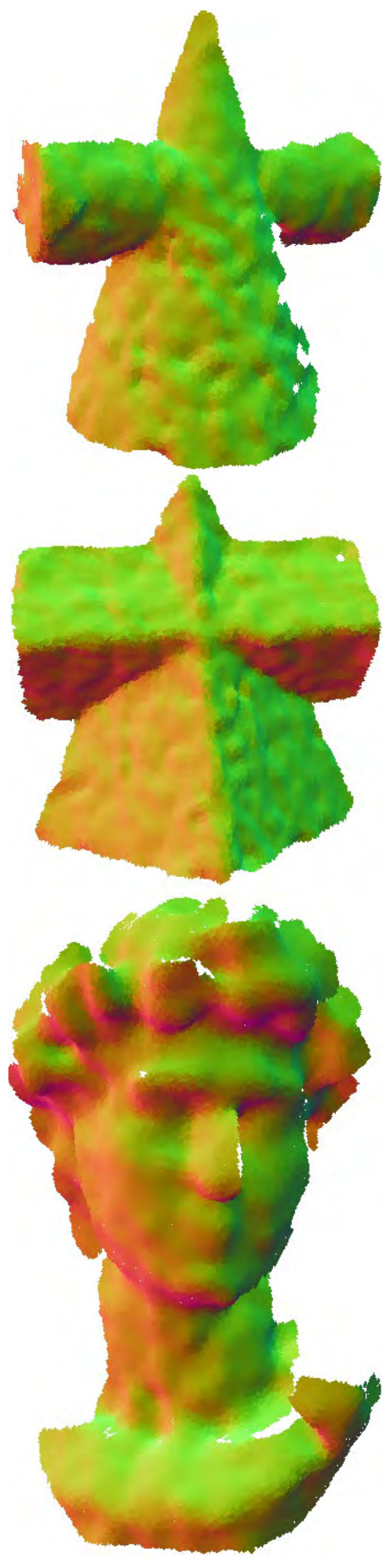}}
	\subfloat[DMR]{\label{fig:kinect-e}\includegraphics[width=0.124\textwidth]{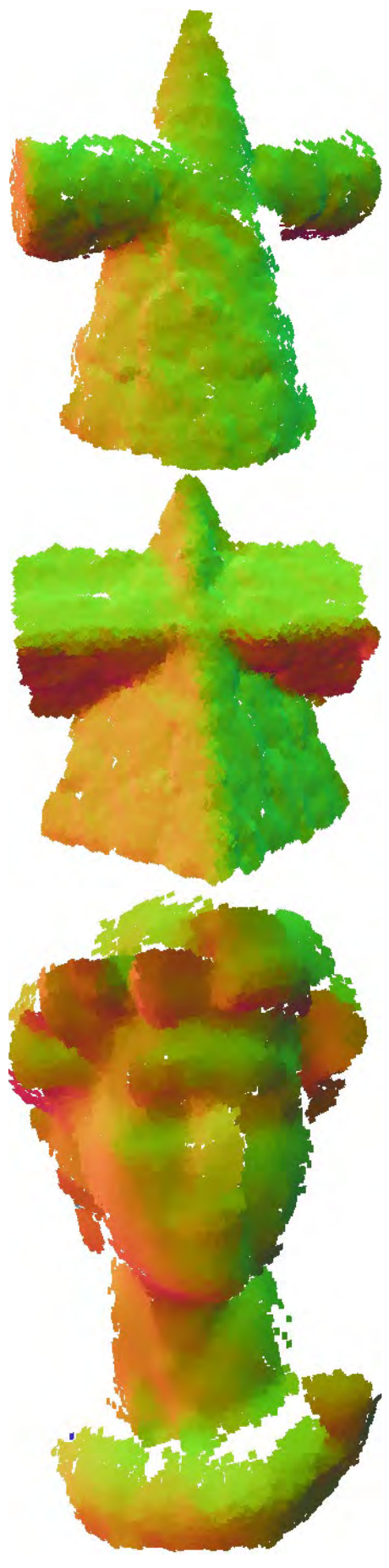}}
	\subfloat[PCN]{\label{fig:kinect-f}\includegraphics[width=0.124\textwidth]{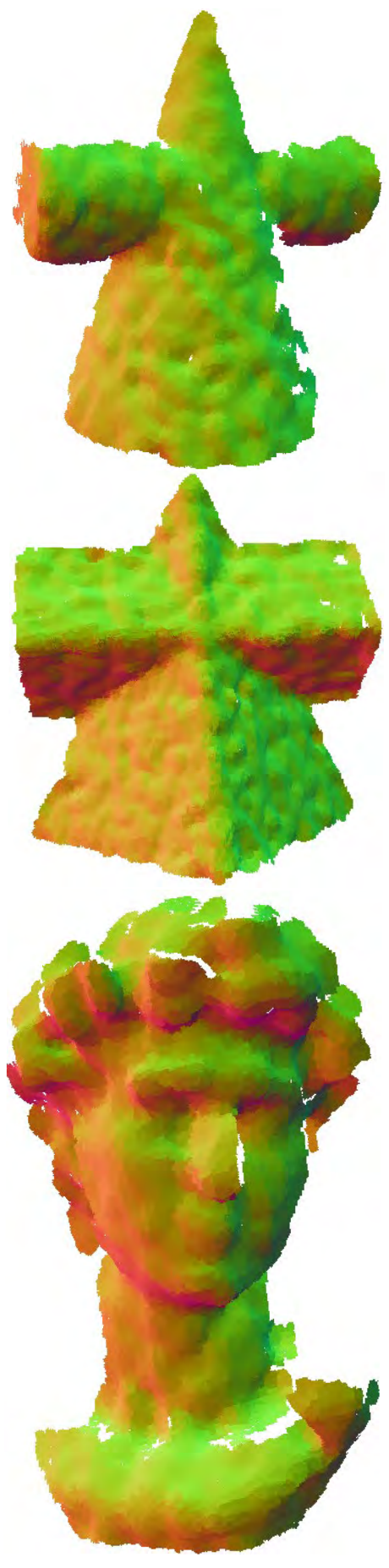}}
	\subfloat[PF]{\label{fig:kinect-g}\includegraphics[width=0.124\textwidth]{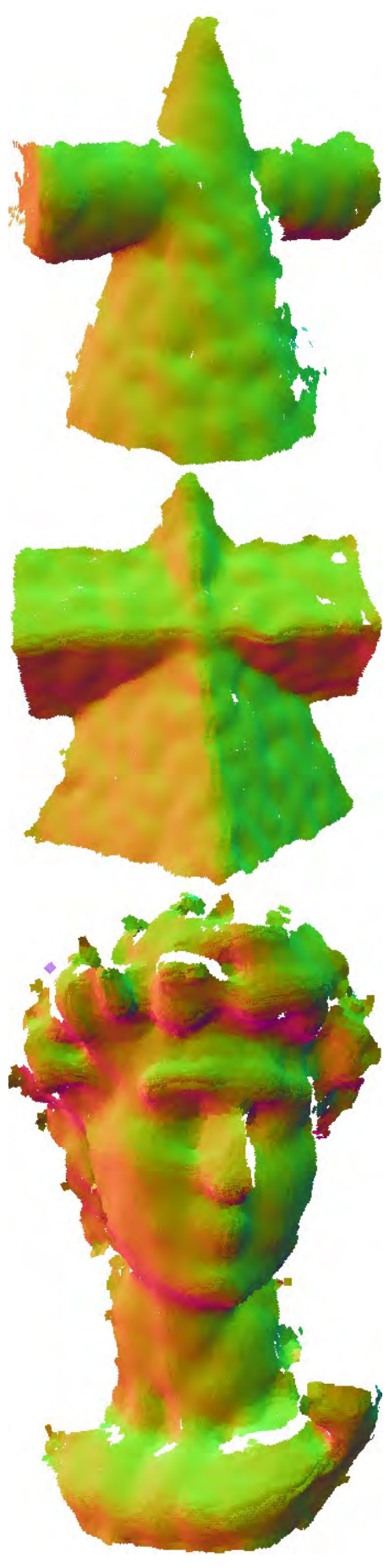}}
	\subfloat[Ours]{\label{fig:kinect-h}\includegraphics[width=0.124\textwidth]{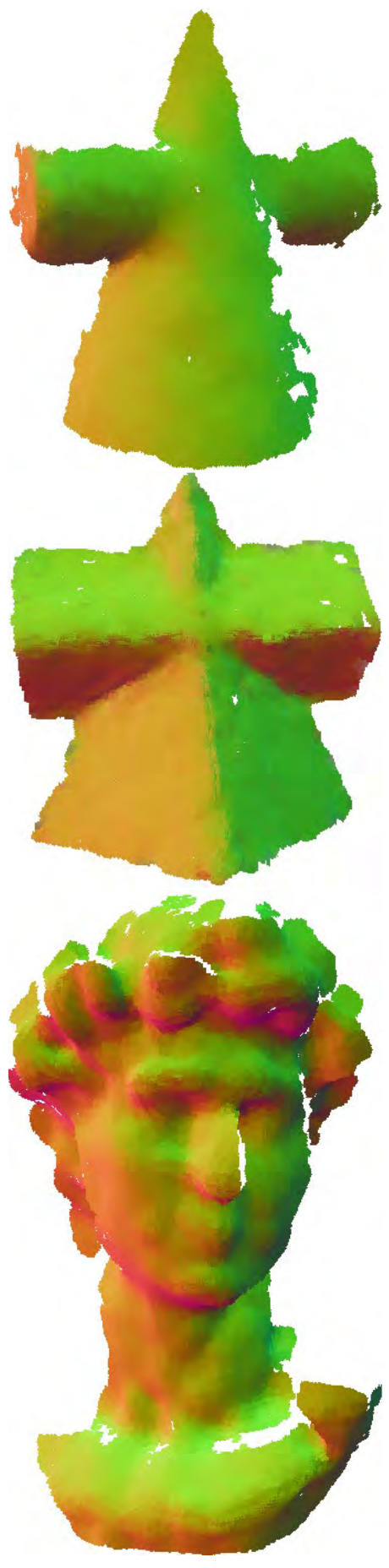}}
	\caption{ Denoising results of data captured by Kinect. From left to right: noisy input, results produced by WLOP, RIMLS, EC-Net, DMR, PCN, PF, and our method.
	}
	\label{fig:kinect}
\end{figure*}

\begin{figure*}[htb]
	\centering
	\subfloat[Noisy]{\label{Laser-a}\includegraphics[width=0.125\textwidth]{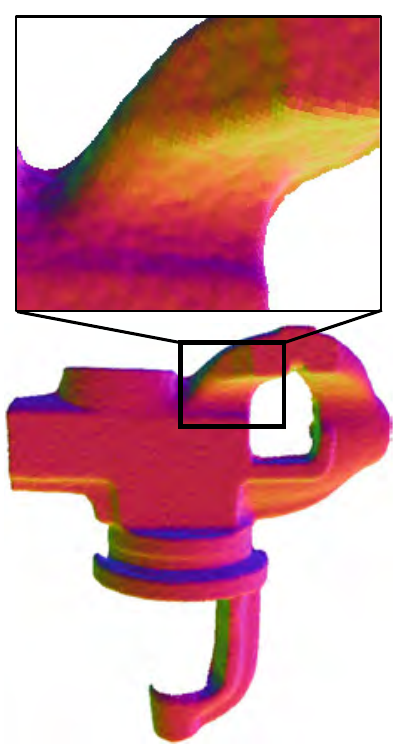}}
	\subfloat[WLOP]{\label{Laser-b}\includegraphics[width=0.125\textwidth]{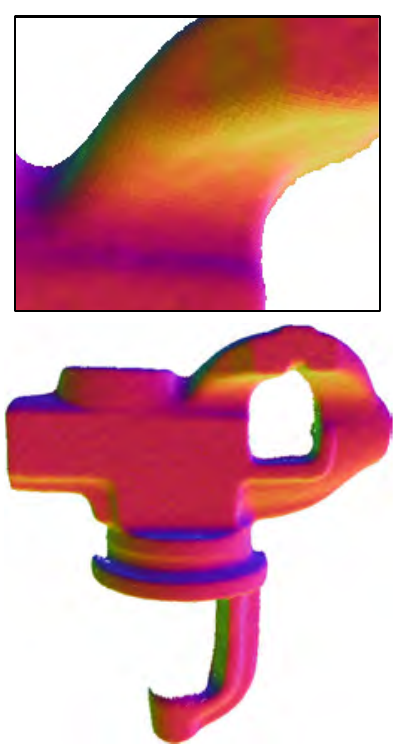}}
	\subfloat[RIMLS]{\label{Laser-c}\includegraphics[width=0.125\textwidth]{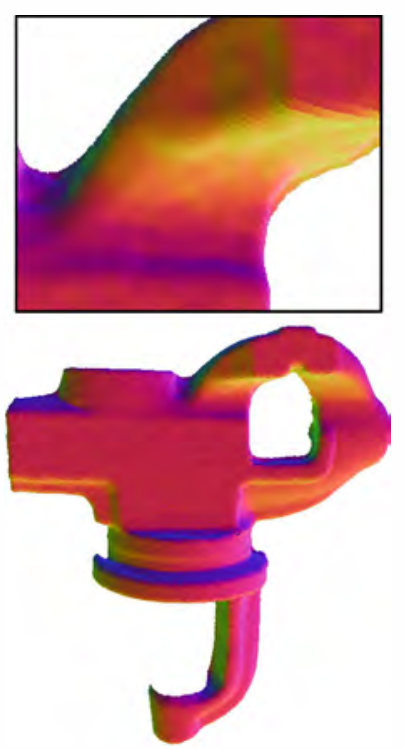}}
	\subfloat[EC-Net]{\label{Laser-d}\includegraphics[width=0.125\textwidth]{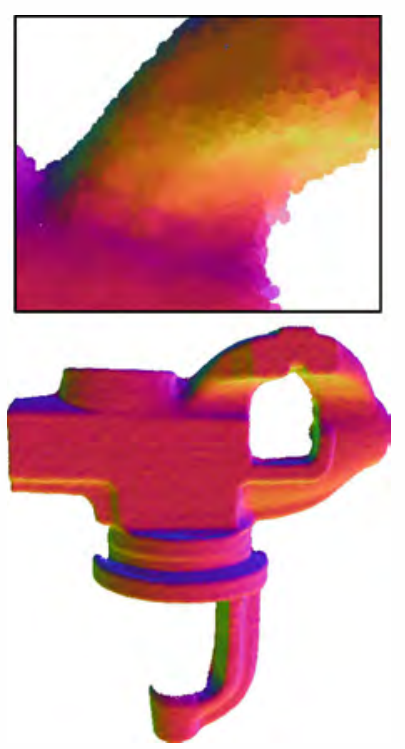}}
	\subfloat[DMR]{\label{Laser-e}\includegraphics[width=0.125\textwidth]{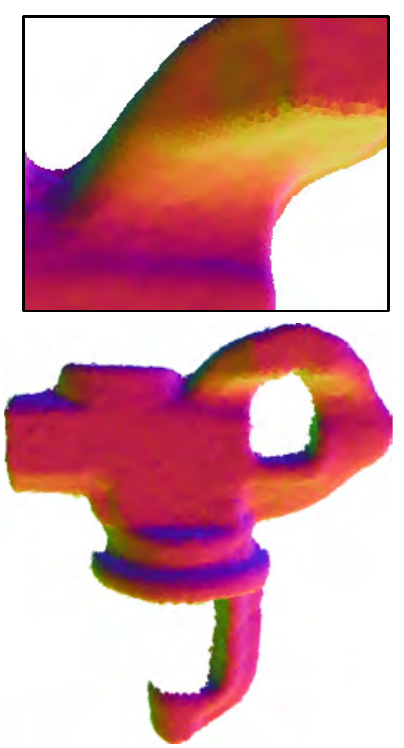}}
	\subfloat[PCN]{\label{Laser-f}\includegraphics[width=0.125\textwidth]{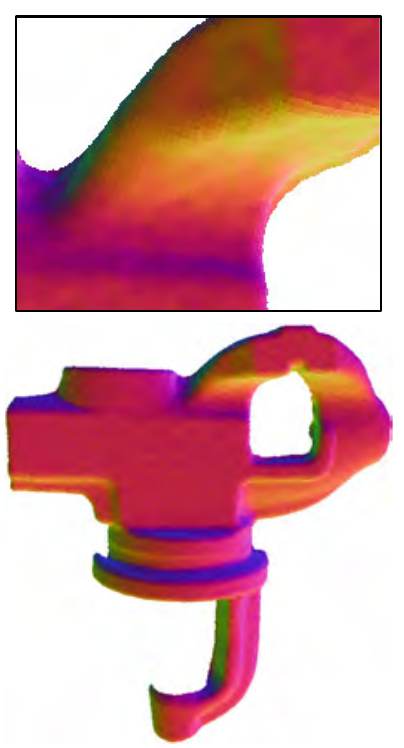}}
	\subfloat[PF]{\label{Laser-g}\includegraphics[width=0.125\textwidth]{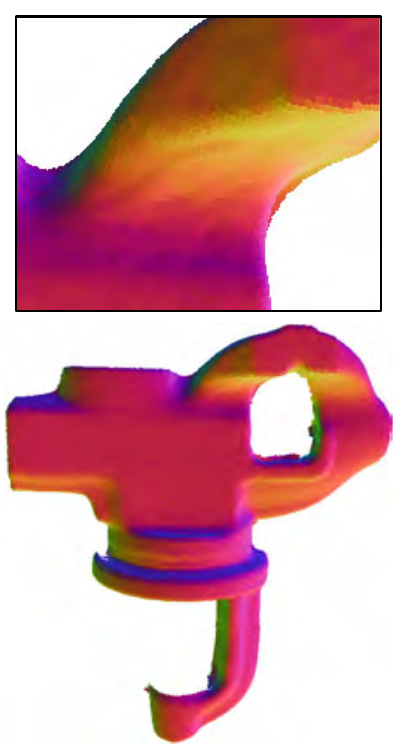}}
	\subfloat[Ours]{\label{Laser-h}\includegraphics[width=0.125\textwidth]{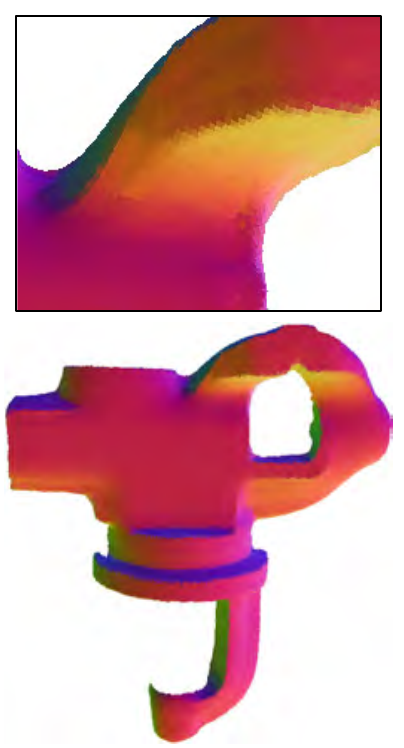}}
	\caption{ Denoising results of the laser scanned point cloud. From left to right: noisy input, results produced by WLOP, RIMLS, EC-Net, DMR, PCN, PF, and our method. The zoomed views, highlight that our method better keeps geometrical features.
	}
	\label{fig:Laser}
\end{figure*}

\begin{figure*}
	\begin{minipage}[b]{0.3\linewidth}
		   \centering
			\subfloat[Real scanned outdoor scene]{\label{RueMadame-a}\includegraphics[height=6.6cm,width=1\linewidth]{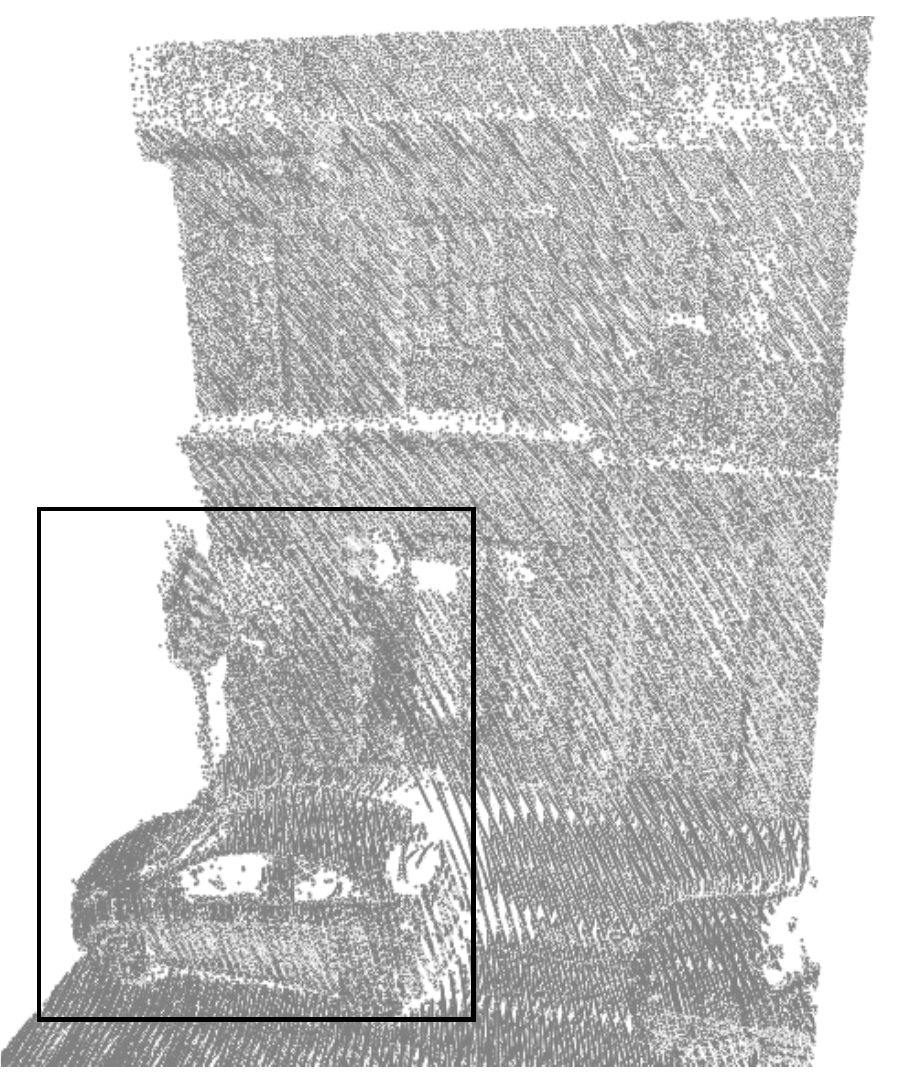}}
	\end{minipage}
	\medskip
	\begin{minipage}[b]{0.7\linewidth}
		 \centering
		\subfloat[Noisy]{\label{RueMadame-b}\includegraphics[width=0.25\linewidth]{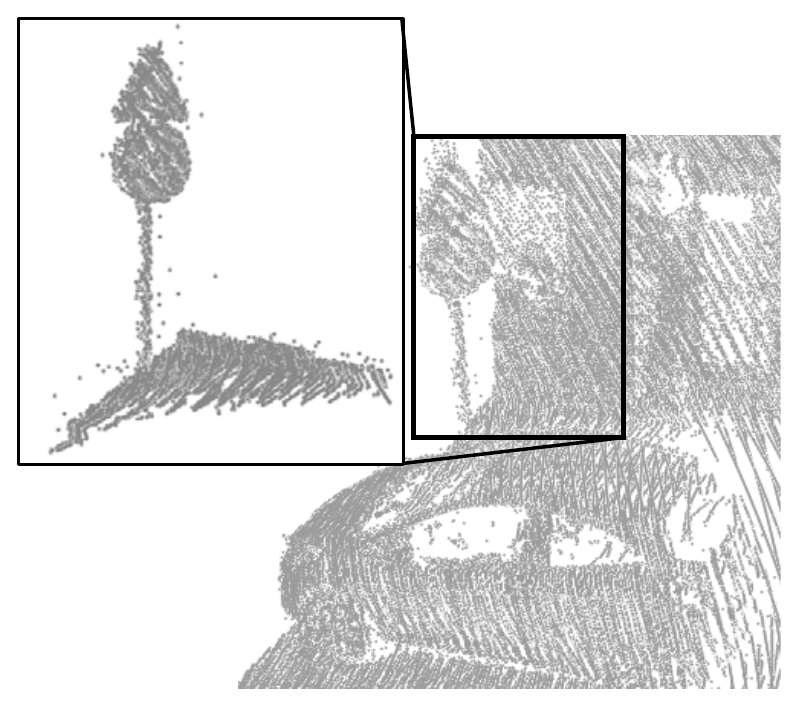}}
		\subfloat[WLOP]{\label{RueMadame-c}\includegraphics[width=0.25\linewidth]{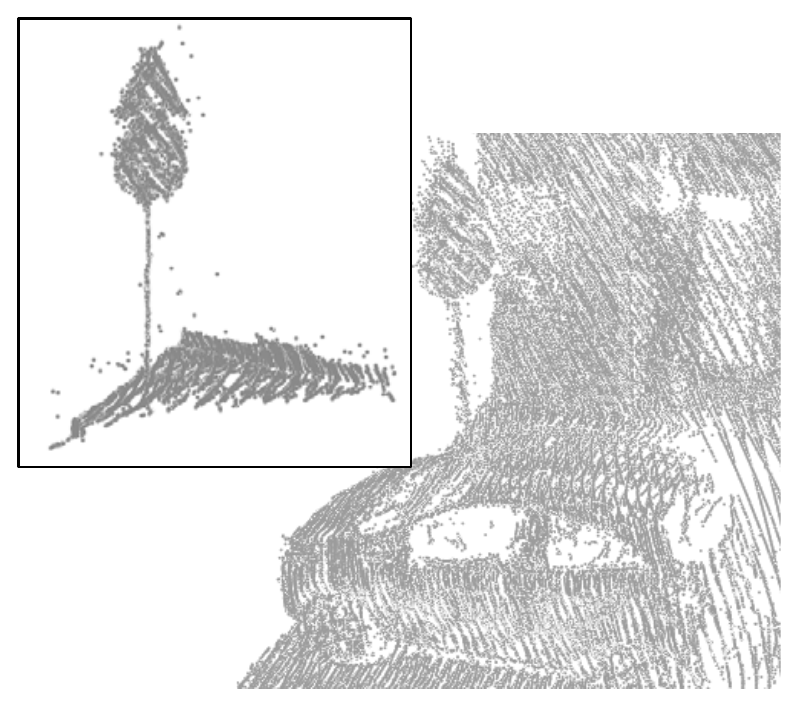}}
		\subfloat[RIMLS]{\label{RueMadame-d}\includegraphics[width=0.25\linewidth]{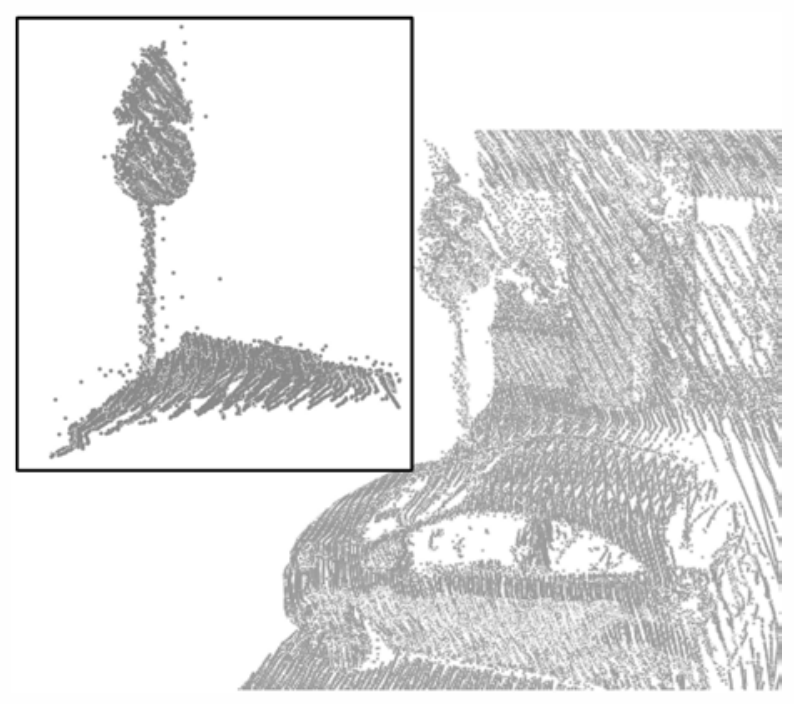}}
		\subfloat[EC-Net]{\label{RueMadame-e}\includegraphics[width=0.25\linewidth]{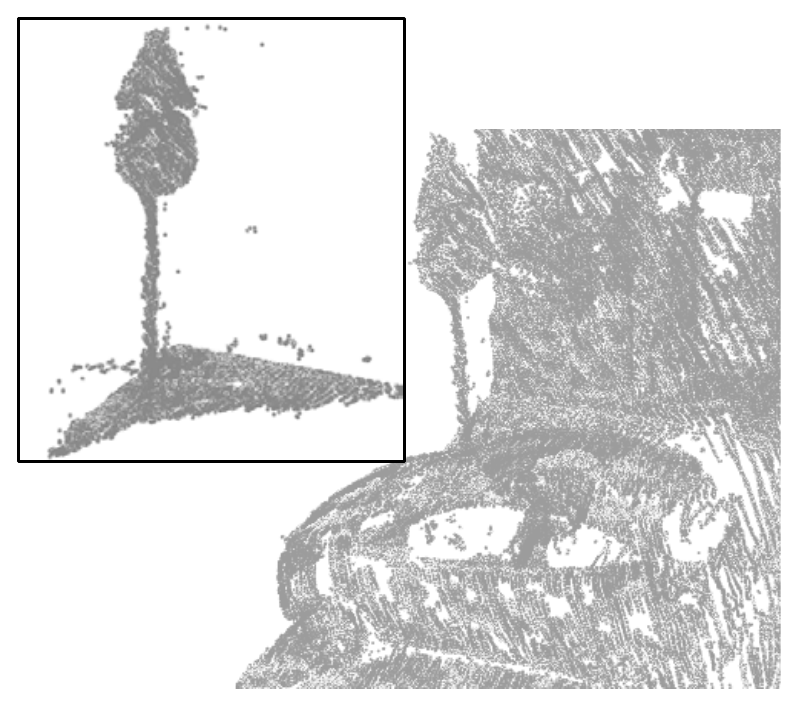}}\\
		\subfloat[DMR]{\label{RueMadame-f}\includegraphics[width=0.25\linewidth]{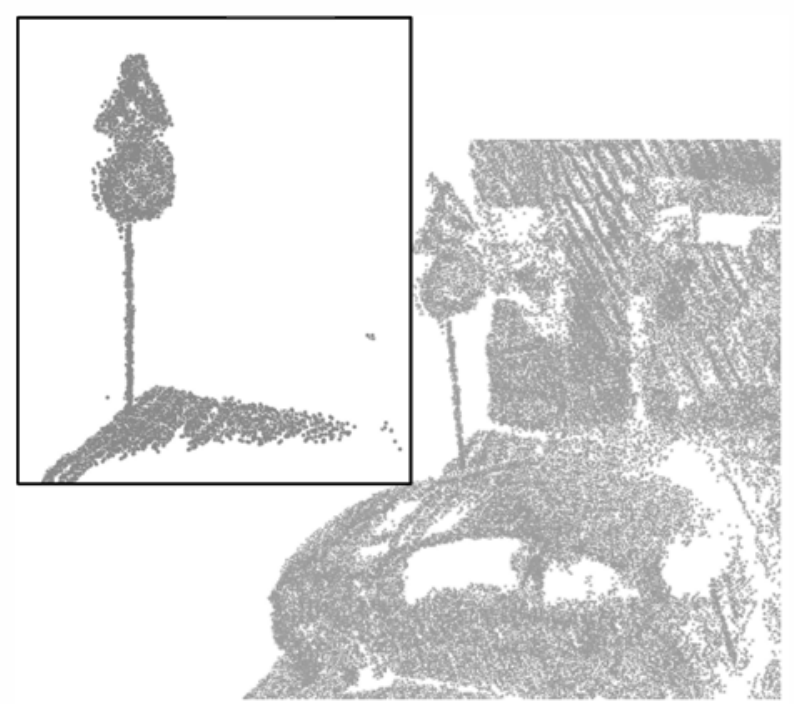}}
		\subfloat[PCN]{\label{RueMadame-g}\includegraphics[width=0.25\linewidth]{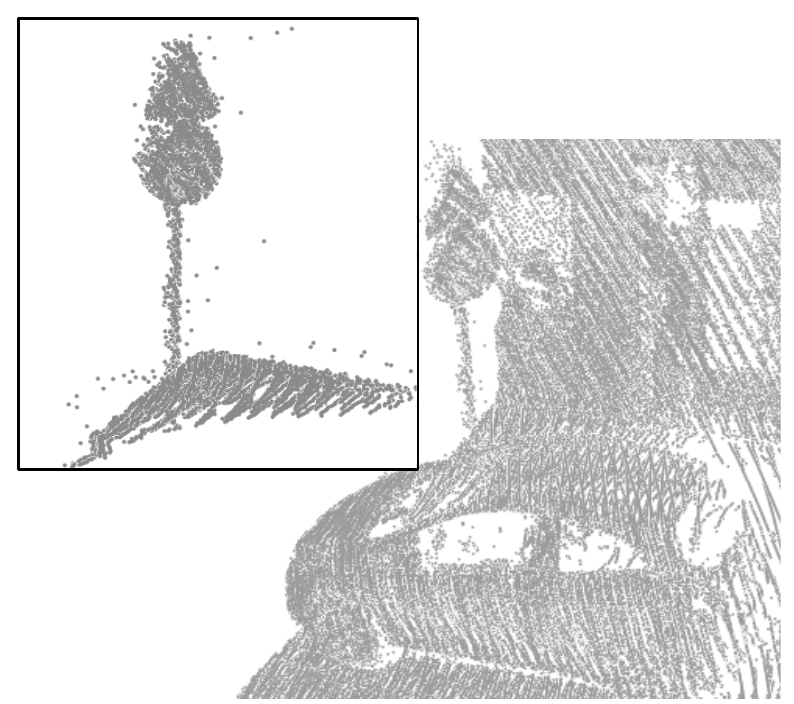}}
		\subfloat[PF]{\label{RueMadame-h}\includegraphics[width=0.25\linewidth]{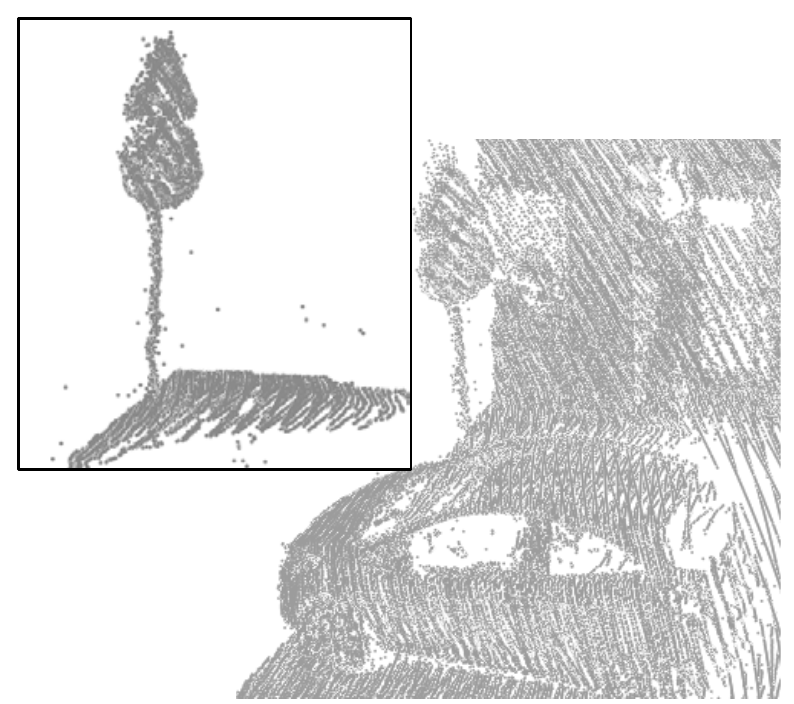}}
		\subfloat[Ours]{\label{RueMadame-j}\includegraphics[width=0.25\linewidth]{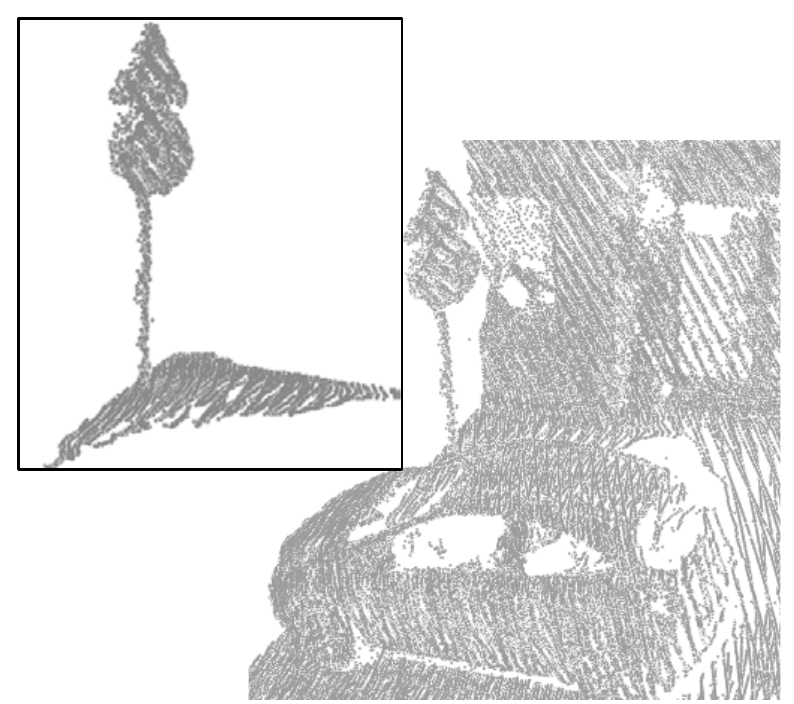}}
	\end{minipage}

     \caption{Denoising results of a real scanned point cloud scene. The zoomed views highlight that our method better removes heavy noise and avoids introducing additional artifacts.}
     \label{fig:RueMadame}
\end{figure*}

\textbf{Scanned data.}
To further validate the efficacy of our method, we conducted experiments on real scanning data without retraining the method. Fig. \ref{fig:kinect} shows the results for Kinect scanning data. The first and second rows reveal that for CAD surfaces, the other methods retained additional bumps on the denoised results, while our method effectively prevents visible artifacts, demonstrating superior performance in preserving geometric features and recovering smooth regions. For the non-CAD surface (David) in the third row of Fig. \ref{fig:kinect}, RILMS, PF, and our method generated more natural results than other methods. Yet, numerical metrics in Table~\ref{tab:denoisedCDErrors} indicate that our method outperforms RIMLS and PF in dealing with these scanned surfaces which contain complex surface characteristics.

To evaluate our method's effectiveness further, we tested it on laser-scanned data. As the zoomed views in Figure \ref{fig:Laser} indicate, while all the tested methods can remove noise and preserve geometric features to some extent, our method produces more appealing results, preserving neat structural features and recovering clean smooth regions. 
Finally, we tested our method on raw outdoor scenes from the Paris-rue-Madame Database, as illustrated in Fig. \ref{fig:RueMadame}. Compared to the other methods, our approach produces better denoised results, with fewer outliers while removing large noise in the scene.

\begin{figure*}[htb]
	\centering
	\subfloat[Noisy]{\label{fig:NonUniform-a}\includegraphics[width=0.125\textwidth]{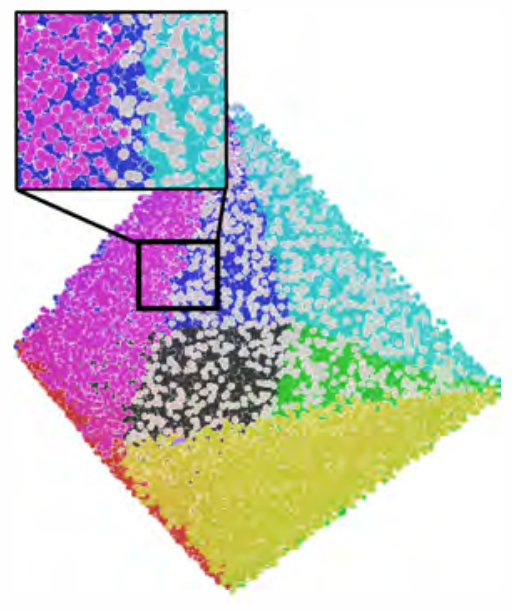}}
	\subfloat[WLOP]{\label{fig:NonUniform-b}\includegraphics[width=0.125\textwidth]{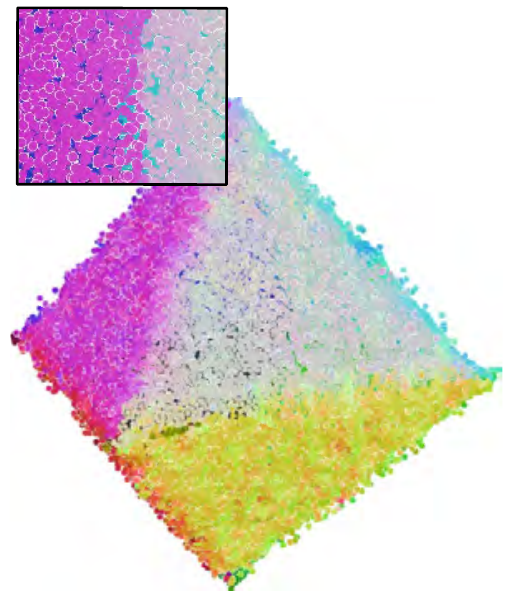}}
	\subfloat[RIMLS]{\label{fig:NonUniform-c}\includegraphics[width=0.125\textwidth]{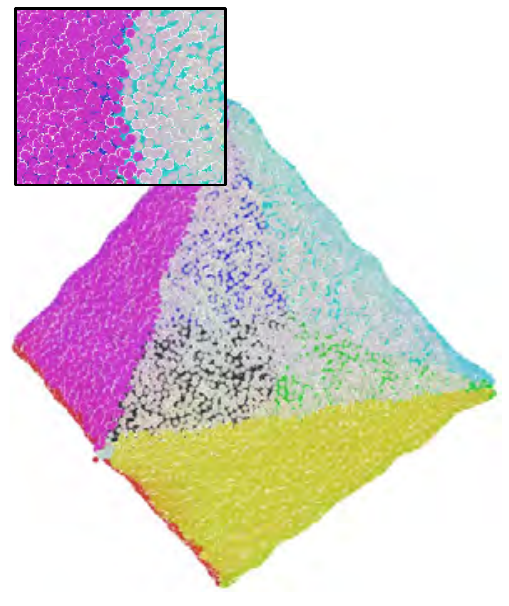}}
	\subfloat[EC-Net]{\label{fig:NonUniform-d}\includegraphics[width=0.125\textwidth]{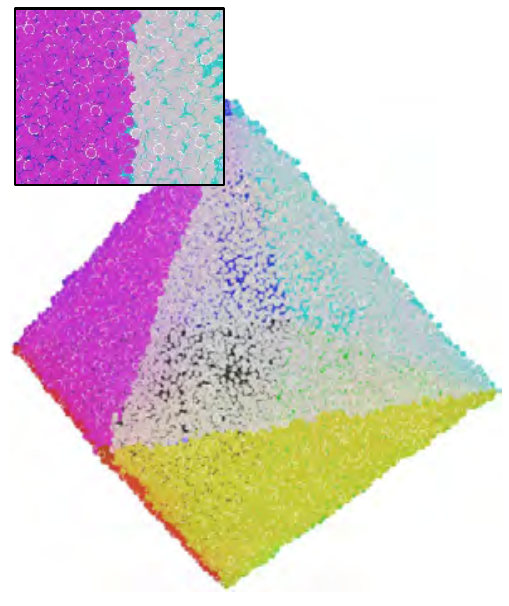}}
	\subfloat[DMR]{\label{fig:NonUniform-e}\includegraphics[width=0.125\textwidth]{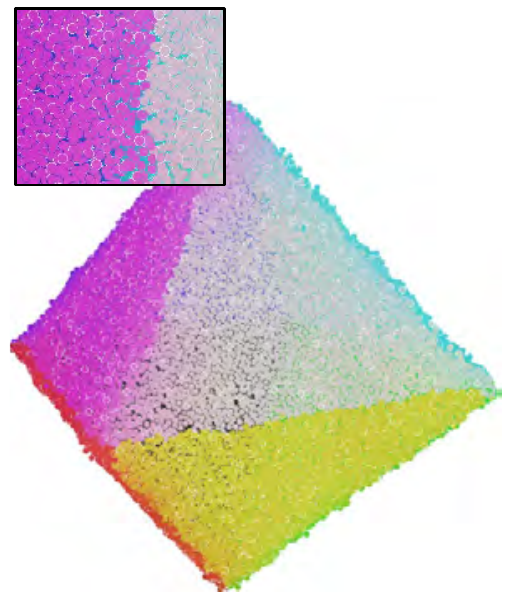}}
	\subfloat[PCN]{\label{fig:NonUniform-f}\includegraphics[width=0.125\textwidth]{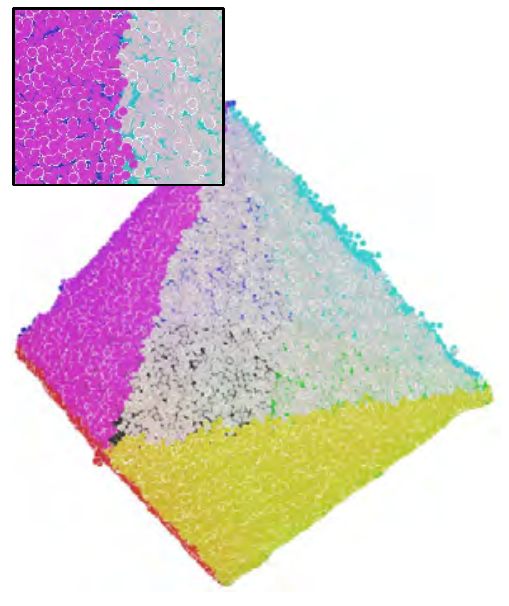}}
	\subfloat[PF]{\label{fig:NonUniform-g}\includegraphics[width=0.125\textwidth]{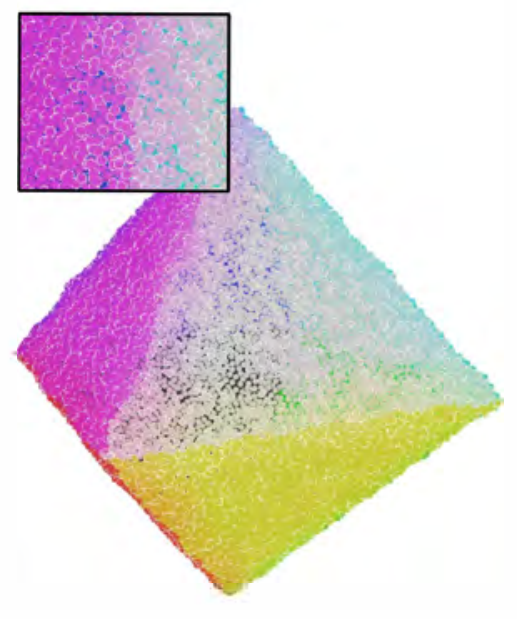}}
	\subfloat[Ours]{\label{fig:NonUniform-h}\includegraphics[width=0.125\textwidth]{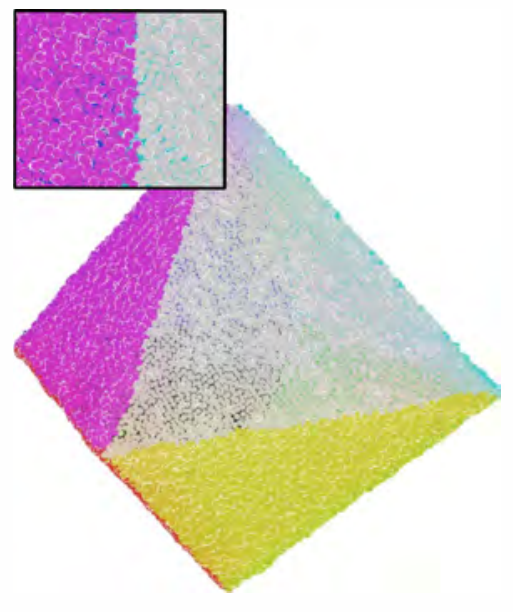}}
	\caption{ Denoising results of irregular sampling data. From left to right: noisy input, results produced by WLOP, RIMLS, EC-Net, DMR, PCN, PF, and our method.
	}
	\label{fig:NonUniform}
\end{figure*}

\begin{figure}[htb]
	\centering
	\subfloat[1\%]{\label{fig:noiselevel-a}\includegraphics[width=0.125\textwidth]{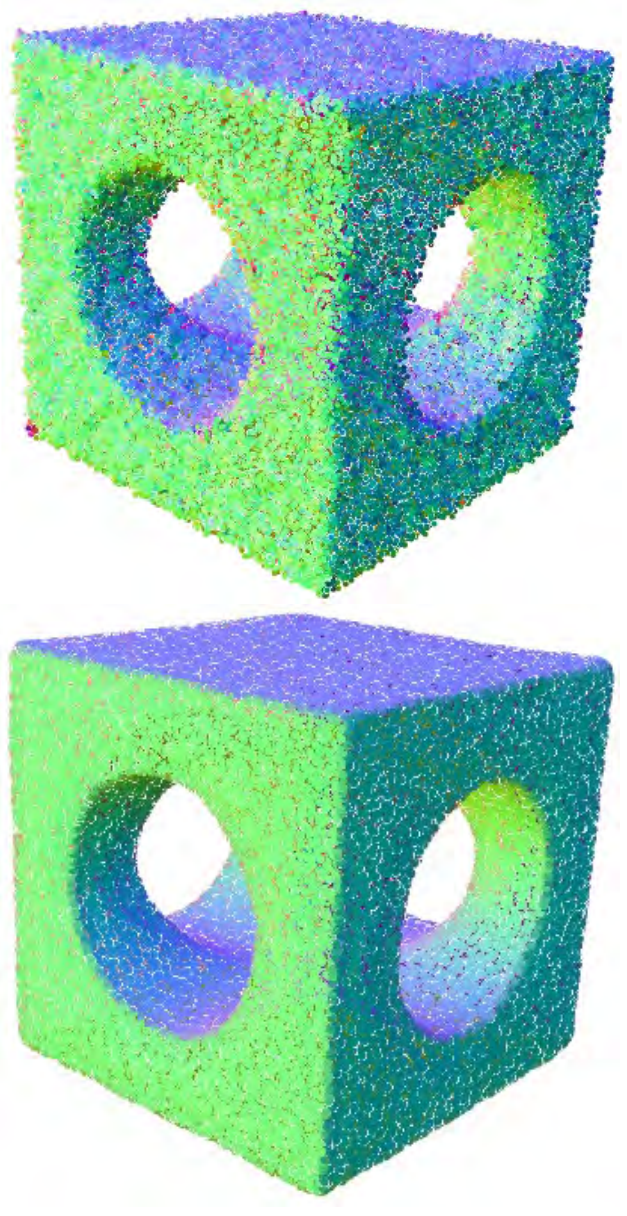}}
	\subfloat[2\%]{\label{fig:noiselevel-b}\includegraphics[width=0.125\textwidth]{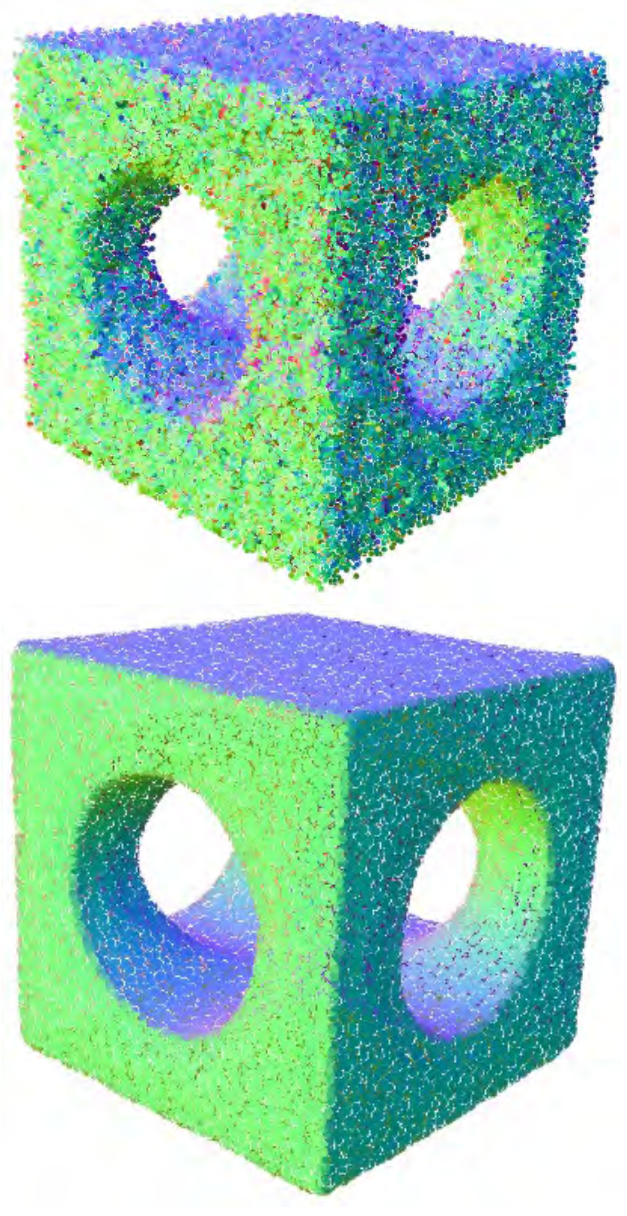}}
	\subfloat[3\%]{\label{fig:noiselevel-c}\includegraphics[width=0.125\textwidth]{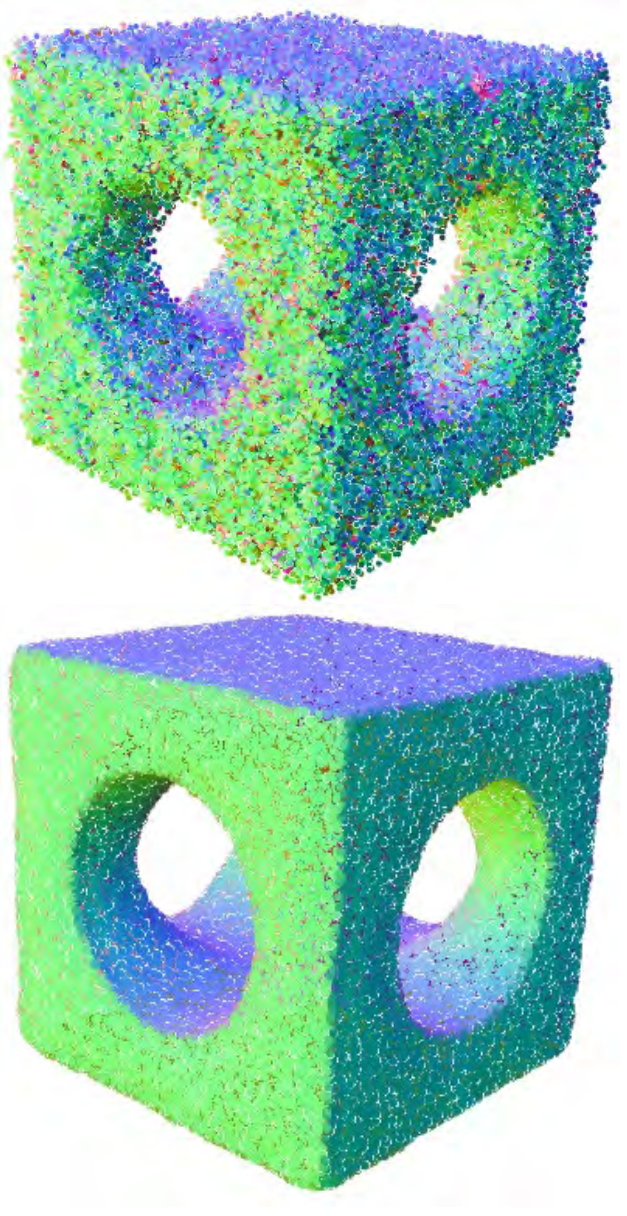}}
	\subfloat[4\%]{\label{fig:noiselevel-d}\includegraphics[width=0.125\textwidth]{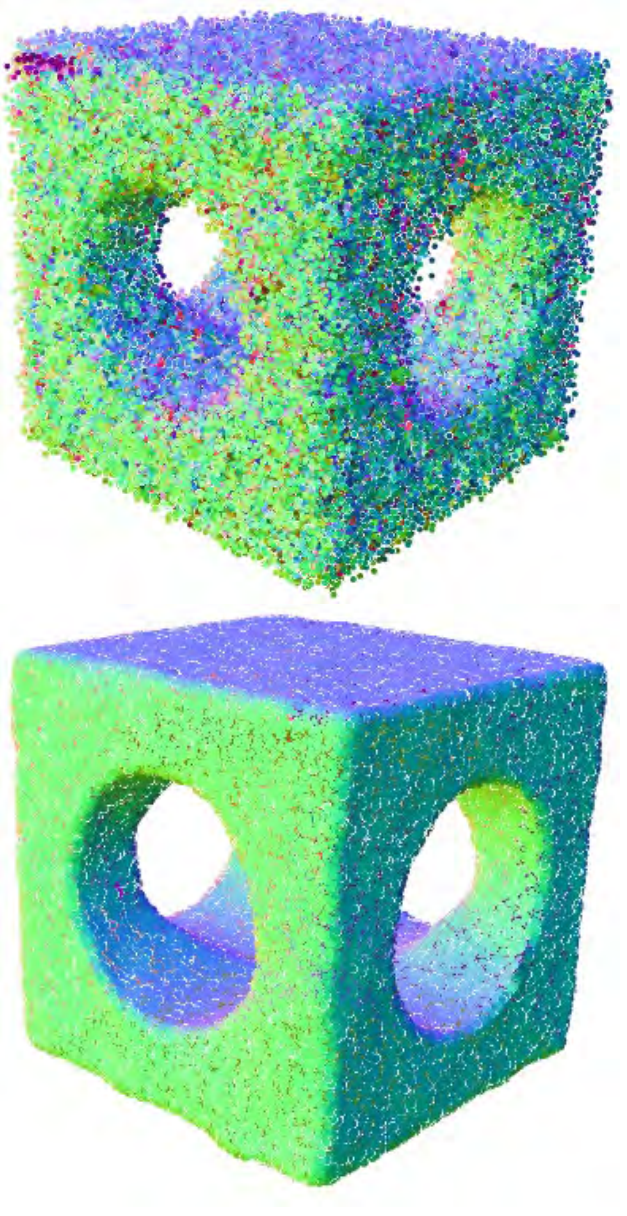}}
	\caption{ Denoising results of Cylinder corrupted by different levels of noise. The first row shows noisy point clouds (1\%, 2\%, 3\%, and 4\% noise), while the second row shows the corresponding denoising results produced by our method.
	}
	\label{fig:noiselevel}
\end{figure}

\begin{figure}[htb]
	\centering
	\subfloat{\label{fig:outliers}\includegraphics[width=0.25\textwidth]{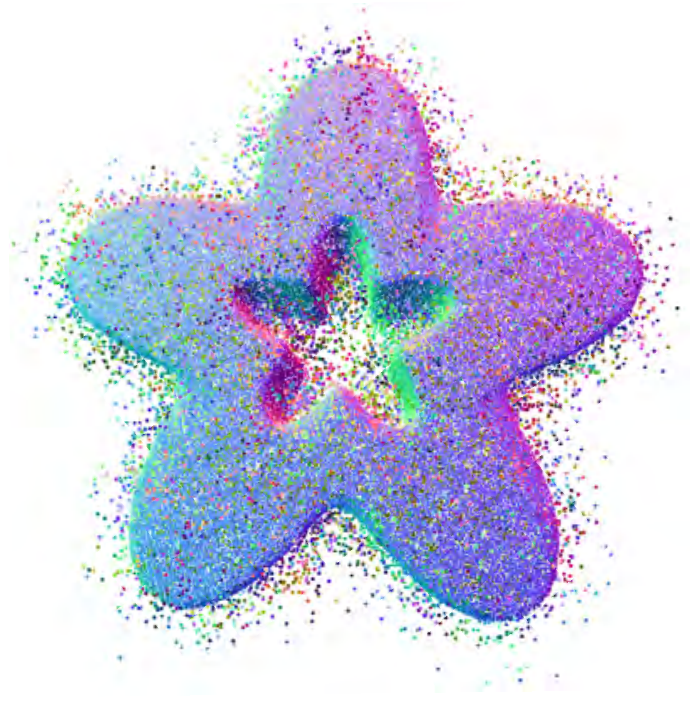}}
	\subfloat{\label{fig:outliers-remove}\includegraphics[width=0.25\textwidth]{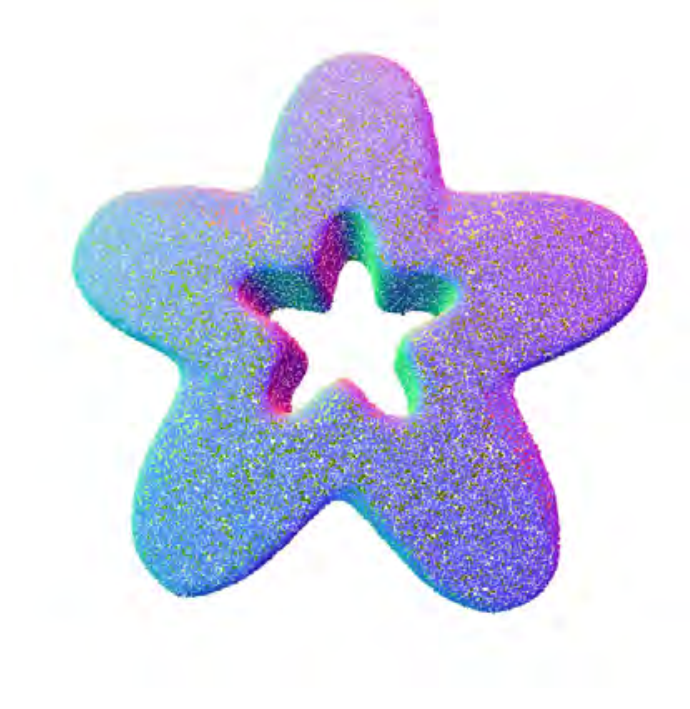}}
		\caption{Denoising result of Trim\_star with outliers. Noisy input (left) and the corresponding denoising result (right).
	}
	\label{fig:outliers}
\end{figure}

\textbf{Robustness tests.} To further verify the robustness of our method, we demonstrate the performance of our method against irregular sampling, different noise levels, and outliers in the following.
We show the effectiveness of our method against non-uniform sampling in Fig. \ref{fig:NonUniform}. Although the noisy surface suffers from irregular sampling, our method is noticeably better than all other compared methods, which can produce a compelling result that preserves sharp edges and corners.
Fig. \ref{fig:noiselevel} shows the robustness of our method against different noise levels. As we can see in Figs. \ref{fig:noiselevel-a}, \ref{fig:noiselevel-b}, and \ref{fig:noiselevel-c}, our method can remove noise while preserving sharp features well when the noise level is increased.
However, when the noise level is larger than the geometric feature sizes, our method may blur some geometric features and cannot produce satisfactory results; see Fig. \ref{fig:noiselevel-d}.
Since our method takes into account the handling of outliers, it can deal with the surface corrupted by a larger number of outliers effectively; see Fig. \ref{fig:outliers}.

\begin{table*}[htp]
	\centering    \footnotesize
	\caption{Quantitative evaluation of results in Figs. \ref{fig:fandisk}, \ref{fig:elephant}, \ref{fig:bunny_hi}, and \ref{fig:kinect}. For each result, we list CD ($\times 10^{-5}$ ) and P2S ($\times 10^{-3}$ ) errors and running time (in seconds).}
	
	\begin{tabular}{cccccccccccccccccccccc}
		\toprule
		
		\footnotesize Model &
		\multicolumn{2}{c}{\footnotesize WLOP}&
		\multicolumn{2}{c}{\footnotesize RIMLS}&
		\multicolumn{2}{c}{\footnotesize EC-NET}&
		\multicolumn{2}{c}{\footnotesize DMR}&
		\multicolumn{2}{c}{\footnotesize PCN}&
		\multicolumn{2}{c}{\footnotesize PF}&
		\multicolumn{2}{c}{\footnotesize Ours}
		\\
		\midrule
		\footnotesize PartLp &
		\multicolumn{2}{r|}{18.09, 10.10; 89.0} &
		\multicolumn{2}{r|}{2.78, 2.66; 708.2} &
		\multicolumn{2}{r|}{4.15, 5.96; 51.1} &
		\multicolumn{2}{r|}{3.32, 3.18; 14.3} &
		\multicolumn{2}{r|}{1.84, 1.34; 704.0} &
		\multicolumn{2}{r|}{1.02, 1.02; 152.1} &
		\multicolumn{2}{r}{\textbf{0.93}, \textbf{0.96}; 311.1}\\
		
		\footnotesize Boxunion &
		\multicolumn{2}{r|}{4.80, 3.71; 94.7} &
		\multicolumn{2}{r|}{2.57, 2.31; 669.3} &
		\multicolumn{2}{r|}{4.29, 5.93; 49.8} &
		\multicolumn{2}{r|}{2.98, 3.11; 20.6} &
		\multicolumn{2}{r|}{1.98, 1.94; 726.0} &
		\multicolumn{2}{r|}{1.29, 1.11; 170.2} &
		\multicolumn{2}{r}{\textbf{1.21}, \textbf{1.06}; 303.5}\\
		
		\footnotesize Elephant &
		\multicolumn{2}{r|}{35.10, 8.20; 107.3} &
		\multicolumn{2}{r|}{11.42, 5.40; 551.7} &
		\multicolumn{2}{r|}{9.83, 4.73; 52.2} &
		\multicolumn{2}{r|}{7.01, 5.41; 14.7} &
		\multicolumn{2}{r|}{3.65, 3.11; 726.5} &
		\multicolumn{2}{r|}{4.48, 2.86; 163.9} &
		\multicolumn{2}{r}{\textbf{2.87}, \textbf{2.78}; 310.9}\\
		
		\footnotesize Bunny\_{}Hi &
		\multicolumn{2}{r|}{1.29, 1.82; 56.2} &
		\multicolumn{2}{r|}{1.06, 1.64; 187.7} &
		\multicolumn{2}{r|}{0.90, 2.16; 52.2} &
		\multicolumn{2}{r|}{1.87, 2.40; 15.2} &
		\multicolumn{2}{r|}{1.03, 1.32; 736.1} &
		\multicolumn{2}{r|}{0.81, 1.00; 155.4} &
		\multicolumn{2}{r}{\textbf{0.70}, \textbf{0.82}; 311.3}\\
		
		\footnotesize Cone &
		\multicolumn{2}{r|}{20.97, 7.50; 10.7} &
		\multicolumn{2}{r|}{18.04, 5.73; 27.2} &
		\multicolumn{2}{r|}{19.74, 7.17; 6.7} &
		\multicolumn{2}{r|}{28.19, 6.90; 2.4} &
		\multicolumn{2}{r|}{19.47, 6.92; 47.9} &
		\multicolumn{2}{r|}{17.54, \textbf{5.71}; 29.5} &
		\multicolumn{2}{r}{\textbf{17.08}, \textbf{5.71}; 34.1}\\

		\footnotesize Pyramid &
		\multicolumn{2}{r|}{20.18, 6.60; 9.1} &
		\multicolumn{2}{r|}{17.26, 5.77; 21.8} &
		\multicolumn{2}{r|}{19.86, 7.04; 8.6} &
		\multicolumn{2}{r|}{22.21, 7.07; 2.8} &
		\multicolumn{2}{r|}{20.16, 6.84; 56.3} &
		\multicolumn{2}{r|}{18.20, 5.90; 20.4} &
		\multicolumn{2}{r}{\textbf{17.17}, \textbf{5.66}; 45.1}\\
		
		\footnotesize David &
		\multicolumn{2}{r|}{15.78, 5.56; 13.8} &
		\multicolumn{2}{r|}{19.03, 5.45; 40.0} &
		\multicolumn{2}{r|}{15.27, 6.02; 9.7} &
		\multicolumn{2}{r|}{26.47, 6.96; 2.9} &
		\multicolumn{2}{r|}{15.30, 5.87; 67.9} &
		\multicolumn{2}{r|}{16.37, 5.51; 26.2} &
		\multicolumn{2}{r}{\textbf{15.01}, \textbf{5.41}; 49.1}\\
		\bottomrule
	\end{tabular}
	\label{tab:denoisedCDErrors}
\end{table*}

\textbf{Quantitative evaluation.} We observe from the aforementioned qualitative comparisons that our method can generate visually better denoised results than competing methods.
Here, we compare them numerically. We utilize Chamfer distance (CD) and Point-to-surface distance (P2S) to measure the fidelity of the denoised result to the ground truth point cloud. The evaluation metrics are widely used in work \cite{chen2022repcd, zhang2020pointfilter, rakotosaona2020pointcleannet}.
We first compare the examples in Figs. \ref{fig:fandisk}, \ref{fig:elephant}, \ref{fig:bunny_hi} and \ref{fig:kinect} and list the evaluation results in Table \ref{tab:denoisedCDErrors}.
As we can see, our method outperforms the competing ones because our CD and P2S errors are significantly smaller than all the other compared methods.
Then we further compare our method to those learning-based methods (EC-Net, DMR, PCN, and PF) in the test set and show the evaluation results in Table \ref{tab:datasetCDerror}. 
Our method (PCA) takes low-quality initial normals estimated by PCA as input, and our method (AdaFit) takes the better normals estimated by AdaFit as input.
We observe that, except for under the noise level 2.5\%, our method (PCA) produces the second-best results on CD and P2S metrics. Moreover, our method (AdaFit) achieves the best results on both metrics in most cases, which show the superiority of our method.

\begin{table*}[htp] 
\footnotesize
	\centering 
	\caption{Quantitative comparisons for learning-based denoising methods on the synthetic dataset. Bold represents the best results, and underline denotes the second-best results.}

\begin{tabular}{ccccccc}
\toprule
 &\multicolumn{6}{l}{CD ($\times 10^{-5}$), P2S ($\times 10^{-3}$)} \\
\cmidrule(l){2-7}
Noise Level                & EC-Net       & DMR          & PCN          & PF          & Ours (PCA)         &  Ours (AdaFit) \\
\midrule
0.25\%                     & 0.67, 0.88  & 2.12, 2.16  & 0.96, 1.15  & 0.70, {0.55} & \underline{0.61}, \underline{0.52}  & \textbf{0.60}, \textbf{0.43}  \\
0.5\%                      & 1.08, 1.61  & 2.33, 2.32  & 1.33, 1.48  & 0.92, 0.78 & \underline{0.80}, \textbf{0.65}   & \textbf{0.78}, \underline{0.68}  \\
1\%                        & 4.47, 4.56  & 3.49, 3.27  & 2.11, 2.25  & 1.62, 1.36 & \underline{1.31}, \underline{1.30}   & \textbf{1.24}, \textbf{1.24}  \\
1.5\%                      & 12.90, 8.38  & 5.70, 4.90   & 3.73, 3.54  & 2.77,{2.32} & \underline{2.76},  \underline{2.05}  & \textbf{1.86}, \textbf{1.88}  \\
2.5\%                      & 51.05, 16.66 & 19.96, 9.97 & 20.16, 8.92 & \underline{5.11}, \underline{3.40} & 5.62, 3.73  & \textbf{4.96}, \textbf{3.38}    \\ 
\bottomrule
\end{tabular} \label{tab:datasetCDerror}
\end{table*}

\textbf{Computational time.} We also record the running time of all the tested methods in Table \ref{tab:denoisedCDErrors}.
As seen, DMR is the fastest method, while PCN is the slowest one.
Our method ranks fourth in speed among the learning-based methods.
In particular, the traditional methods (WLOP, RIMLS) require trial-and-error efforts to tune parameters to produce satisfactory results in practice; thus, we only discuss the inference time of the deep learning methods (EC-Net, DMR, PCN, PF, and ours).
DMR and EC-Net are faster than the other deep learning methods (PCN, PF, and ours).
The reason is that DMR and EC-Net have an analogous upsampling process and only use a few patches of the noisy point cloud as input.
In contrast, the PCN, PF, and our method use a pointwise manner to deal with the noisy input; thus, the inference times of these three methods are higher.
Moreover, due to the network complexity, our method takes less time than the PCN and more time than the PF.
In summary, although the inference time of our method seems to be slightly computationally intensive, it can generate more appealing results in terms of visual quality and error metrics in most examples.

\begin{figure*}[htb]
		\centering
		\captionsetup[subfloat]{labelsep=none,format=plain,labelformat=empty}
		\subfloat[(a)\; Jet]{\label{syn-normal-a}\includegraphics[width=0.183\textwidth]{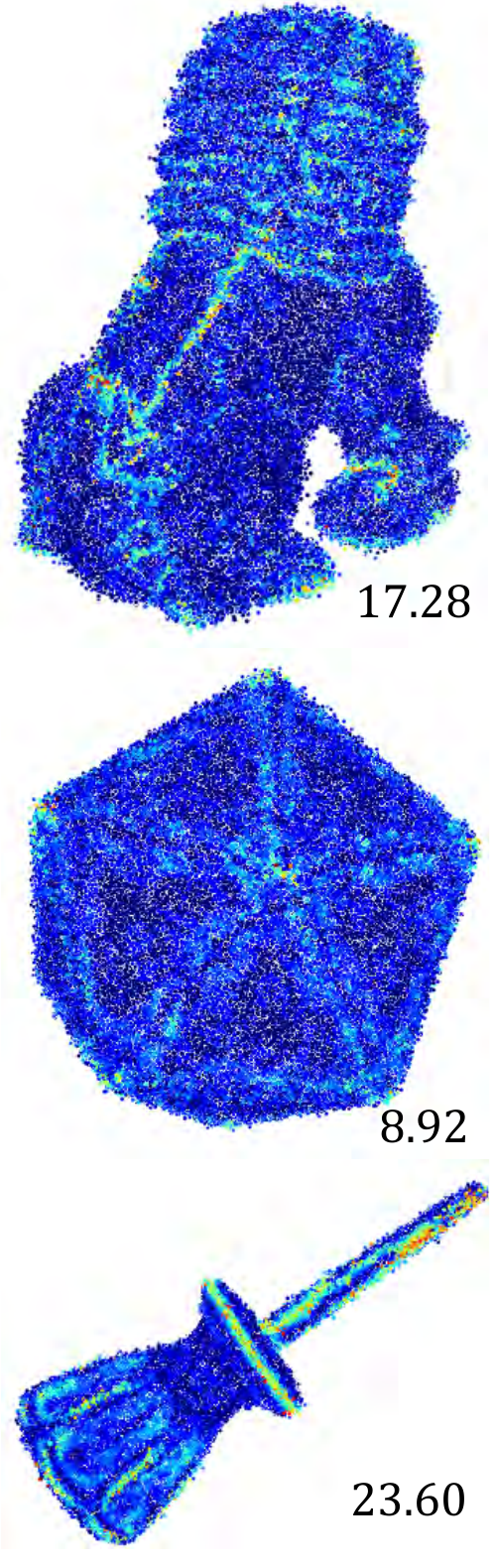}}
		\subfloat[(b)\;PCPNet]{\label{syn-normal-b}\includegraphics[width=0.183\textwidth]{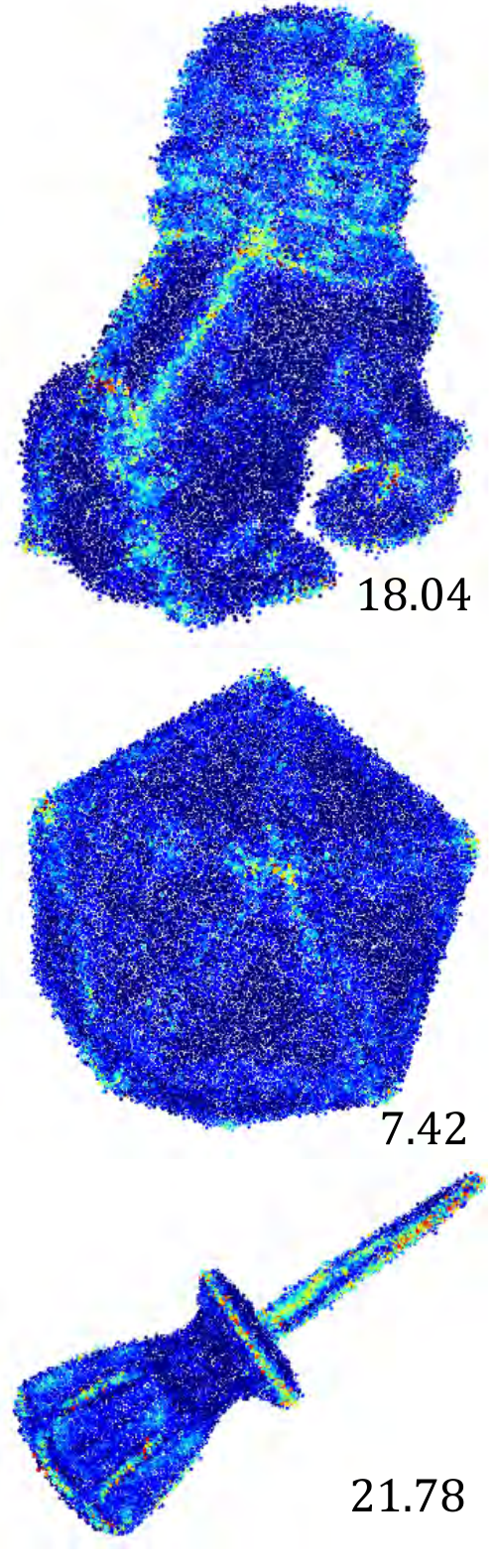}}
		\subfloat[(c)\;DeepFit]{\label{syn-normal-c}\includegraphics[width=0.183\textwidth]{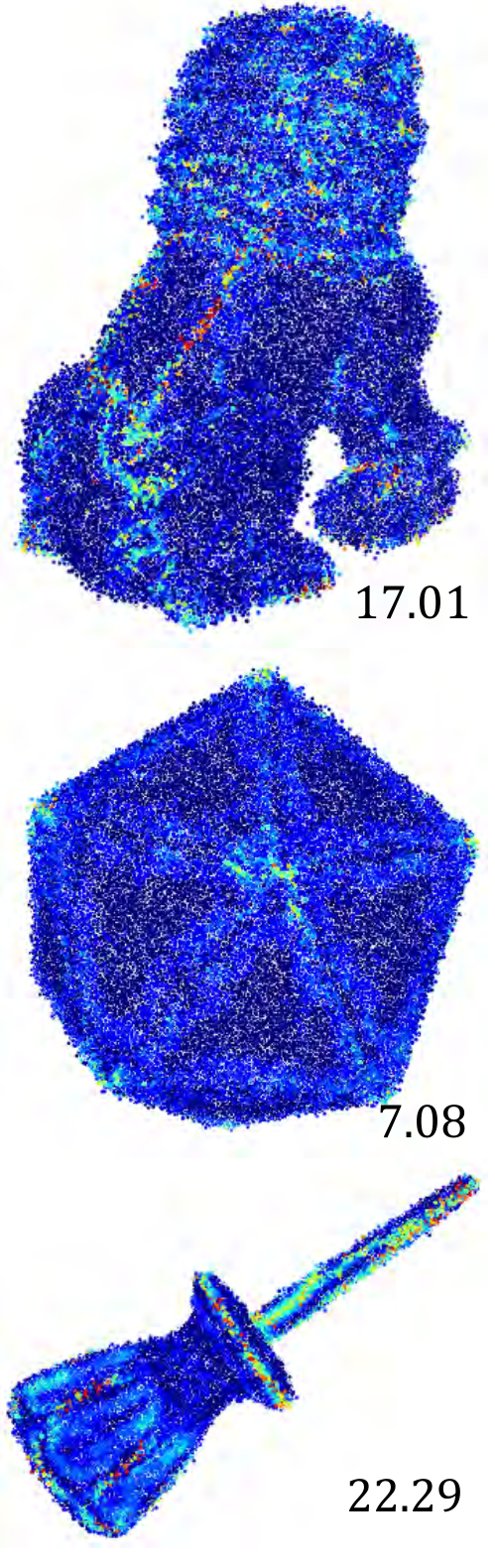}}
		\subfloat[(d)\;AdaFit]{\label{syn-normal-d}\includegraphics[width=0.183\textwidth]{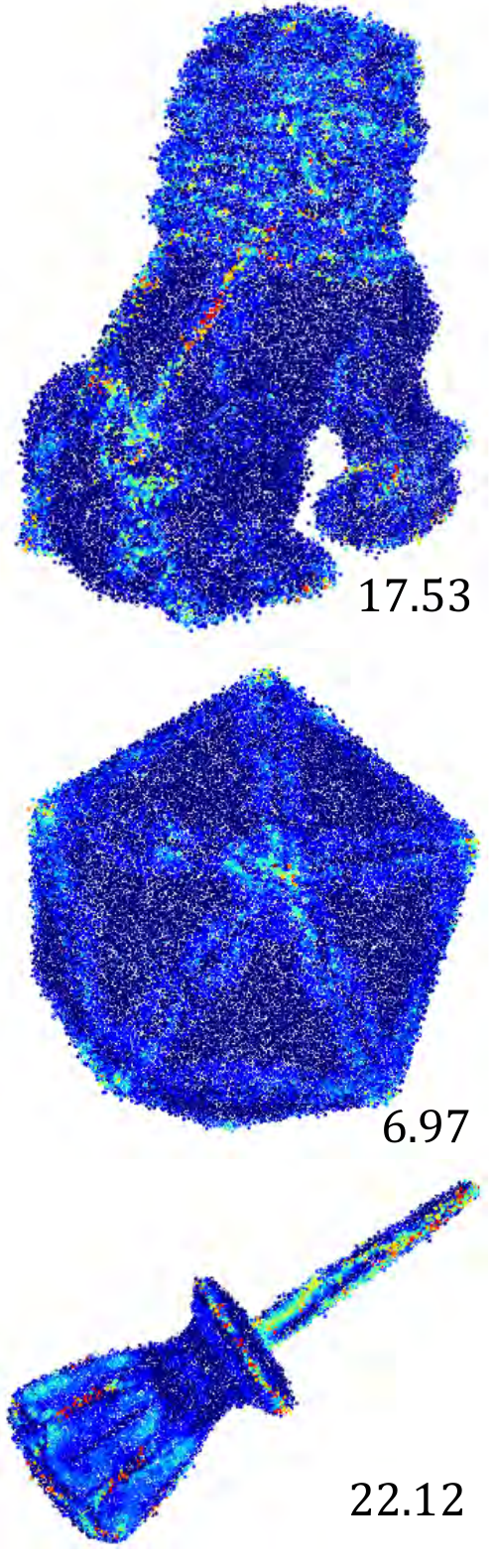}}
		\subfloat[(e)\;Ours]{\label{syn-normal-e}\includegraphics[width=0.183\textwidth]{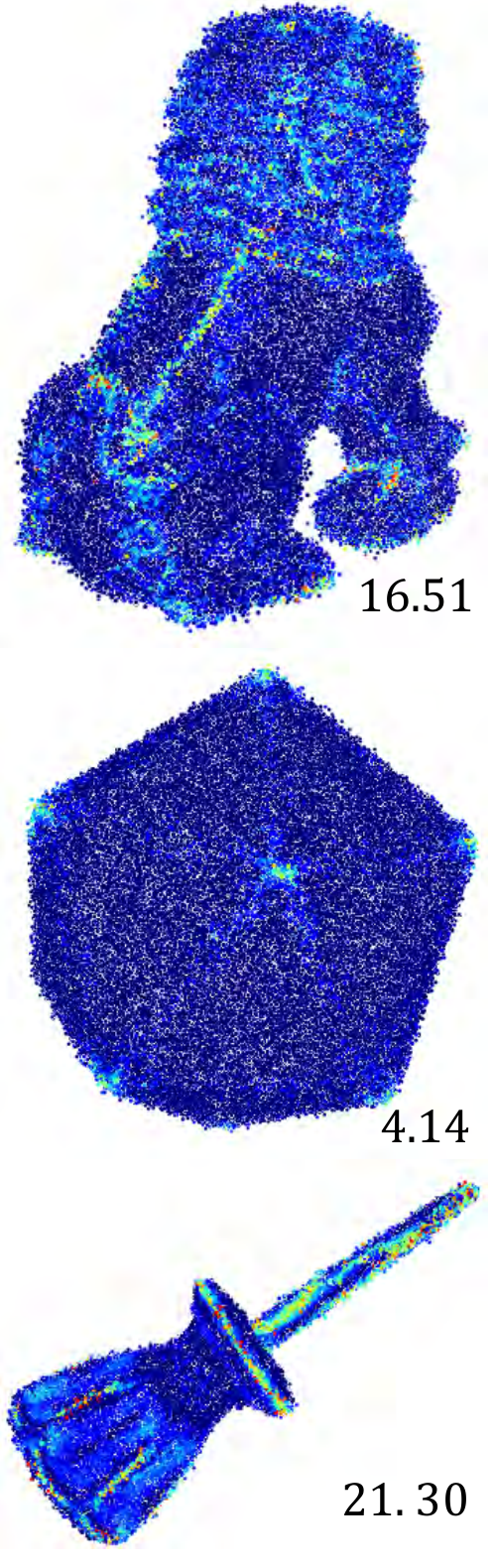}}
		\subfloat[]{\label{syn-normal-e}\includegraphics[width=0.05\textwidth]{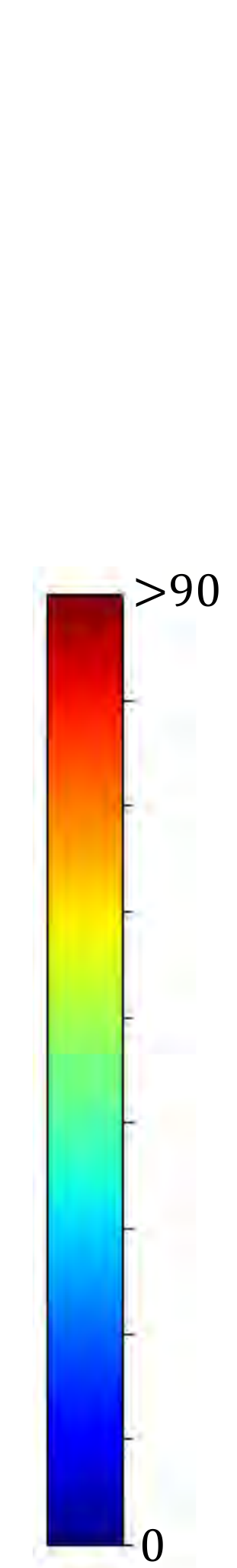}}
		\caption{Visualization of normal error for synthetic data with 0.5\% noise. From left to right: Jet, PCPNet, DeepFit, AdaFit, and our method. The point clouds are color-coded based on angular difference, with a color map given by the color bar on the right. The numerical value denotes RMSE ($\times 10^{-3}$), and a lower error is better.}
		\label{fig:syn-normal}
\end{figure*}
	
\begin{figure*}[htb]
		\centering
		\subfloat[RGB image]{\label{nyu-a}\includegraphics[width=0.164\textwidth]{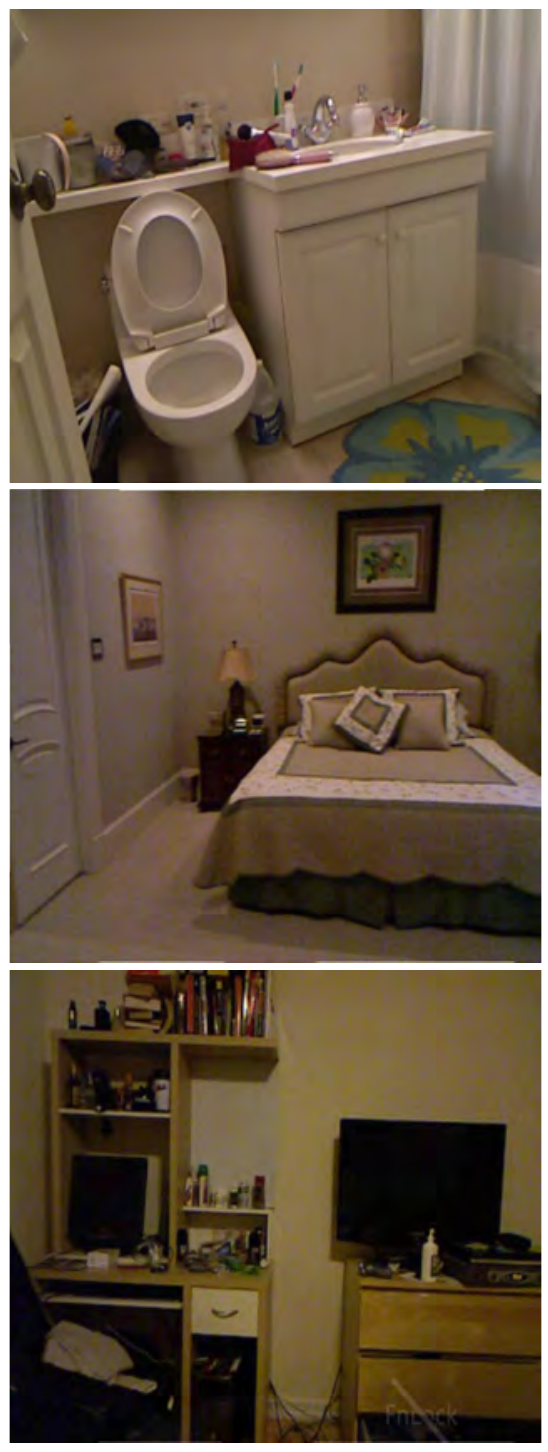}}
		\subfloat[Jet]{\label{nyu-b}\includegraphics[width=0.164\textwidth]{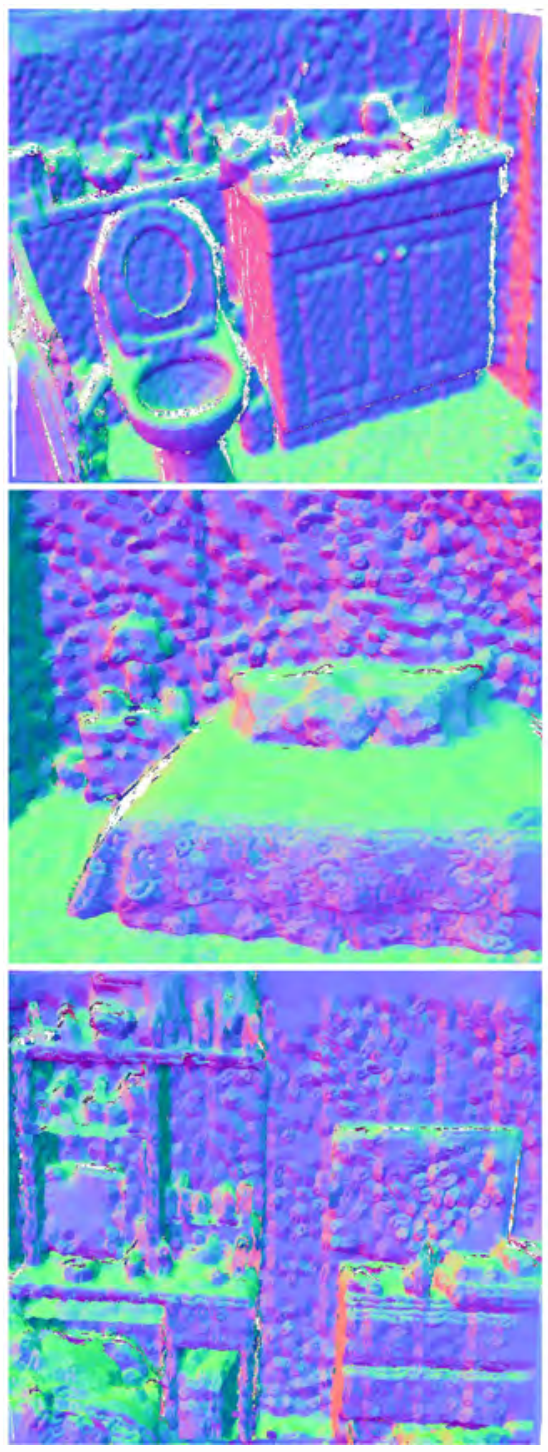}}
		\subfloat[PCPNet]{\label{nyu-c}\includegraphics[width=0.164\textwidth]{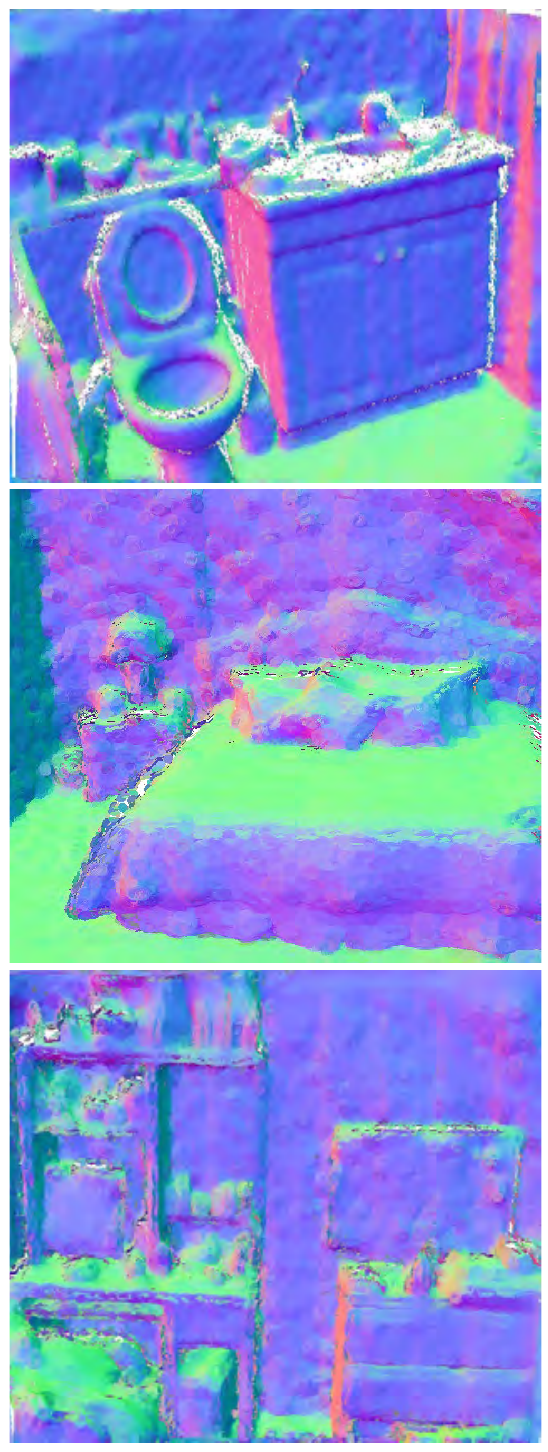}}
	\subfloat[DeepFit]{\label{nyu-d}\includegraphics[width=0.164\textwidth]{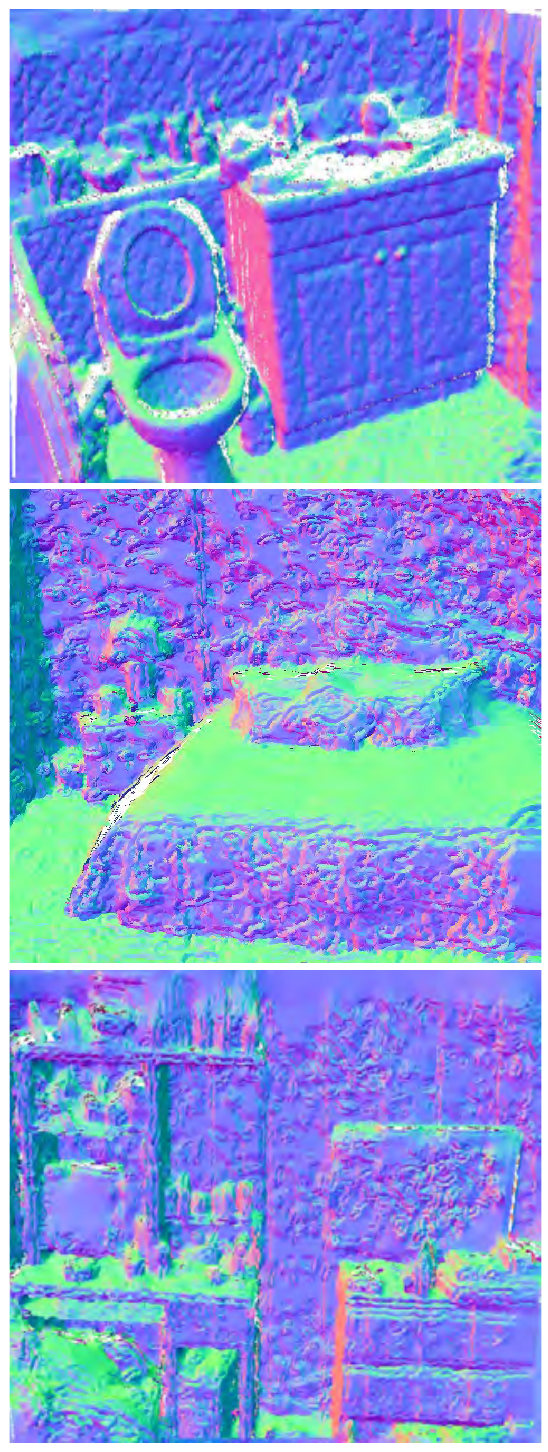}}
		\subfloat[AdaFit]{\label{nyu-e}\includegraphics[width=0.164\textwidth]{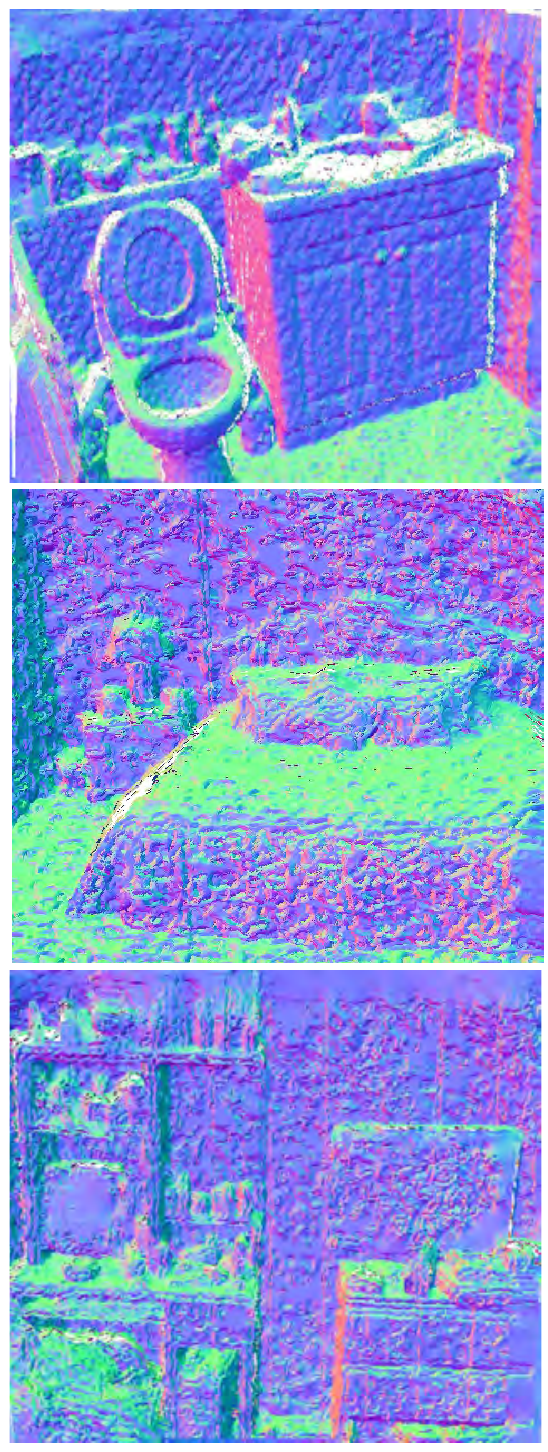}}
		\subfloat[Ours]{\label{nyu-f}\includegraphics[width=0.164\textwidth]{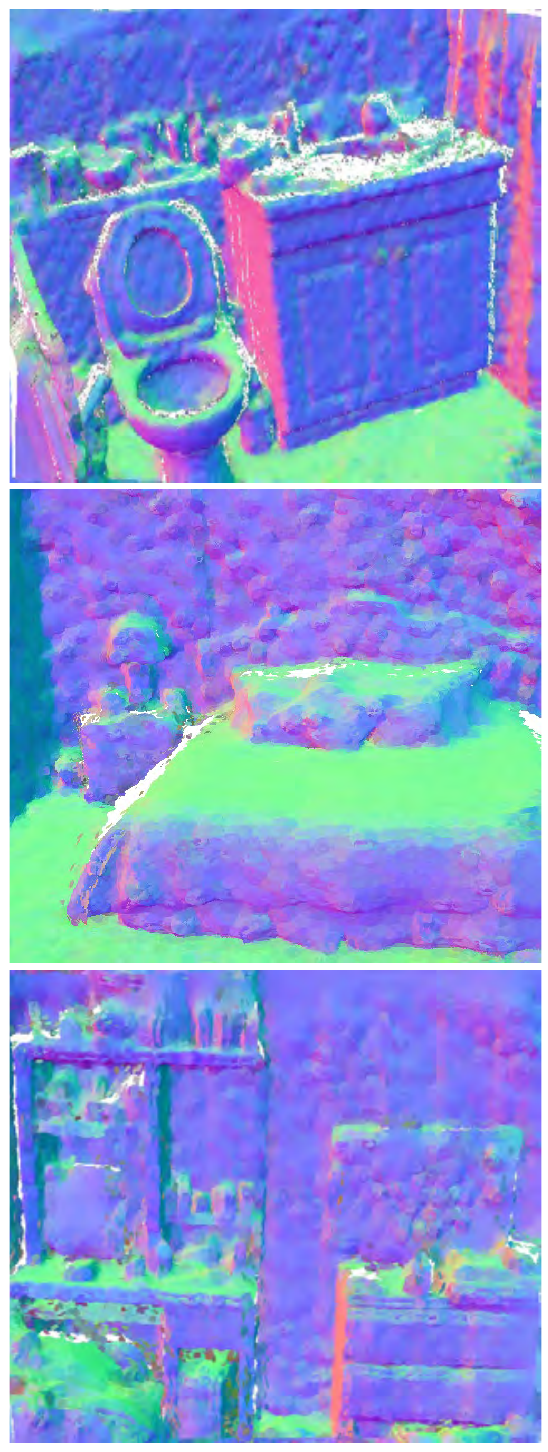}}
		\caption{Visual comparison of normal estimation results on scanned point clouds from the NYU Depth V2 dataset\cite{silberman2012indoor}. From left to right: RGB image (noisy input), Jet, PCPNet, DeepFit, AdaFit, and our method.
		}
		\label{fig:nyu}
\end{figure*}

\subsection{Experiments for Normal Filtering}
The normal filtering task also plays an important role in our method.
We compare our method with state-of-the-art normal filtering methods, including the traditional method of Jet \cite{cazals2005estimating} and the deep learning methods of PCPNet \cite{guerrero2018pcpnet}, DeepFit \cite{ben2020deepfit} and AdaFit \cite{zhu2021adafit}. We retrain PCPNet, DeepFit, and AdaFit over our training set. For PCPNet and AdaFit, we retrain the multi-scale version of the network provided by the authors.

\textbf{Qualitative comparisons.} Fig. \ref{fig:syn-normal} visualizes the error map for normals, where the error is defined as the angular deviations between the filtered normals and the ground truth normals.
As we can see, Jet and PCPNet produce higher errors in geometric feature regions. DeepFit can deal with smoothly curved regions well, but it blurs small-scale features.
AdaFit effectively recovers small-scale features and fine details but slightly oversmoothes sharp features.
In contrast, our method produces the desired results in terms of preserving sharp and detailed features.
To test the generalization capability of our method, we also present Fig. \ref{fig:nyu} for comparisons between our method and four other methods applied to real indoor scenes acquired by Kinect sensors (NYU Depth V2 dataset \cite{silberman2012indoor}).
Note that we do not train all the tested methods on scanned scenes.
For a fair comparison, we only use the filtered normals produced by our method instead of retaining denoised point coordinates.
As we can see in Fig. \ref{fig:nyu}, Jet retains considerable noise in the results.
PCPNet can smooth noisy surfaces effectively but flatten medium- and small-scale geometric features.
DeepFit and AdaFit preserve geometric features well even for small-scale and detailed features, although they induce some bumps in the results.
The reason for inducing the bumpy artifacts may be as follows.
Since both noise and geometric details are high-frequency information, DeepFit and Adafit may erroneously restore some high level noise as geometric features.
Compared to DeepFit and AdaFit, our method can better preserve structure features, although some geometric details are flattened slightly.
Moreover, our method tends to produce visually cleaner results without noticeable artifacts.

\begin{table}[htp]\footnotesize
	\centering
	\caption{Quantitative comparisons for normal estimation for classical geometric and learning-based methods. Bold represents the best results, and underline denotes the second-best results.}
\resizebox{\linewidth}{!}{
\begin{tabular}{cccccccc}
		\toprule
		&\multicolumn{7}{l}{RMSE ($\times 10^{-3}$)}\\
		
		\cmidrule(l){2-8}
		Noise level& PCA& Jet & PCPNet & DeepFit & AdaFit& Ours & Ours\\

            & &  &  &  & & (PCA) & (AdaFit)\\
  
		\midrule
		0.25\% & 13.48 & 13.10& 12.59 & 11.69 & 11.09 & \underline{10.97} & \textbf{10.68}\\
		0.5\% & 16.54 & 16.29 & 15.48 & 15.01 & 14.92 & \underline{14.78}  & \textbf{14.27}\\
		1\% & 25.82 & 22.96&21.72 & 21.54 & \underline{21.15} & 21.38  & \textbf{19.36}\\
		1.5\% & 32.87 &28.58&26.66 &27.25 & \underline{25.76} & 26.04  & \textbf{25.41}\\
		2.5\% & 42.95 &37.97&35.86 &36.29 & \underline{32.81} & 33.32  & \textbf{31.67}\\
  \bottomrule
	\end{tabular}} \label{tab:RMSE}
\end{table}

\textbf{Quantitative Comparisons.}
We adopt the root mean squared error of the angle difference (abbreviated as RMSE) to quantify normal filtering results \cite{guerrero2018pcpnet}, \cite{zhu2021adafit}.
Lower RMSE values indicate better results.
The results of our method (PCA) have the lowest RMSE values for all the tested examples, as shown in Fig. \ref{fig:syn-normal}. We also compare our method with the other four on the test set introduced in subsection \ref{subsec:experimentalSettings}.
Our method (PCA) can produce the second-best results when the noise is low and moderate (0.25\%, 0.5\% noise).
AdaFit achieves the second-best results as the noise level increases (1.0\%, 1.5\%, and 2.5\% noise), but our results are comparable to those of AdaFit. Furthermore, by using better initial normals as input, our method (AdaFit) can significantly outperform the other compared methods and achieves the smallest RMSE values at all noise levels, as shown in Table \ref{tab:RMSE}.

\begin{table}[htp]
	\centering    \footnotesize
	\caption{Quantitative comparisons of DR, CL, and our PCDNF. We list CD and P2S for the denoising task and RMSE for the normal filtering task.}

 %实际结果
\begin{tabular}{cccc}
\toprule
 &\multicolumn{3}{l}{CD ($\times 10^{-5}$), P2S ($\times 10^{-3}$); RMSE ($\times 10^{-3}$)} \\
\cmidrule(l){2-4}
Noise Level                & DR       & CL          & Ours (PCA)  \\
\midrule
0.25\%                     & \enspace 1.03,  0.54; --  &  0.64, \textbf{0.48}; 12.65  & \textbf{0.61}, 0.52; \textbf{10.97} \\
0.5\%                      & \enspace 1.28,  0.82; --  &  0.85, 0.73; 15.51  & \textbf{0.80}, \textbf{0.65}; \textbf{14.78}       \\
1\%                        & \enspace 1.92,  1.46; --  &  1.36, 1.32; 21.40  & \textbf{1.31}, \textbf{1.30}; \textbf{21.38}     \\
1.5\%                      & \enspace 3.22,  2.31; --  & 2.89, 2.54; \textbf{25.45}   & \textbf{2.76}, \textbf{2.05}; 26.04     \\
2.5\%                      & 10.66, 6.12; -- & 6.27, 4.14; \textbf{32.14} & \textbf{5.62}, \textbf{3.73}; 33.32     \\ 
\bottomrule
\end{tabular}
	\label{tab:extra_compare}
\end{table}

\begin{table}[htp]
	\centering    \footnotesize
	\caption{Ablation analysis: quantitative comparisons of different network versions. For each version,
 we list CD for the denoising task and RMSE for the normal filtering task.}
\resizebox{\linewidth}{!}{	
\begin{tabular}{cccccc}
\toprule
 &\multicolumn{5}{l}{CD ($\times 10^{-5}$), RMSE ($\times 10^{-3}$)} \\
\cmidrule(l){2-6}
Noise Level                & \emph{V1}       & \emph{V2}          & \emph{V3}          & \emph{V4}          & Full         \\
\midrule
0.25\%                     & 0.64, 15.27  & 0.62, 13.42  & 0.63, --  & --, \textbf{10.95} & \textbf{0.61}, 10.97  \\
0.5\%                      & 0.87, 18.15  & 0.81, 16.30  & 0.82, --  & --, 15.18 & \textbf{0.80}, \textbf{14.78}     \\
1\%                        & 1.52, 22.81  & 1.33, 21.58  & 1.39, --  & --, 21.60 & \textbf{1.31}, \textbf{21.38}    \\
1.5\%                      & 2.99, 27.84  & 2.88, 26.62   & 2.80, --  & --, 26.68 & \textbf{2.76},  \textbf{26.04}   \\
2.5\%                      & 7.23, 35.41 & 6.46, 34.53 & 6.40, -- &  --, 34.58  & \textbf{5.62}, \textbf{33.32}    \\ 
\bottomrule
\end{tabular}}
	\label{tab:ablationStuides}
\end{table}

\subsection{PCDNF versus DR}
To further demonstrate the effectiveness of PCDNF, we compare it to DR, a method recently proposed by Chen et al. \cite{huwei_deep_resampling}. 
DR uses gradient fields to model the distribution of degraded point clouds, which can predict the gradient field over the point cloud that converges points toward the underlying surface. 
Quantitative comparisons of results produced by DR and our method are listed in Table \ref{tab:extra_compare}.
As we can see, our method yields lower CD and P2S values than the competing method DR for the denoising task, indicating that our results are more faithful to the ground truth.

\subsection{PCDNF versus CL}
During the final stages of the paper's completion, we became aware of a concurrent work proposed by Edirimuni et al. \cite{Lu_contrastive}, which also addresses the challenges of point cloud denoising and normal filtering. Although both methods share similar goals, they differ in approaches: CL uses a contrastive learning mechanism to tackle both tasks, whereas PCDNF takes a multitask perspective. Specifically, PCDNF utilizes two network branches for denoising and normal filtering, respectively. This design allows the two tasks to benefit each other mutually. In contrast, CL uses only a single network branch to predict the combined vector (point position offset and normal vector) for denoising points and smoothing normals. Therefore, CL does not consider the association and interaction of the two tasks (denoising and normal filtering), which is the main difference between our method and CL.

Table \ref{tab:extra_compare} presents the evaluation results of our method and CL using the CD and RMSE metrics to assess the performance of denoising and normal filtering. We observe that our method outperforms CL in the denoising task for most cases except for the 0.25\% noise level. For the normal filtering task, our method also achieves better results at the noise levels 0.25\%, 0.5\%, and 1.0\%. However, CL achieves lower MSAE values when the noise level is high, specifically at 1.5\% and 2.5\%. Therefore, we believe that PCDNF and CL are complementary approaches, each offering unique advantages within their respective paradigms.

\subsection{Ablation Studies}
We verify the individual contributions of the major modules in our network by conducting the following four ablation studies. 
\begin{itemize}
	\item \emph{Removing feature selection and refinement modules (V1)}.
	\item \emph{Removing feature refinement module (V2)}.
	\item  \emph{The normal filtering network branch is removed (V3)}.
       \item  \emph{The denoising network branch is removed (V4)}.
\end{itemize}

For each ablated variant, we evaluate it in the validation set to choose the best hyperparameters. We perform quantitative comparisons of our method and four ablated variants in the test set and record the evaluation in Table \ref{tab:ablationStuides}.
From the table, we have the following observations. All four variants show lower performance than our full pipeline. Each pipeline module is necessary to ensure high-quality denoising and normal filtering results.
More specifically, by comparing \emph{V1} with \emph{V2}, it can be seen that the feature selection module is necessary for removing noise by selecting those feature points with similar characteristics to the denoised point.
By comparing \emph{V2} with our full pipeline, we can see that the additional feature refinement module positively impacts denoising accuracy.
The feature refinement module consists of two units (feature augmentation and fusion).
The roles of these two units are demonstrated in Figs. \ref{fig:featureAugmentation} and \ref{fig:featureFusion}, and are explained in subsection \ref{subsec:featureRefinement}.
We use a two-branch network structure for point cloud denoising and normal filtering tasks. 
To demonstrate the positive interaction between these two tasks, we design two variants (\emph{V3} and \emph{V4}) that perform only the denoising task and normal estimation task.
As we can see, our full network can achieve the best quantitative results in terms of CD and RMSE. Specifically, comparing variant \emph{V3} with our full network, our additional normal filtering branch helps our method produce the best denoising results. Comparing variant \emph{V4} with our full network, our denoising branch can help our method yield the best normal filtering results. The above ablation studies confirm the effectiveness of the major modules in our network and the mutual promotion of the two tasks (point cloud denoising and normal filtering).

\begin{figure*}[htb]
	\centering
	\subfloat{}\includegraphics[width=1.0\textwidth]{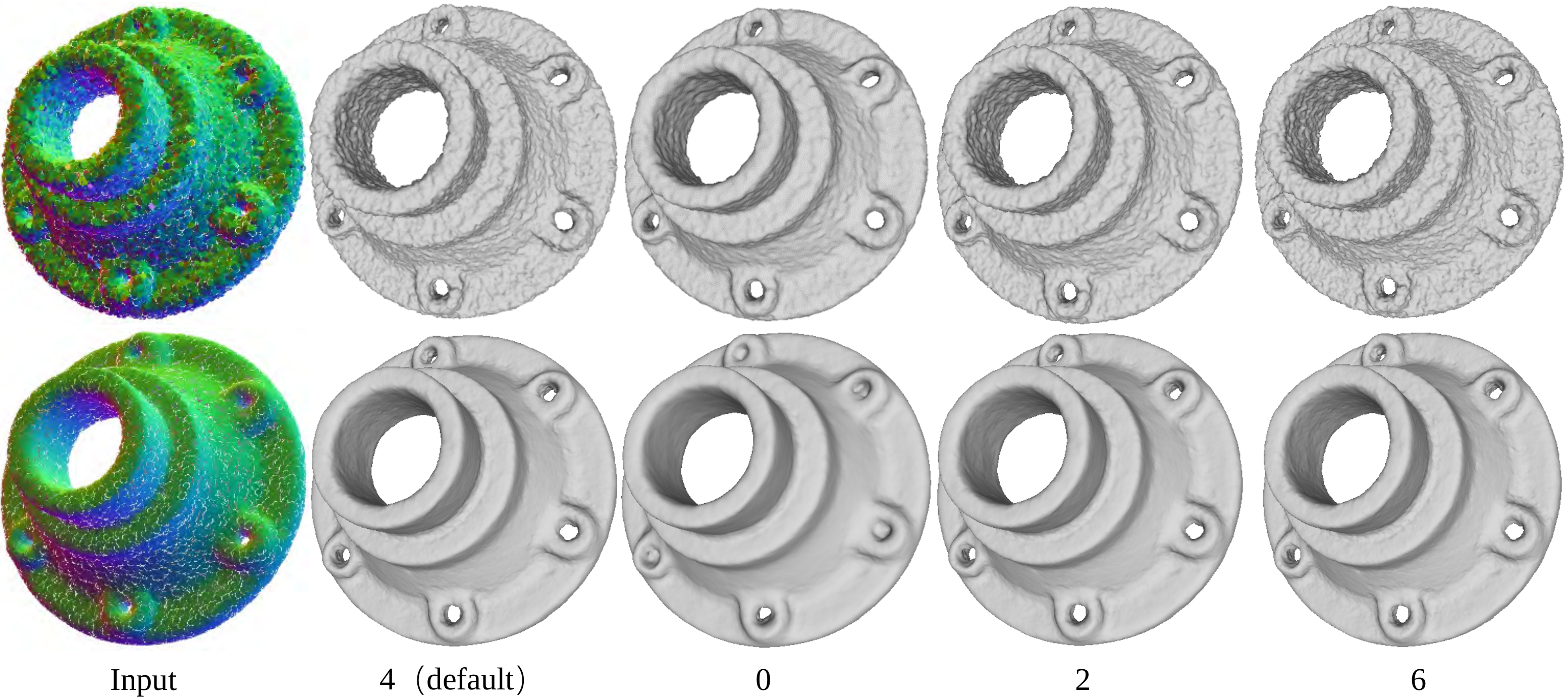}
	\caption{The screened Poisson surface reconstruction (sPSR) algorithm balances smoothness and accuracy via an interpolation weight $\alpha$. The top row shows the surfaces produced by applying sPSR directly to the noisy input with weight $\alpha=4$, $0$, $2$, $6$. The second row shows the reconstructed surfaces, which are computed by applying sPSR (with the same interpolation weight as the top row) to the denoised input. Our reconstructed surfaces are not sensitive to the perturbation of $\alpha$.
	}
	\label{fig:application1}
\end{figure*}

\begin{figure}[htb]
	\centering
	\subfloat[]{\label{application2-a}\includegraphics[width=0.16\textwidth]{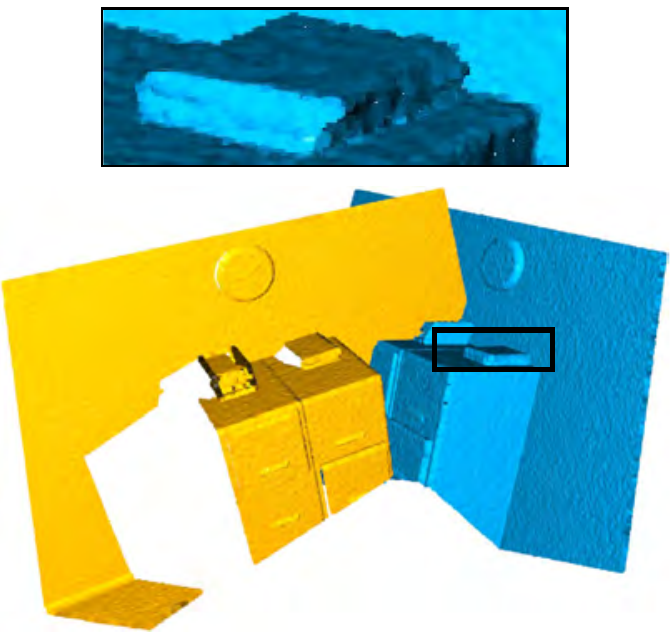}}
	\subfloat[]{\label{application2-b}\includegraphics[width=0.16\textwidth]{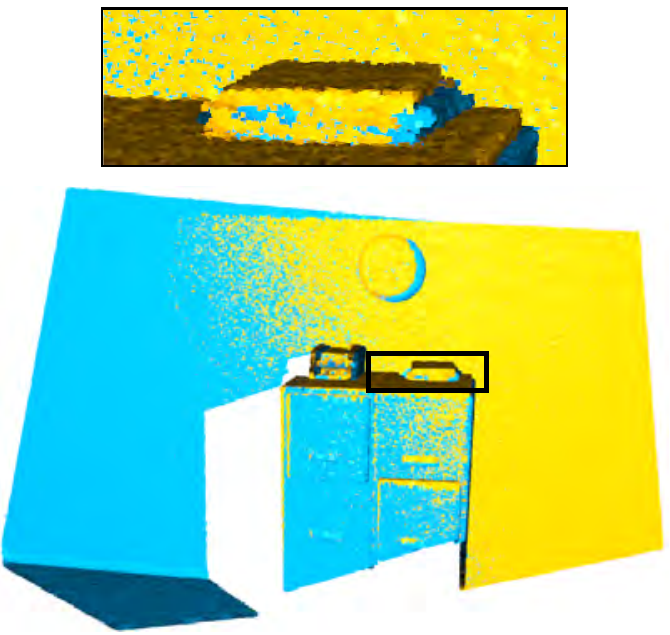}}
	\subfloat[]{\label{application2-c}\includegraphics[width=0.16\textwidth]{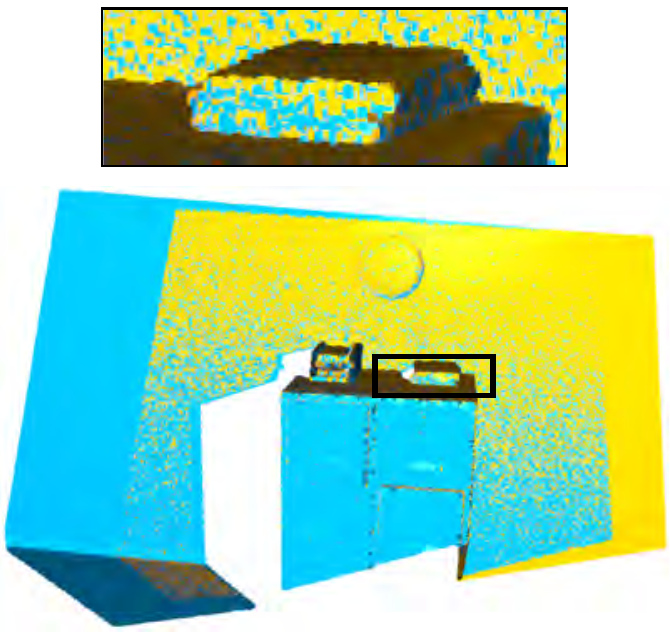}}
	\caption{Application of point cloud registration. (a) Noisy input. (b) Registration result of (a). (c) Registration result after applying our denoising method to (a). }
	\label{fig:application2}
\end{figure}

\begin{figure}[htb]
	\centering
	\subfloat[]{\label{application3-a}\includegraphics[width=0.16\textwidth]{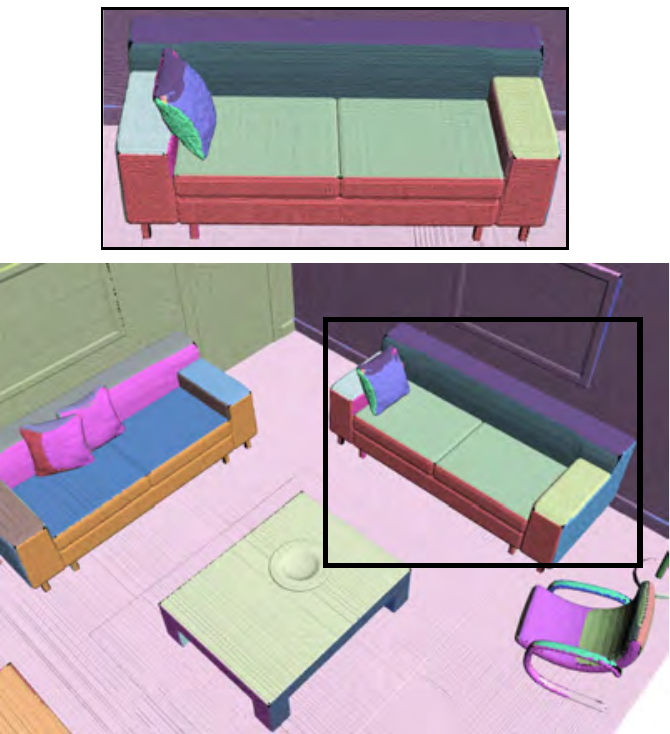}}
	\subfloat[]{\label{application3-b}\includegraphics[width=0.16\textwidth]{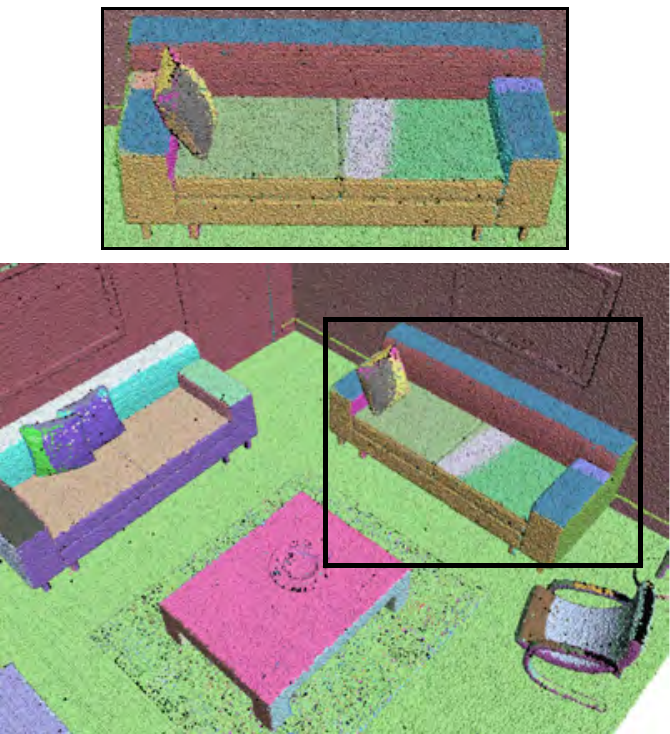}}
	\subfloat[]{\label{application3-c}\includegraphics[width=0.16\textwidth]{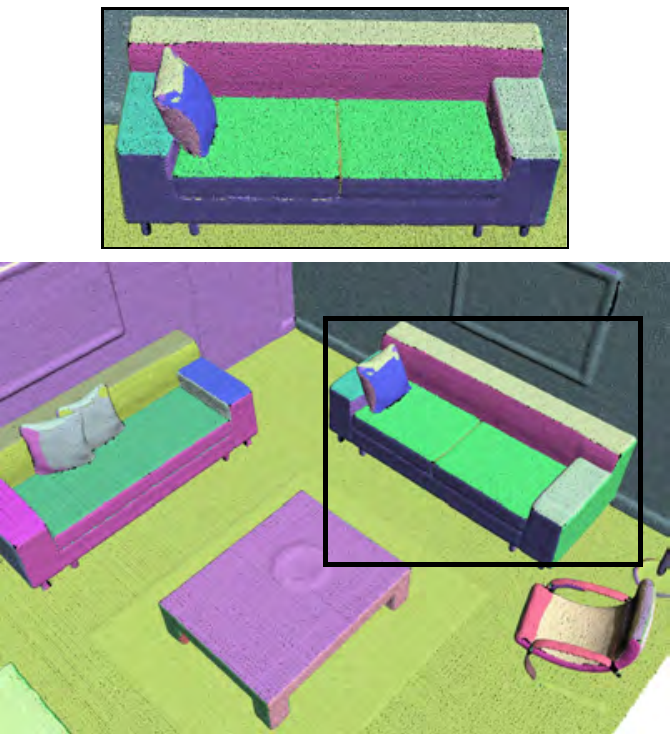}}
	\caption{Application of RANSAC plane fitting. (a) Clean point cloud. (b) The corresponding noisy point cloud. (c) Plane segmentation result produced after applying our method to (b).}
	\label{fig:application3}
\end{figure}

\subsection{Limitations}
To some extent, the quality of the initial normals has an impact on the results of our method. 
Our method (with PCA) outperforms AdaFit at lower noise levels but is less effective at higher noise levels, as Table \ref{tab:RMSE} shows. The reason for this is that our method prioritizes denoising and treats normal estimation as a secondary task. Consequently, our subnet for smoothing raw normals lacks the complexity needed to accurately estimate normals in the presence of large noise. In order to maintain the simplicity and generality of input, we choose to use raw normals estimated by PCA, which can negatively affect the performance of our method in the presence of high noise levels. Nonetheless, due to our multitask and iteration mechanisms, our method can incrementally enhance denoising and normal filtering results, making them comparable to state-of-the-art methods. Furthermore, when better initial normals (estimated by AdaFit) are used as input, our method can concurrently achieve the best denoising and normal filtering results, as demonstrated in Tables \ref{tab:datasetCDerror} and \ref{tab:RMSE}.

\subsection{Applications}
To demonstrate the effectiveness of our method across  various applications, we utilize our denoised outputs as inputs for surface reconstruction \cite{kazhdan2013screened}, point cloud registration \cite{zhou2016fast}, and plane fitting \cite{schnabel2007efficient} tasks.
In Fig. \ref{fig:application1},
we observe that the reconstruction results from noisy point clouds exhibit significant artifacts, whereas those from our denoised input are high-quality and preserve features.
Fig. \ref{fig:application2} demonstrates that our denoising method can enhance the accuracy of the point cloud registration task, as outputs after applying our method are more precise than those from raw inputs.
We perform plane fitting \cite{schnabel2007efficient} on an indoor scene point cloud. As Fig. \ref{fig:application3} shows,  using our denoised output as input for the RANSAC algorithm \cite{schnabel2007efficient} results in a more reliable plane segmentation outcome, which is closer to the segmentation outcome from the corresponding clean point cloud.

\section{Conclusion} \label{sec:conclusion}
We have proposed a learning method for denoising point clouds via joint normal filtering. Our key insight is that denoising and normal filtering tasks are inseparably intertwined. Our method takes the noisy point cloud and corresponding initial normals as input and uses an end-to-end approach to predict denoised points and filtered normals simultaneously. Our method comprises two innovative modules: the shape-aware selector and feature refinement.
The shape-aware selector reduces the negative effects of noise and outliers on feature learning, thereby enhancing denoising and filtering performance. The feature refinement has advantages in recovering structure and detailed features. Our experiments demonstrate that incorporating normal filtering improves the denoising performance significantly. Our method achieves a new SOTA for the denoising task, with substantial improvements in visual quality and quantitative evaluation. Although our method is not specifically designed for normal filtering, it performs favorably against most SOTA normal filtering methods.

To our knowledge, this is the first work to couple the interdependent tasks of point cloud denoising and normal filtering within a single deep neural network. We believe that there is a wealth of opportunity for exploring future directions. 
For instance, we can improve the normal filtering subnet to obtain more accurate normal estimations, especially in situations with large noise. Moreover, we can further improve the subnet to address the normal orientation ambiguity problem.
As a pointwise denoising approach, our method incurs high computational costs for training and inference. To address this, we plan to develop a patchwise framework to improve runtime performance.

\section*{Acknowledgments}
Zheng Liu was supported by National Key R$\&$D Program of China (No. 2022YFB3904100).
Renjie Chen was partially supported by NSF of China (62072422) and NSF of Anhui Province, China (2008085MF195).
Ying He was partially supported by the Ministry of Education, Singapore, under its Academic Research Fund Grants (MOE-T2EP20220-0005 \& RT19/22).
% This work was supported by NSF of China (Nos. 62072422, 12001144, 62025207, 62076227, and 61702467), National Key R$\&$D Program of China (No. 2020YFC1523102), NSF of Anhui Province, China (No. 2008085MF195), Youth Science and Technology Foundation of Gansu (No. 20JR5RA050), NSF of Zhejiang Province, China (No. LQ20A010007), and Zhejiang Lab (No. 2019NB0AB03).

\bibliographystyle{IEEEtran}
\bibliography{references}

\begin{thebibliography}{10}
\providecommand{\url}[1]{#1}
\csname url@samestyle\endcsname
\providecommand{\newblock}{\relax}
\providecommand{\bibinfo}[2]{#2}
\providecommand{\BIBentrySTDinterwordspacing}{\spaceskip=0pt\relax}
\providecommand{\BIBentryALTinterwordstretchfactor}{4}
\providecommand{\BIBentryALTinterwordspacing}{\spaceskip=\fontdimen2\font plus
\BIBentryALTinterwordstretchfactor\fontdimen3\font minus
  \fontdimen4\font\relax}
\providecommand{\BIBforeignlanguage}[2]{{%
\expandafter\ifx\csname l@#1\endcsname\relax
\typeout{** WARNING: IEEEtran.bst: No hyphenation pattern has been}%
\typeout{** loaded for the language `#1'. Using the pattern for}%
\typeout{** the default language instead.}%
\else
\language=\csname l@#1\endcsname
\fi
#2}}
\providecommand{\BIBdecl}{\relax}
\BIBdecl

\bibitem{wei2021geodualcnn}
M.~Wei, H.~Chen, Y.~Zhang, H.~Xie, Y.~Guo, and J.~Wang,
  ``Geo\textsc{D}ual\textsc{CNN}: Geometry-supporting dual convolutional neural
  network for noisy point clouds,'' \emph{IEEE Trans. Vis. Comput. Graph.},
  2021.

\bibitem{chen2022repcd}
H.~Chen, Z.~Wei, X.~Li, Y.~Xu, M.~Wei, and J.~Wang,
  ``Re\textsc{PCD}-\textsc{N}et: Feature-aware recurrent point cloud denoising
  network,'' \emph{Int. J. Comput. Vision}, vol. 130, no.~3, pp. 615--629,
  2022.

\bibitem{zhang2020pointfilter}
D.~Zhang, X.~Lu, H.~Qin, and Y.~He, ``Pointfilter: Point cloud filtering via
  encoder-decoder modeling,'' \emph{IEEE Trans. Vis. Comput. Graph.}, vol.~27,
  no.~3, pp. 2015--2027, 2021.

\bibitem{oztireli2009feature}
A.~C. {\"O}ztireli, G.~Guennebaud, and M.~Gross, ``Feature preserving point set
  surfaces based on non-linear kernel regression,'' \emph{Comput. Graph.
  Forum}, vol.~28, no.~2, pp. 493--501, 2009.

\bibitem{lu2017gpf}
X.~Lu, S.~Wu, H.~Chen, S.-K. Yeung, W.~Chen, and M.~Zwicker, ``\textsc{GPF}:
  \textsc{GMM}-inspired feature-preserving point set filtering,'' \emph{IEEE
  Trans. Vis. Comput. Graph.}, vol.~24, no.~8, pp. 2315--2326, 2017.

\bibitem{chen2019multi}
H.~Chen, M.~Wei, Y.~Sun, X.~Xie, and J.~Wang, ``Multi-patch collaborative point
  cloud denoising via low-rank recovery with graph constraint,'' \emph{IEEE
  Trans. Vis. Comput. Graph.}, vol.~26, no.~11, pp. 3255--3270, 2019.

\bibitem{chen2019structure}
H.~Chen, J.~Huang, O.~Remil, H.~Xie, J.~Qin, Y.~Guo, M.~Wei, and J.~Wang,
  ``Structure-guided shape-preserving mesh texture smoothing via joint low-rank
  matrix recovery,'' \emph{Comput-Aided Des.}, vol. 115, pp. 122--134, 2019.

\bibitem{lu2020low}
X.~Lu, S.~Schaefer, J.~Luo, L.~Ma, and Y.~He, ``Low rank matrix approximation
  for 3d geometry filtering,'' \emph{IEEE Trans. Vis. Comput. Graph.}, vol.~28,
  no.~4, pp. 1835--1847, 2020.

\bibitem{liu2020feature}
Z.~Liu, X.~Xiao, S.~Zhong, W.~Wang, Y.~Li, L.~Zhang, and Z.~Xie, ``A
  feature-preserving framework for point cloud denoising,'' \emph{Comput-Aided
  Des.}, vol. 127, p. 102857, 2020.

\bibitem{Liu2022MeshTGV}
Z.~Liu, Y.~Li, W.~Wang, L.~Liu, and R.~Chen, ``Mesh total generalized variation
  for denoising,'' \emph{IEEE Trans. Vis. Comput. Graph.}, vol.~28, no.~12, pp.
  4418--4433, 2022.

\bibitem{rakotosaona2020pointcleannet}
M.-J. Rakotosaona, V.~La~Barbera, P.~Guerrero, N.~J. Mitra, and M.~Ovsjanikov,
  ``Point\textsc{C}lean\textsc{N}et: Learning to denoise and remove outliers
  from dense point clouds,'' \emph{Comput. Graph. Forum}, vol.~39, no.~1, pp.
  185--203, 2020.

\bibitem{lu2020deep}
D.~Lu, X.~Lu, Y.~Sun, and J.~Wang, ``Deep feature-preserving normal estimation
  for point cloud filtering,'' \emph{Comput-Aided Des.}, vol. 125, p. 102860,
  2020.

\bibitem{wang2019dynamic}
Y.~Wang, Y.~Sun, Z.~Liu, S.~E. Sarma, M.~M. Bronstein, and J.~M. Solomon,
  ``Dynamic \textsc{g}raph \textsc{cnn} for learning on point clouds,''
  \emph{ACM Trans. Graph.}, vol.~38, no.~5, pp. 1--12, 2019.

\bibitem{zhou2022point}
L.~Zhou, G.~Sun, Y.~Li, W.~Li, and Z.~Su, ``Point cloud denoising review: from
  classical to deep learning-based approaches,'' \emph{Graph. Models}, vol.
  121, p. 101140, 2022.

\bibitem{alexa2003computing}
M.~Alexa, J.~Behr, D.~Cohen-Or, S.~Fleishman, D.~Levin, and C.~T. Silva,
  ``Computing and rendering point set surfaces,'' \emph{IEEE Trans. Vis.
  Comput. Graph.}, vol.~9, no.~1, pp. 3--15, 2003.

\bibitem{amenta2004defining}
N.~Amenta and Y.~J. Kil, ``Defining point-set surfaces,'' \emph{ACM Trans.
  Graph.}, vol.~23, no.~3, pp. 264--270, 2004.

\bibitem{fleishman2005robust}
S.~Fleishman, D.~Cohen-Or, and C.~T. Silva, ``Robust moving least-squares
  fitting with sharp features,'' \emph{ACM Trans. Graph.}, vol.~24, no.~3, pp.
  544--552, 2005.

\bibitem{lipman2007parameterization}
Y.~Lipman, D.~Cohen-Or, D.~Levin, and H.~Tal-Ezer, ``Parameterization-free
  projection for geometry reconstruction,'' \emph{ACM Trans. Graph.}, vol.~26,
  no.~3, p.~22, 2007.

\bibitem{huang2009consolidation}
H.~Huang, D.~Li, H.~Zhang, U.~Ascher, and D.~Cohen-Or, ``Consolidation of
  unorganized point clouds for surface reconstruction,'' \emph{ACM Trans.
  Graph.}, vol.~28, no.~5, pp. 176:1--7, 2009.

\bibitem{huang2013edge}
H.~Huang, S.~Wu, M.~Gong, D.~Cohen-Or, U.~Ascher, and H.~R. Zhang, ``Edge-aware
  point set resampling,'' \emph{ACM Trans. Graph.}, vol.~32, no.~1, pp.
  9:1--9:12, 2013.

\bibitem{preiner2014continuous}
R.~Preiner, O.~Mattausch, M.~Arikan, R.~Pajarola, and M.~Wimmer, ``Continuous
  projection for fast $\textsc{L}_1$ reconstruction.'' \emph{ACM Trans.
  Graph.}, vol.~33, no.~4, pp. 47--1, 2014.

\bibitem{avron2010}
H.~Avron, A.~Sharf, C.~Greif, and D.~Cohen-Or, ``$\ell_1$-sparse reconstruction
  of sharp point set surfaces,'' \emph{ACM Trans. Graph.}, vol.~29, no.~5, pp.
  135:1--12, 2010.

\bibitem{sun2015denoising}
Y.~Sun, S.~Schaefer, and W.~Wang, ``Denoising point sets via $\ell_0$
  minimization,'' \emph{Comput. Aided Geom. Des.}, vol.~35, pp. 2--15, 2015.

\bibitem{mattei2017point}
E.~Mattei and A.~Castrodad, ``Point cloud denoising via moving
  $\mathrm{RPCA}$,'' \emph{Comput. Graph. Forum}, vol.~36, no.~8, pp. 123--137,
  2017.

\bibitem{digne2017sparse}
J.~Digne, S.~Valette, and R.~Chaine, ``Sparse geometric representation through
  local shape probing,'' \emph{IEEE Trans. Vis. Comput. Graph.}, vol.~24,
  no.~7, pp. 2238--2250, 2017.

\bibitem{wang2022rethinking}
J.~Wang, J.~Jiang, X.~Lu, and M.~Wang, ``Rethinking point cloud filtering: A
  non-local position based approach,'' \emph{Comput-Aided Des.}, vol. 144, p.
  103162, 2022.

\bibitem{sun2022structure}
G.~Sun, C.~Chu, J.~Mei, W.~Li, and Z.~Su, ``Structure-aware denoising for
  real-world noisy point clouds with complex structures,'' \emph{Comput-Aided
  Des.}, vol. 149, p. 103275, 2022.

\bibitem{huweifeaturegraph}
W.~Hu, X.~Gao, G.~Cheung, and Z.~Guo, ``Feature graph learning for 3d point
  cloud denoising,'' \emph{IEEE Trans. Signal Process.}, vol.~68, pp.
  2841--2856, 2020.

\bibitem{qi2017pointnet}
C.~R. Qi, H.~Su, K.~Mo, and L.~J. Guibas, ``Point\textsc{N}et: Deep learning on
  point sets for 3d classification and segmentation,'' in \emph{Proceedings of
  the IEEE Conference on Computer Vision and Pattern Recognition (CVPR)}, 2017,
  pp. 652--660.

\bibitem{qi2017pointnet++}
C.~R. Qi, L.~Yi, H.~Su, and L.~J. Guibas, ``Point\textsc{N}et++: Deep
  hierarchical feature learning on point sets in a metric space,'' in
  \emph{Proceedings of the Annual Conference on Neural Information Processing
  Systems (NIPS)}, 2017, pp. 5100--5109.

\bibitem{liu2021deep}
S.-L. Liu, H.-X. Guo, H.~Pan, P.-S. Wang, X.~Tong, and Y.~Liu, ``Deep implicit
  moving least-squares functions for 3d reconstruction,'' in \emph{Proceedings
  of the IEEE Conference on Computer Vision and Pattern Recognition (CVPR)},
  2021, pp. 1788--1797.

\bibitem{Lu_IterativePFN}
D.~de~Silva~Edirimuni, X.~Lu, Z.~Shao, G.~Li, A.~Robles{-}Kelly, and Y.~He,
  ``{IterativePFN}: True iterative point cloud filtering,'' in
  \emph{Proceedings of the IEEE Conference on Computer Vision and Pattern
  Recognition (CVPR)}, 2023.

\bibitem{roveri2018pointpronets}
R.~Roveri, A.~C. {\"O}ztireli, I.~Pandele, and M.~Gross,
  ``Point\textsc{P}ro\textsc{N}ets: Consolidation of point clouds with
  convolutional neural networks,'' \emph{Comput. Graph. Forum}, vol.~37, no.~2,
  pp. 87--99, 2018.

\bibitem{yu2018ec}
L.~Yu, X.~Li, C.-W. Fu, D.~Cohen-Or, and P.-A. Heng,
  ``E\textsc{c}-\textsc{n}et: an edge-aware point set consolidation network,''
  in \emph{Proceedings of the European Conference on Computer Vision (ECCV)},
  2018, pp. 386--402.

\bibitem{luo2020differentiable}
S.~Luo and W.~Hu, ``Differentiable manifold reconstruction for point cloud
  denoising,'' in \emph{Proceedings of the ACM International Conference on
  Multimedia (ACM MM)}, 2020, pp. 1330--1338.

\bibitem{hermosilla2019total}
P.~Hermosilla, T.~Ritschel, and T.~Ropinski, ``Total \textsc{d}enoising:
  Unsupervised learning of 3d point cloud cleaning,'' in \emph{Proceedings of
  the IEEE International Conference on Computer Vision (ICCV)}, 2019, pp.
  52--60.

\bibitem{pistilli2020learning}
F.~Pistilli, G.~Fracastoro, D.~Valsesia, and E.~Magli, ``Learning
  graph-convolutional representations for point cloud denoising,'' in
  \emph{Proceedings of the European Conference on Computer Vision (ECCV)},
  2020, pp. 103--118.

\bibitem{luo2021score}
S.~Luo and W.~Hu, ``Score-based point cloud denoising,'' in \emph{Proceedings
  of the IEEE International Conference on Computer Vision (ICCV)}, 2021, pp.
  4583--4592.

\bibitem{huwei_deep_resampling}
H.~Chen, B.~Du, S.~Luo, and W.~Hu, ``Deep point set resampling via gradient
  fields,'' \emph{IEEE Trans. Pattern Anal. Mach. Intell.}, vol.~45, no.~3, pp.
  2913--2930, 2023.

\bibitem{Lu_contrastive}
D.~de~Silva~Edirimuni, X.~Lu, G.~Li, and A.~Robles{-}Kelly, ``Contrastive
  learning for joint normal estimation and point cloud filtering,'' \emph{IEEE
  Trans. Vis. Comput. Graph.}, 2023.

\bibitem{kazhdan2013screened}
M.~Kazhdan and H.~Hoppe, ``Screened poisson surface reconstruction,'' \emph{ACM
  Trans. Graph.}, vol.~32, no.~3, pp. 1--13, 2013.

\bibitem{hou2022iterative}
F.~Hou, C.~Wang, W.~Wang, H.~Qin, C.~Qian, and Y.~He, ``Iterative {P}oisson
  surface reconstruction (i\textsc{PSR}) for unoriented points,'' \emph{ACM
  Trans. Graph.}, vol.~41, no.~4, pp. 128:1--13, 2022.

\bibitem{rusu2009fast}
R.~B. Rusu, N.~Blodow, and M.~Beetz, ``Fast point feature histograms
  \textsc{(FPFH)} for 3d registration,'' in \emph{IEEE International Conference
  on Robotics and Automation (ICRA)}, 2009, pp. 3212--3217.

\bibitem{zhang2021fast}
J.~Zhang, Y.~Yao, and B.~Deng, ``Fast and robust iterative closest point,''
  \emph{IEEE Trans. Pattern Anal. Mach. Intell.}, vol.~44, no.~7, 2022.

\bibitem{guerrero2018pcpnet}
P.~Guerrero, Y.~Kleiman, M.~Ovsjanikov, and N.~J. Mitra, ``P\textsc{CPN}et
  learning local shape properties from raw point clouds,'' \emph{Comput. Graph.
  Forum}, vol.~37, no.~2, pp. 75--85, 2018.

\bibitem{zhou2020normal}
J.~Zhou, H.~Huang, B.~Liu, and X.~Liu, ``Normal estimation for 3d point clouds
  via local plane constraint and multi-scale selection,'' \emph{Comput-Aided
  Des.}, vol. 129, p. 102916, 2020.

\bibitem{ben2019nesti}
Y.~Ben-Shabat, M.~Lindenbaum, and A.~Fischer, ``Nesti-net: Normal estimation
  for unstructured 3d point clouds using convolutional neural networks,'' in
  \emph{Proceedings of the IEEE Conference on Computer Vision and Pattern
  Recognition (CVPR)}, 2019, pp. 10\,112--10\,120.

\bibitem{zhou2022fast}
J.~Zhou, W.~Jin, M.~Wang, X.~Liu, Z.~Li, and Z.~Liu, ``Fast and accurate normal
  estimation for point clouds via patch stitching,'' \emph{Comput.-Aided Des.},
  vol. 142, p. 103121, 2022.

\bibitem{ben2020deepfit}
Y.~Ben-Shabat and S.~Gould, ``Deep\textsc{f}it: 3d surface fitting via neural
  network weighted least squares,'' in \emph{Proceedings of the European
  Conference on Computer Vision (ECCV)}, 2020, pp. 20--34.

\bibitem{zhu2021adafit}
R.~Zhu, Y.~Liu, Z.~Dong, Y.~Wang, T.~Jiang, W.~Wang, and B.~Yang,
  ``Ada\textsc{F}it: Rethinking learning-based normal estimation on point
  clouds,'' in \emph{Proceedings of the IEEE International Conference on
  Computer Vision (ICCV)}, 2021, pp. 6118--6127.

\bibitem{zhang2022geometry}
J.~Zhang, J.-J. Cao, H.-R. Zhu, D.-M. Yan, and X.-P. Liu, ``Geometry guided
  deep surface normal estimation,'' \emph{Comput.-Aided Des.}, vol. 142, p.
  103119, 2022.

\bibitem{cao2021latent}
J.~Cao, H.~Zhu, Y.~Bai, J.~Zhou, J.~Pan, and Z.~Su, ``Latent tangent space
  representation for normal estimation,'' \emph{IEEE Trans. Ind. Electron.},
  vol.~69, no.~1, pp. 921--929, 2021.

\bibitem{zhou2022refine}
H.~Zhou, H.~Chen, Y.~Zhang, M.~Wei, H.~Xie, J.~Wang, T.~Lu, J.~Qin, and X.-P.
  Zhang, ``Refine-\textsc{N}et: Normal refinement neural network for noisy
  point clouds,'' \emph{IEEE Trans. Pattern Anal. Mach. Intell.}, 2022.

\bibitem{he2016deep}
K.~He, X.~Zhang, S.~Ren, and J.~Sun, ``Deep residual learning for image
  recognition,'' in \emph{Proceedings of the IEEE Conference on Computer Vision
  and Pattern Recognition (CVPR)}, 2016, pp. 770--778.

\bibitem{silberman2012indoor}
N.~Silberman, D.~Hoiem, P.~Kohli, and R.~Fergus, ``Indoor segmentation and
  support inference from rgbd images,'' in \emph{Proceedings of the European
  Conference on Computer Vision (ECCV)}, 2012, pp. 746--760.

\bibitem{cazals2005estimating}
F.~Cazals and M.~Pouget, ``Estimating differential quantities using polynomial
  fitting of osculating jets,'' \emph{Comput. Aided Geom. Des.}, vol.~22,
  no.~2, pp. 121--146, 2005.

\bibitem{zhou2016fast}
Q.-Y. Zhou, J.~Park, and V.~Koltun, ``Fast global registration,'' in
  \emph{Proceedings of the European Conference on Computer Vision (ECCV)},
  2016, pp. 766--782.

\bibitem{schnabel2007efficient}
R.~Schnabel, R.~Wahl, and R.~Klein, ``Efficient \textsc{RANSAC} for point-cloud
  shape detection,'' \emph{Comput. Graph. Forum}, vol.~26, no.~2, pp. 214--226,
  2007.

\end{thebibliography}

\begin{IEEEbiography}[{\includegraphics[width=1in,height=1.25in,clip,keepaspectratio]{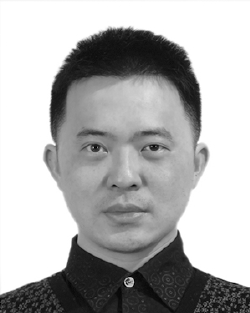}}]{Zheng Liu} is currently an associate professor at China University of Geosciences (Wuhan). He received the Ph.D. degree from Central China Normal University in 2012. From 2013 to 2014, he held a post-doctoral position with School of Mathematical Sciences, University of Science and Technology of China.
His research interests include geometry processing, computer graphics, 3D computer vision, and deep learning.
\end{IEEEbiography}

\begin{IEEEbiography}[{\includegraphics[width=1in,height=1.25in,clip,keepaspectratio]{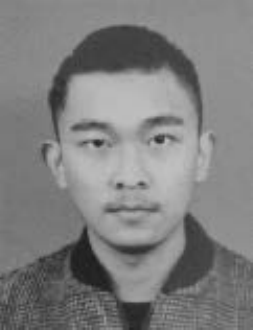}}]{Yaowu Zhao} is currently a M.S. candidate in China University of Geosciences (Wuhan).
He received the B.S. degree from China University of Geosciences (Wuhan), in 2021.
His research interests include geometry processing and deep learning.
\end{IEEEbiography}

\begin{IEEEbiography}[{\includegraphics[width=1in,height=1.25in,clip,keepaspectratio]{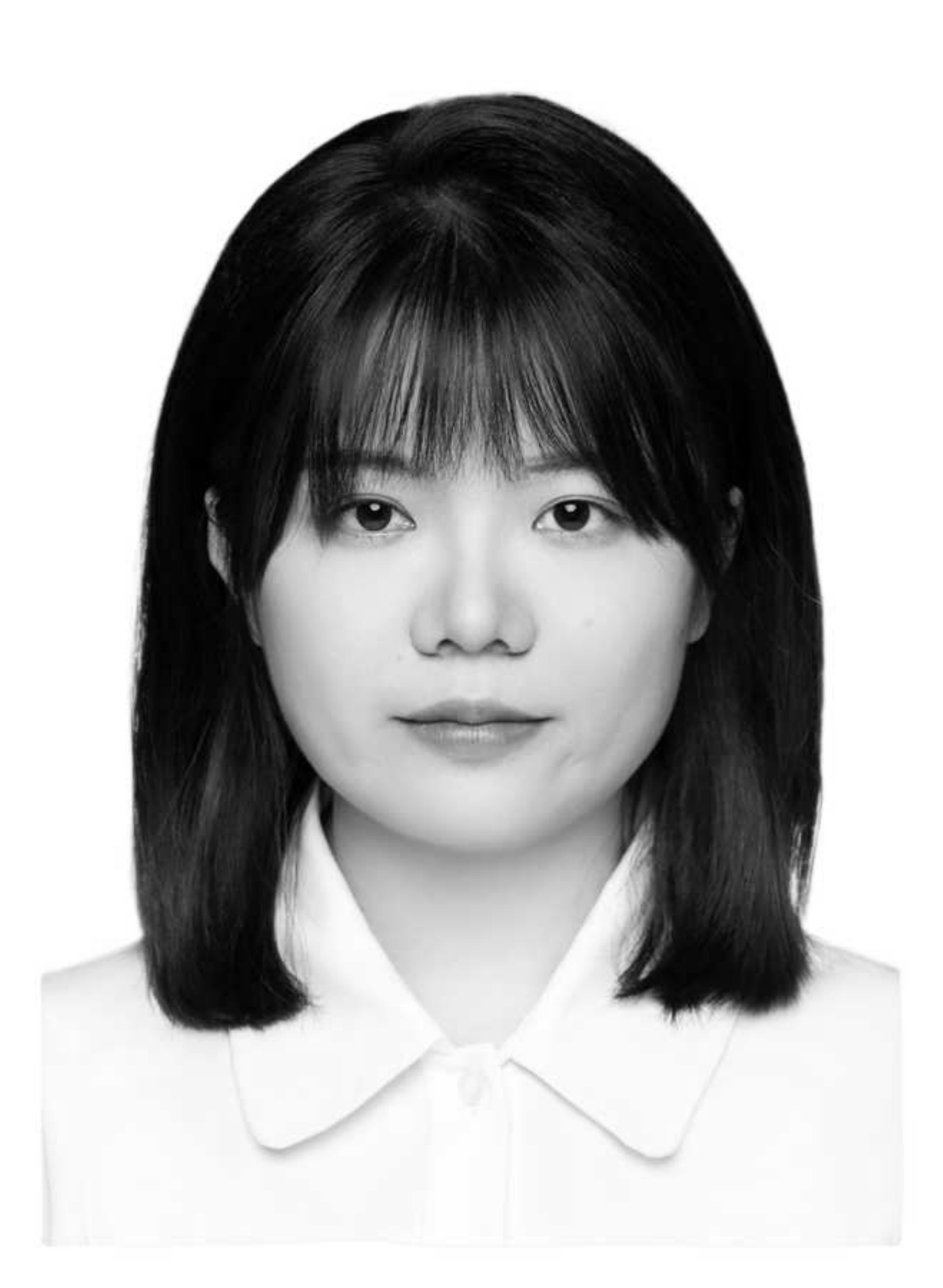}}]{Sijing Zhan} is currently a M.S. candidate at China University of Geosciences (Wuhan).
He received B.S. degree from Jimei University, in 2020.
Her research interests include geometry processing, 3D vision and deep learning.
\end{IEEEbiography}

\begin{IEEEbiography}[{\includegraphics[width=1in,height=1.25in,clip,keepaspectratio]{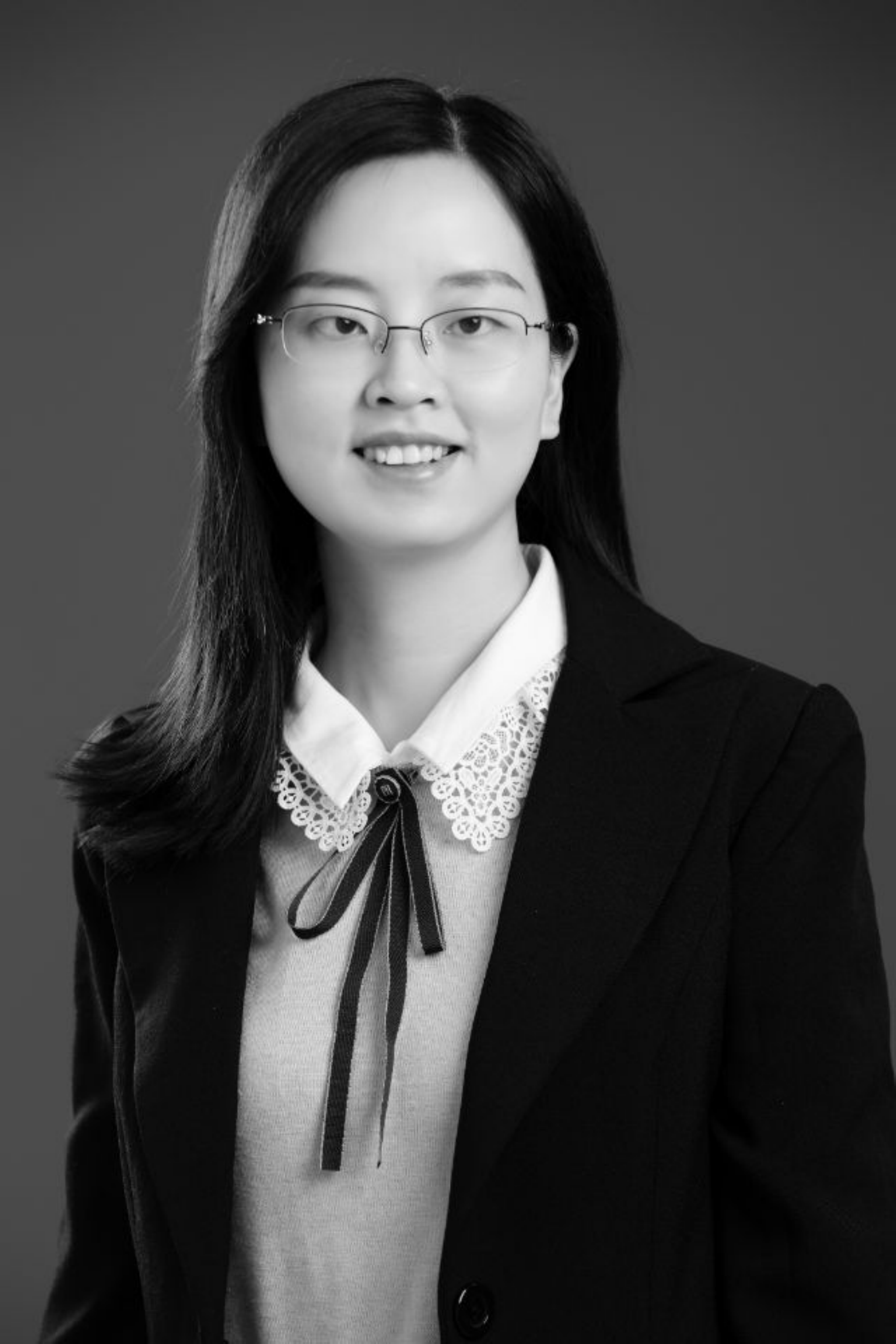}}]
{Yuanyuan Liu} received the B.E.
degree from Nanchang University, Nanchang, China,
in 2005, the M.E. degree from the Huazhong University of Science and Technology, Wuhan, China,
in 2007, and the Ph.D. degree from Central China
Normal University, Wuhan.
She is an associate professor at the China University of
Geosciences, Wuhan. Her research interests include
image processing, computer vision, and pattern
recognition.
\end{IEEEbiography}

\begin{IEEEbiography}[{\includegraphics[width=1in,height=1.25in,clip,keepaspectratio]{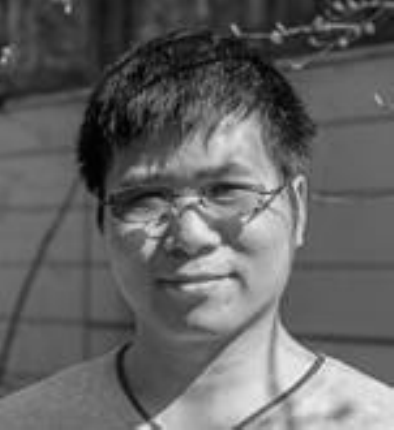}}]{Renjie Chen} is a professor at the University of Science and Technology of China (USTC). He holds a PhD degree from Zhejiang University, China. Before joining USTC, he was a postdoctoral fellow at the Technion--Israel Institute of Technology, a postdoctoral research associate at the University of North Carolina at Chapel Hill, a key researcher in the BeingThere Center in Nanyang Technological University, Singapore, and a senior researcher heading a research group working on 3D geometry and images at the Max Planck Institute for Informatics (MPII) in Saarbrucken, Germany. His research interests includes computer graphics, geometry modeling, computational geometry and glasses-free 3D display.
\end{IEEEbiography}

\begin{IEEEbiography}[{\includegraphics[width=1in,height=1.25in,clip,keepaspectratio]{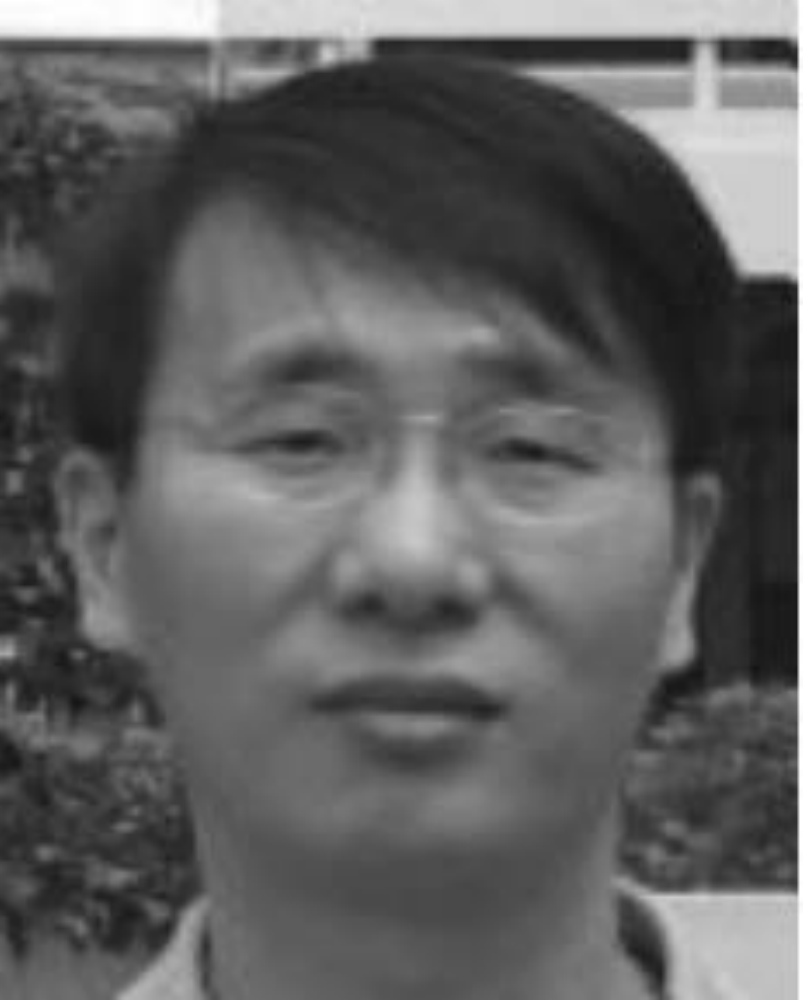}}]{Ying He} is currently an associate professor at
School of Computer Science and Engineering,
Nanyang Technological University, Singapore.
He received the BS and MS degrees in electrical
engineering from Tsinghua University, China,
and the PhD degree in computer science from
Stony Brook University, USA. His research interests
fall into the general areas of visual computing
and he is particularly interested in the
problems which require geometric analysis and
computation.
\end{IEEEbiography}

%\input{Appendix}
%\bigvee\input{SupplementaryMaterial}

\end{document}